\documentclass[sigconf, nonacm]{acmart}
\usepackage{wrapfig, blindtext}
\usepackage{balance}
\usepackage{xr-hyper}
\usepackage{hyperref}
\usepackage{algorithm}
\usepackage{algorithmic}
\usepackage{mathtools}
\usepackage{multirow}
\usepackage{graphics}
\usepackage{textcomp}

\usepackage[utf8]{inputenc}

\usepackage{subfigure}

\newcommand*\commentETH[2][\textbf{[ETH:]} ]{\colorbox{blue}{\small \textcolor{yellow}{#1#2}}}

\newcommand\vldbdoi{10.14778/3494124.3494128}
\newcommand\vldbpages{427 - 436}
\newcommand\vldbvolume{15}
\newcommand\vldbissue{3}
\newcommand\vldbyear{2022}
\newcommand\vldbauthors{Susie Xi Rao, Shuai Zhang, Zhichao Han, Zitao Zhang, Wei Min, Zhiyao Chen, Yinan Shan, Yang Zhao, Ce Zhang}
\newcommand\vldbtitle{\shorttitle} 
\newcommand\vldbavailabilityurl{https://github.com/eBay/xFraud}
\newcommand\vldbpagestyle{plain} 

\usepackage{enumitem}

\makeatletter
\newcommand*{\addFileDependency}[1]{
  \typeout{(#1)}
  \@addtofilelist{#1}
  \IfFileExists{#1}{}{\typeout{No file #1.}}
}
\makeatother

\newcommand*{\myexternaldocument}[1]{%
    \externaldocument{#1}%
    \addFileDependency{#1.tex}%
    \addFileDependency{#1.aux}%
}

\myexternaldocument{RevisionCoverLetter}

\begin{document}

\title{xFraud: Explainable Fraud Transaction Detection}
\author{Susie Xi Rao$^\dagger$, Shuai Zhang$^\dagger$, Zhichao Han$^\star$, Zitao Zhang$^\star$, Wei Min$^\star$, Zhiyao Chen$^\star$, Yinan Shan$^\star$, Yang Zhao$^\star$, Ce Zhang$^\dagger$}
\affiliation{%
    \institution{ETH Zurich$^\dagger$ \hspace{1
    cm} eBay China$^\star$}
}
\email{{raox,shuazhang,ce.zhang}@inf.ethz.ch}\email{{zhihan,zitzhang,wmin,zhiyachen,yshan,yzhao5}@ebay.com}

\begin{abstract}
At online retail platforms, it is crucial to actively detect the risks of transactions to improve customer experience and minimize financial loss. In this work, we propose xFraud, an explainable fraud transaction prediction framework which is mainly composed of a detector and an explainer. The xFraud detector can effectively and efficiently predict the legitimacy of incoming transactions. Specifically, it utilizes a heterogeneous graph neural network to learn expressive representations from the informative heterogeneously typed entities in the transaction logs. The explainer in xFraud can generate meaningful and human-understandable explanations from graphs to facilitate further processes in the business unit. In our experiments with xFraud on real transaction networks with up to 1.1 billion nodes and 3.7 billion edges, xFraud is able to outperform various baseline models in many evaluation metrics while remaining scalable in distributed settings. In addition, we show that xFraud explainer can generate reasonable explanations to significantly assist the business analysis via both quantitative and qualitative evaluations.

\end{abstract}

\maketitle

\pagestyle{\vldbpagestyle}
\begingroup\small\noindent\raggedright\textbf{Full Paper Reference:}\\
\vldbauthors.
\vldbtitle. PVLDB, \vldbvolume(\vldbissue): \vldbpages, \vldbyear.\\
\href{https://doi.org/\vldbdoi}{doi:\vldbdoi}
\endgroup
\begingroup
\renewcommand\thefootnote{}\footnote{\noindent
This work is licensed under the Creative Commons BY-NC-ND 4.0 International License. Visit \url{https://creativecommons.org/licenses/by-nc-nd/4.0/} to view a copy of this license. For any use beyond those
covered by this license, obtain permission by contacting the authors. Copyright is held
by the owner/author(s).\\
}\addtocounter{footnote}{-1}\endgroup

\ifdefempty{\vldbavailabilityurl}{https://github.com/eBay/xFraud}{
\vspace{.3cm}
\begingroup\small\noindent\raggedright\textbf{Artifact Availability:}\\
The source code, data, and/or other artifacts have been made available at \url{https://github.com/eBay/xFraud}.
\endgroup
}

\section{Introduction}


The online retail industry is reshaping our shopping behavior, and the resulting security risks are not negligible. 
Common threats in e-commerce include account acquisition, financial information theft, fake chargeback, money laundry, and many more. For instance, malicious attackers might try to steal customer's credit card information; the login credentials can also be acquired by hackers. These criminal activities can bring negative impacts on user experiences, cause financial losses, and seriously degrade the platform credibility. As such, it is critical to identify fraudulent behaviors and take every precaution to minimize risks. 
\begin{figure}[!t]
    \centering
    \begin{tabular}{c}
       \includegraphics[width=0.65\linewidth]{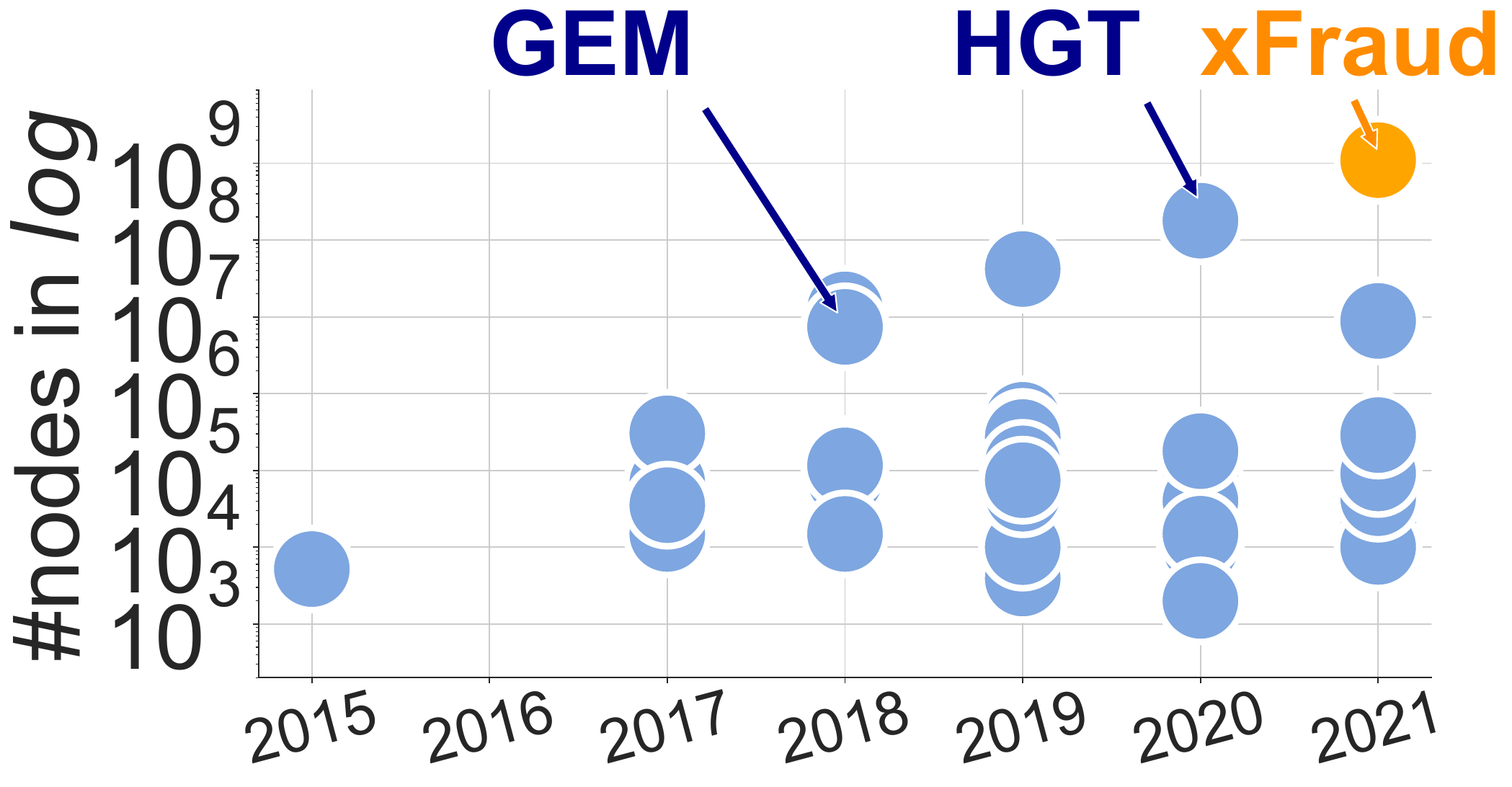}  \\ \includegraphics[width=0.65\linewidth]{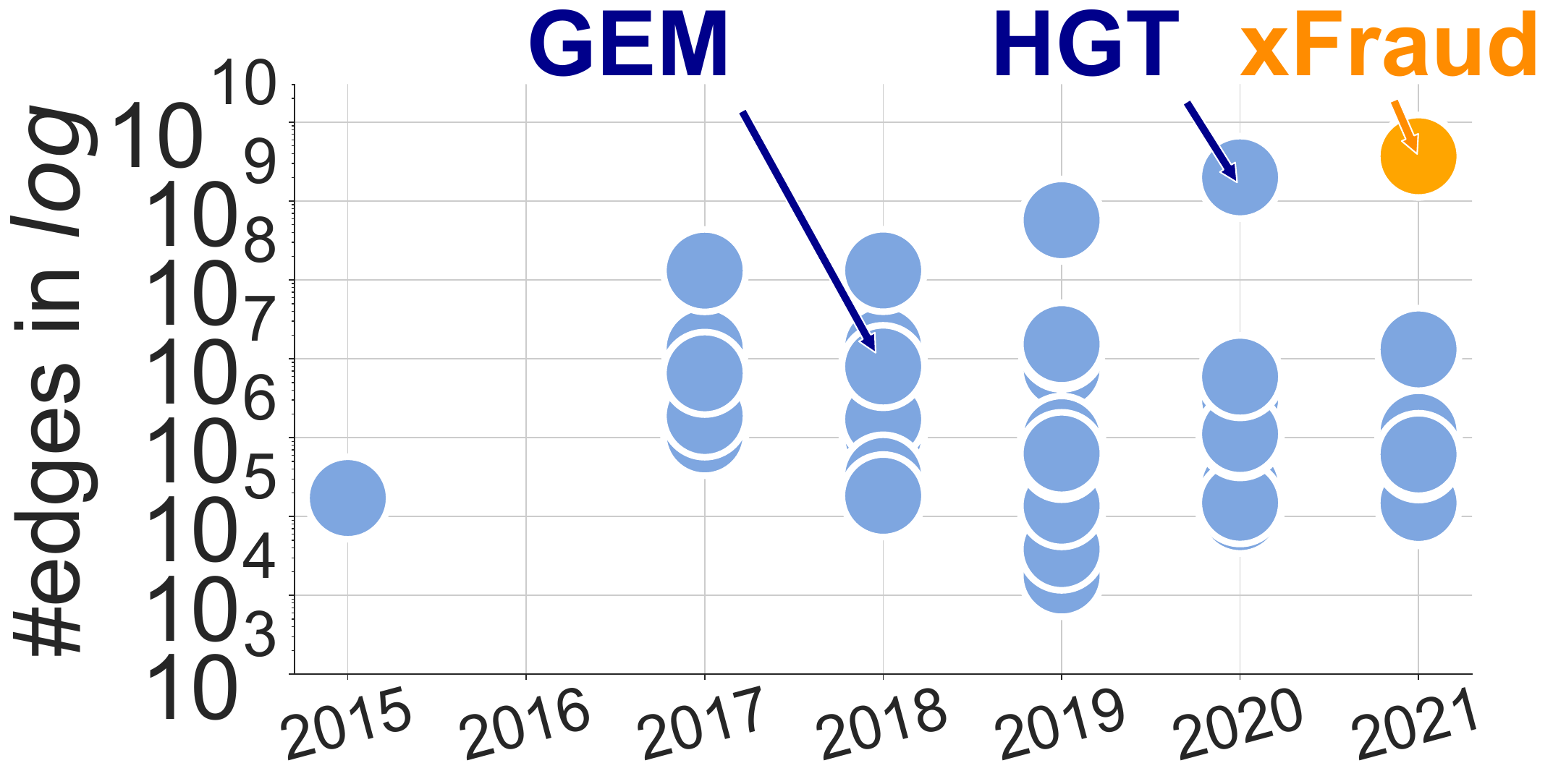} \\
    \end{tabular}
    \caption{The node and edge numbers ($log$) of heterogeneous graph datasets in the literature. Full survey in Appendix~\ref{app:survey-hetero}~\cite{xfraudAppendix}.}
    \label{fig:node-edge-compare}
\end{figure}

\textit{Fraud detection} has been an emerging topic for e-commerce and social media companies. It is studied in various applications: malicious account detection (e.g., social networks~\cite{breuer2020friend}, online payment systems~\cite{zhong2020financial}, and online retailer platforms~\cite{liu2018heterogeneous, cao2017hitfraud,zhu2020modeling,cao2019titant});  anti-money laundry \cite{emmerich2017flowscope,weber2019anti}; spam reviews and news detection \cite{wang2019fdgars,shu2020hierarchical}.
Among these \textit{transaction fraud detection} is an important topic~\cite{cao2017hitfraud,zhu2020modeling,cao2019titant}. In this work, we focus on automatic fraudulent transaction detection in a real-world e-commerce environment at eBay. For an incoming transaction, we aim to predict whether it is legitimate or not.

\textbf{Challenges.}
Despite recent efforts in automatic fraudulent transaction detection~\cite{shu2020hierarchical,dhawan2019spotting, kaghazgaran2018combating,zhong2020financial, hu2019cash, breuer2020friend, eswaran2017zoobp, nilforoshan2019slicendice, liu2017holoscope, wang2018deep,wen2020asa, zhang2019key, liu2018heterogeneous, li2019spam} with 
machine learning such as LSTM and graph neural networks (GNNs),
we realize that three challenges still linger when
it comes to our application scenario at eBay.

\textbf{\textit{(1. Information Heterogeneity)}} In our system, there are heterogeneous types of information concerning a transaction such as payment tokens, shipping addresses, email. Intuitively, such information is indicative in fraud detection. \textit{How to effectively utilize such information in an end-to-end ML model?} 

\textbf{\textit{(2. Scalability and Efficiency)}} Our platform can produce millions of transactions involving millions of users in a short span of time, which requires the detecting system to be \textit{efficient and scalable for practical use}.
Specifically, Figure~\ref{fig:node-edge-compare} shows the
landscape of heterogeneous graphs emerging in the last six years.
In this paper, we are tackling a workload, to our best knowledge,
that consists of one of the largest \textit{heterogeneous} graphs for graph neural networks (see a more detailed survey in Appendix~\ref{app:survey-hetero}~\cite{xfraudAppendix}). This poses unique challenges in 
the system design and optimizations.

\textbf{\textit{(3. Explainability)}} 
Flagging a transaction to be fraudulent is not a trivial process. False decisions are likely to cause trouble to our customers and significantly degrade the platform's credibility. In general, this process requires extensive human efforts in cautiously reviewing the model's prediction, which is inefficient and costly. 
\textit{How can we explain the outcome of 
an ML model, and more importantly, how 
close these explanations are to those developed by 
human experts in the business unit (BU)?} 

\textbf{Our Approach.} To tackle the aforementioned challenges, we propose xFraud, an explainable fraud detection framework at eBay. xFraud is not only able to efficiently and effectively predict the legitimacy of a transaction but can also generate human readable explanations in to assist flagging transaction frauds. xFraud advances previous work in two ways.

First, xFraud builds upon a heterogeneous GNN to tackle \textit{transaction fraud detection}.
Specifically, xFraud consists of a detector and an explainer. In the detector, we tackle a transaction fraud detection task from the graph perspective. Different from the existing works~\cite{ma2018graphrad, liu2019geniepath, liang2019uncovering}, a heterogeneous graph of different node types (e.g., transactions, addresses, payment tokens) is constructed. To capture the heterogeneous relation patterns and learn more expressive node representations, a self-attentive heterogeneous graph neural network is adopted. The detector can automatically aggregate information from different types of nodes via disparate paths without manually predefined meta-paths. This is important because we do not need to predefine the meta-paths of risk propagation and preprocess the path representations, as required in~\cite{hu2019cash, shu2020hierarchical, zhong2020financial, cao2017hitfraud}.
Additionally, we explore an efficient sampler in the message aggregation procedure, which empirically reduces the inference time significantly while achieves a competitive performance. 

Second, to the best of our knowledge, we provide the first quantitative evaluation of input-level GNN explanation methods~\cite{ying2019gnnexplainer, li2020gnnexplainer-fin} and its agreement with humans in a real-world application scenario.
Unlike traditional classification tasks, the claim that a transaction is fraudulent should be made very cautiously to avoid harming customer experience and degrading the platform's credibility. As such, we integrate an explainer into our framework that can provide intuitive explanations for model predictions. Equipped with these explanations, our auditors, regulators, or decision makers know how a transaction is flagged by the detector, thus making more sensible decisions. 
One open question is \textit{how well these explanations 
agree with the insights from human experts}.
To this end, we conduct an extensive quantitative study to measure the agreement between human perception, GNNExplainer~\cite{ying2019gnnexplainer}, and centrality measures, as well as provide case studies on how these 
explanations can help in practice.
This study also reveals an interesting
tradeoff between GNN-based explanations and traditional 
topological measures (e.g., centrality), which allows 
us to design a hybrid explainer that outperforms both strategies.

\textbf{In summary, our technical contributions are as follows.}

\textbf{(1)} We propose a heterogeneous GNN model (detector) to identify transaction frauds. Our model  captures the heterogeneity in transaction graphs and applies to industrial-scale datasets. xFraud detector provides concrete analytical angles of fraudulent activities.
Compared with HGT~\cite{hu2020hgt}, in xFraud, we design a new sampling mechanism, inspired by the sparsity of our underlying graph; compared with GEM~\cite{liu2018heterogeneous}, xFraud uses a heterogeneous GNN architecture, which allows it to outperform a GEM-style model significantly.
    
\textbf{(2)} We add explainability into xFraud with a hybrid explainer. xFraud explainer computes the contributions of its neighboring node types and edges when predicting a node, and it also attends to global topological features learned from centrality measures. Thus, it enables explicit case studies of network predictions, which is beneficial for model trustworthiness. We perform the first quantitative evaluation between GNNExplainer and human judgments. We also compare edge weights computed via centrality measures with the weights learned by GNNExplainer, through which we identify a trade-off and propose a hybrid explainer in xFraud.
    
\textbf{(3)} We conduct careful system design and optimizations,
    which allow us to scale out, to our best knowledge, to one of 
    the 
    largest heterogeneous graphs being reported for ML workloads
    so far.
    
\textbf{(4)} We conduct experiments on 
real-world transaction networks to show the efficiency of xFraud in detecting transaction frauds and in facilitating the analysis of graph structural patterns.


\begin{figure*}[t]
    \centering
    \includegraphics[width=0.8\linewidth]{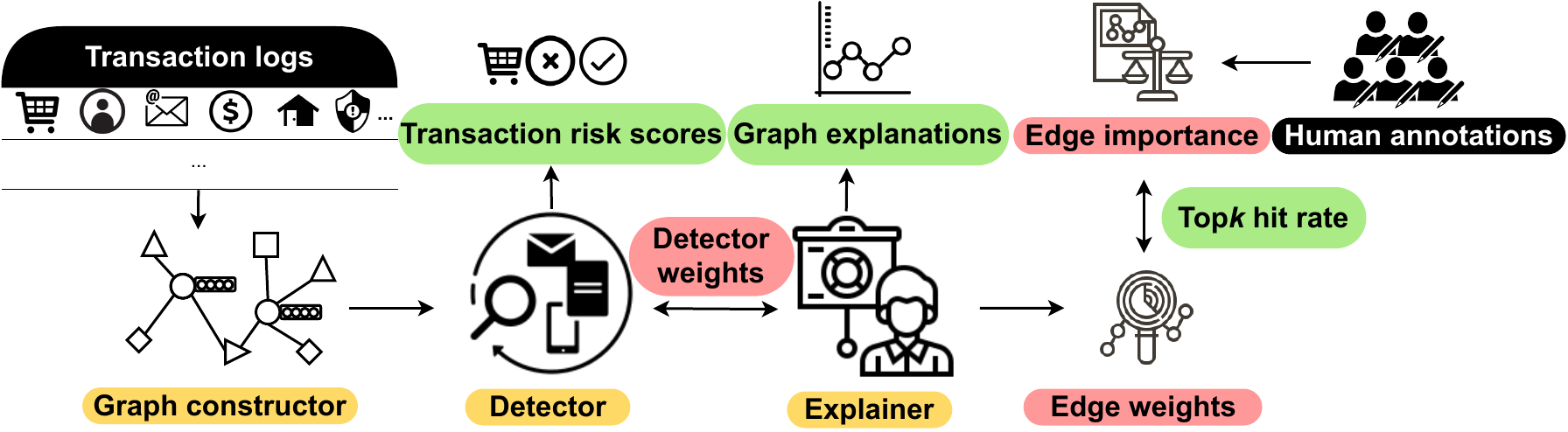}
    \caption{xFraud pipeline.}
    \label{fig:pipeline}
\end{figure*} 
\begin{figure*}[t]
    \centering
    \includegraphics[width=0.9\linewidth]{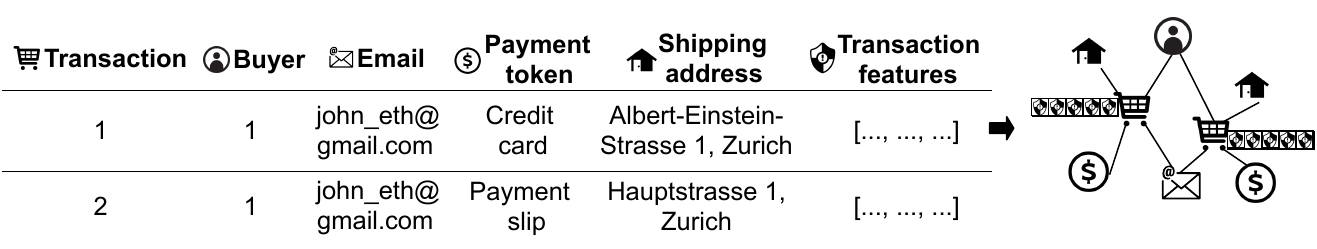}
    \caption{Transactions $\rightarrow$ a heterogeneous graph.}
    \label{fig:constructor}
\end{figure*} 

\section{Related Work}\label{sec:lit}

xFraud builds upon recent successes of 
applying heterogeneous graphs to fraud detection
(e.g., MAHINDER~\cite{zhong2020financial}
and GEM~\cite{liu2018heterogeneous} from Alibaba)
and also recent efforts of
GNN explainability \cite{ying2019gnnexplainer}.
However, it also
makes significant improvements over these
previous efforts, discussed as follows.

\textbf{Fraud Detection.}
There are two lines of studies on fraud detection systems. One line of work leverages graph information with non-GNN methods (see anti-money laundry~\cite{emmerich2017flowscope}, spam detection~\cite{shu2020hierarchical, dhawan2019spotting, kaghazgaran2018combating}, fraudster user detection~\cite{zhong2020financial, hu2019cash, kumar2018rev2, breuer2020friend, eswaran2017zoobp, nilforoshan2019slicendice, liu2017holoscope, hooi2016fraudar, li2018hgsuspector, wang2018deep}, fraud transaction detection
~\cite{cao2019titant, ren2019ensemfdet, zhu2020modeling, cao2017hitfraud}). These models can be event/sequence based~\cite{cao2019titant, breuer2020friend, zhu2020modeling, li2019spam}, or meta-path based~\cite{hu2019cash, shu2020hierarchical, zhong2020financial, cao2017hitfraud}. 
Zhong et al. \cite{zhong2020financial} propose MAHINDER which 
uses \textit{heterogeneous graphs} in the context of 
defaulter detection by pre-defining 
a set of \textit{meta-paths} in a heterogeneous graph of users and merchants. 
The preprocessed meta-path feature representations are trained with an attention mechanism and LSTM to measure the importance of nodes, links, and meta-paths at different timestamps. 
In xFraud, we focus on a different scenario,
aiming at flagging each transaction of a user in various risk scenarios, as a legitimate user does not imply that all its transaction records are legitimate, e.g., once its payment token has been stolen. 
More importantly, we \textit{focus on 
methods that do not need to define meta-paths a priori, instead are able to automatically learn these
patterns using a GNN}.

The other line of work uses GNN methods~\cite{li2019spam, wen2020asa, zhang2019key, liu2018heterogeneous, wang2019fdgars, weber2019anti, ma2018graphrad, liu2019geniepath, liang2019uncovering}. Homogeneous graph has been widely applied in e-commerce applications (see spam review detection~\cite{wang2019fdgars}, anti-money laundry~\cite{weber2019anti}, risky/malicious account detection~\cite{ma2018graphrad, liu2019geniepath, liang2019uncovering}). Recently, people start to solve real-world anomaly detection problems using heterogeneous graph (see spam review detection~\cite{li2019spam}, suspicious user detection~\cite{wen2020asa, zhang2019key, liu2018heterogeneous}) or to combine homogeneous and heterogeneous graphs \cite{li2019spam}, because it allows aggregating information propagation through various types of nodes/edges. GEM~\cite{liu2018heterogeneous} by Liu et al.~has utilized attention mechanisms in a device-account heterogeneous graph to capture user activity and device embeddings in each subgraph neighborhood. The heterogeneous graph in GEM is then fed into a GCN.

\textbf{Explainability in GNN.}
Recently, how to interpret and explain GNN predictions has gained spotlight. There are two levels to explain: input level (GraphConsis\cite{liu2020alleviating}, GNNExplainer \cite{ying2019gnnexplainer}, GraphLIME \cite{huang2020graphlime}, \cite{baldassarre2019explainability, li2020gnnexplainer-fin}), and model level (XGNN \cite{yuan2020xgnn}). GNNExplainer \cite{ying2019gnnexplainer}, a GNN model agnostic explanation framework, proposes explaining the GNN predictions by maximizing the mutual information gain of the true node labels and the predicted labels using informative features. GNNExplainer enables a visualization of important subgraph patterns, which assists users to understand the feature contribution and node label propagation. GNN explainability in the financial domain has been addressed by Li et al.~\cite{li2020gnnexplainer-fin}. They have extended GNNExplainer techniques by (1) adding a regularization term that ensures at least one edge connected to each node is selected in the subgraph, (2) adding edge weighted graph attention to calculate the edge weights in the subgraph. Using GCN in a node classification task, they applied an explainer to identify informative graph patterns on financial transaction data like bitcoin over the counter (OTC) and account matching in bank transactions.

\section{The xFraud Framework}

Figure \ref{fig:pipeline} illustrates the xFraud pipeline in a nutshell. First, we build a graph constructor to convert transaction logs into a graph abstraction (Sec. \ref{sec:hetG-constructor}), which is then fed into a detector to generate a transaction risk score for each transaction record (Sec. \ref{sec:detector-all}). Then, to build a learnable hybrid explainer (Sec. \ref{sec:gnnexplainer}), we combine the task-aware measures of predictions generated by the GNNExplainer, and the task-agnostic centrality measures. 

One highlight of xFraud over previous efforts
is its emphasis on explainability, especially 
its evaluation using insights
from real-world experts in the business unit.
We evaluate the efficacy of the explainer quantitatively (Sec. \ref{sec:explainer-eval-quantitative}) and qualitatively (Sec. \ref{sec:explainer-eval-qualitative}). Quantitatively, we calculate the agreement (top\textit{k} hit rate) between human annotations and explainer weights. Concretely, we first obtain human ground truth on edge importance in risk propagation. Then, we calculate edge importance scores from node importance scores with various aggregation methods. At last, we report the top\textit{k} hit rate between edge importance scores computed from human annotations and edge weights generated by the hybrid explainer. Qualitatively, we study in detail many cases where xFraud explainer assists the BU in better understanding complex fraudulent patterns. 


\subsection{Heterogeneous Graph Construction}\label{sec:hetG-constructor}
Think of the critical entities involved under fraud scenarios. A credit card might be linked to both a legitimate user and a fraudulent user at different stages. The latter happens in a card stolen case. A common shipping address such as a warehouse is sometimes used in frauds. This linkage tends to be stable, compared with stolen financial instruments.
If we formulate fraud detection as a semisupervised learning problem in an inductive setting \cite{hamilton2017inductive-sage} in a heterogeneous graph, we have the specification of the problem formulation as follows. In a heterogeneous transaction graph $\mathcal{G}$, $v \in \mathcal{V}$ has a type $\tau(v) \in \mathcal{A}$, where $\mathcal{A}:= \{txn, pmt, email, addr, buyer\}$, referring to \textit{transaction, payment token, email, shipping address, buyer}, respectively\footnote{For this study, we choose these attributes based on the homophilic tests \cite{baesens2015fraud}. It is shown that fraud exhibits homophilic effects \cite{min2018behavior}, and entities with strong homophilic effects are considered in this work.}. If a transaction has relation with another type of node in $\{$\textit{pmt, email, addr, buyer}$\}$, we put an edge between those two nodes in the heterogeneous graph. Each $txn$ node carries node attributes provided by a risk identification system. A transaction is represented as an ID in the transaction log. The item category in the purchase order relevant to one transaction (item-type info) is encoded in the transaction features. Each transaction is flagged legit or fraud. Figure \ref{fig:constructor} illustrates how to construct such a heterogeneous graph based on two transaction records sharing several entities.
\subsection{The detector of xFraud}\label{sec:detector-all}
As we have seen in the literature, it is common to define meta-paths when analyzing graph structured data and then to extract corresponding features of nodes and edges on the meta-paths before feeding the features into a machine learning or deep learning model. However, in a fraud detection scenario, under many circumstances, it is by nature impossible to enumerate every possible scenario and their influential meta-paths. This is also one of the primary intuitions why a heterogeneous GNN is a desirable choice: it allows a network to learn the importance of meta-paths by itself  based on the network structure and message passing.

\subsubsection{\textbf{xFraud detector}}\label{sec:detector}

We are inspired by Transformer \cite{vaswani2017attention} and HGT \cite{hu2020hgt}, when designing the xFraud detector incl.~  heterogeneous mutual attention and heterogeneous message passing with key, value, and query vector operations (self-attention mechanism). We do not allow target-specific aggregation on different node types, so that we reduce the cost in computing different weights for various node types. We see a better performance in our detector (see discussion in Sec.~\ref{sec:scalable-xfraud}), when shared weights among different types of nodes are used. 
%
Moreover, we do not adopt relative temporal encoding in HGT when processing transactions with timestamps. Reasons are, we would like to keep track of all transactions a buyer executes, as well as the linking entities a transaction involves.  We also model the relations between buyers and transactions. This makes our system adaptable to guest checkouts and their pertaining chargebacks, as those transactions could not be linked to any buyer accounts, but they could be linked to suspicious third-party payment accounts or billing email addresses. In this manner, our system is able to capture disguised/missed fraud patterns of guest checkouts that could otherwise be neglected by (1) representing a transaction using buyers and timestamps as in HGT or by (2) representing the transactions in a homogeneous graph. 


{\bf Comparison to GEM.}
Looking at the application scenarios, xFraud might seem similar to GEM (a malicious account detection system developed by Alibaba). However, xFraud differs from GEM~\cite{liu2018heterogeneous} in that: (1) GEM is a system which directly applies a vanilla GCN to a heterogeneous graph, while our proposed xFraud considers the heterogeneous property of graphs in the underlying architecture (e.g., sampler, heterogeneous graph convolution). In this paper, we choose GEM as a representative of heterogeneous GCN in the evaluation. 
(2) We focus on a different application and have different node types. GEM focuses on fraudsters' detection, while we aim to find the anomaly transactions (a user may have both fraudulent and normal transactions due to account hacking). (3) The GEM model does not provide any explanation for its predictions. We have extensively discussed the GNN explainability and conducted experiments to understand the xFraud explainer. 

\begin{figure*}[!t]
    \centering
    \includegraphics[width=0.7\linewidth]{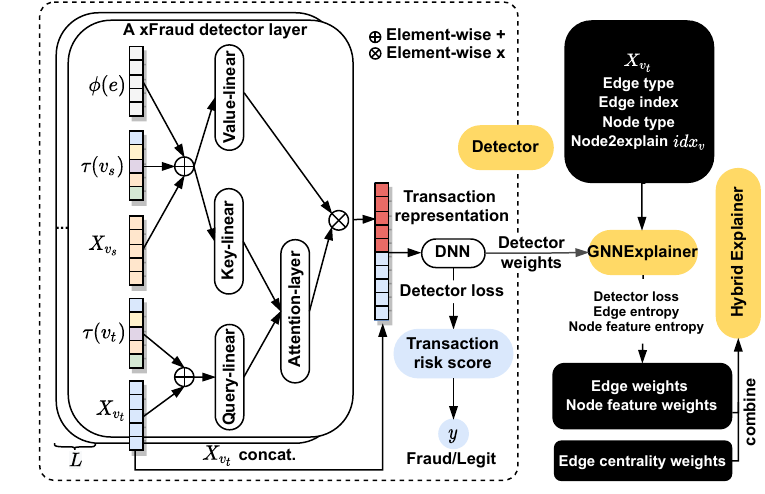}
    \caption{xFraud detector and explainer.} 
    \label{fig:detector-explainer}
\end{figure*}
In Figure \ref{fig:detector-explainer}, we show the detector architecture in detail (left).

\textbf{(1)} The detector takes a heterogeneous graph as input, incl.~ target and source node features $X_{v_{t}}$, $X_{v_{s}}$; target and source node types $\tau(v_t)$, $\tau(v_s)$; edge type $\phi(e)$. For the \textit{txn} nodes, we have node features computed by a company risk identifier. For the other node types, the initial node features are empty and only get their inputs after the first convolution layer. The type features are in one-hot encoding of types. 

\textbf{(2)} $L$ \textit{heterogeneous convolution layers} process the graph with self-attention mechanism: the input layer $\text{L}^{(0)}$ takes transaction features, node type embeddings (source and target), edge type embeddings as input, which are transformed into query, key, and value vectors. Attention scores are calculated for the source and target nodes and then layer-wise normalized, which are then fed into a \textit{ReLU} activation function that emits input for the next convolution layer. In Sec.~\ref{sec:hgnn}, we introduce with equations how  heterogeneous mutual attention and message passing function in one heterogeneous convolution layer. 

\textbf{(3)} After $L$ heterogeneous convolution layers, a \textit{tanh} activation is applied to the transaction representations generated by GNN. Then these representations are concatenated with the original transaction features and fed into a feedforward connected network with two hidden layers. We then apply dropout, layer normalization, and \textit{ReLU} transformation to calculate a predicted risk score and a label.

\textbf{(4)} The loss function of xFraud detector is the cross entropy of the true label and the probability score calculated by \textit{softmax} (see eq.~\ref{eq:detector-loss} in Appendix~\ref{app:gnnexplainer}~\cite{xfraudAppendix}).

\subsubsection{\textbf{Heterogeneous Convolution Layer in xFraud Detector.}}\label{sec:hgnn}
We discuss the details of a xFraud detector layer shown in Figure~\ref{fig:detector-explainer}. For a tuple $\langle\tau(v_s), \phi(e), \tau(v_t)\rangle$, where $e=(v_s,v_t)$, we initialize (1) the node type embeddings $\tau(v)^{emb}$ and the edge type embeddings $\phi(e)^{emb}$ with zero weights; (2) the attention weight matrices of source node $W^{att}_{\tau(v_s)}$ and of target node $W^{att}_{\tau(v_t)}$ with random weights subject to uniform distributions; and (3) the weight matrices for key, query, and value vectors denoted by $W^K$, $W^Q$, $W^V$, respectively, also with random values subject to uniform distributions.
In a nutshell, a general attention-based heterogeneous convolution layer of the node $v_t$ has three components, attention, message, and aggregate as shown in
\begin{equation}
      H^l[v_t] \leftarrow Aggregate (Attention(v_s, v_t) \cdot Message(v_s)) .
\end{equation}
For each target node $v_t$, we create \textit{query, key}, and \textit{value} vector representations for self-attention mechanism with multiheads. 

To construct the $i$th \textit{query} vector for the target node $v_{t}$, we start with an input to the first layer by taking the transaction features of the target node $X_{\tau(v_t)}^{txn}$ and its node type embedding $\tau(v_t)^{emb}$ to calculate
\begin{equation}
   Q^i(v_t) = \text{Q-Linear}^i_{\tau(v_t)}\Big(X_{\tau(v_t)}^{txn}+\tau(v_t)^{emb}\Big) ,
\end{equation}
Then, for $H^{(l-1)}$, where $l \neq 1$, we compute
\begin{equation}
    Q^i(v_t) = \text{Q-Linear}^i_{\tau(v_t)}\Big(H^{(l-1)}[v_t]\Big) , 
\end{equation}
where $H^{(l-1)}[v_t]$ is the node representation of the node $v_t$ on the $H^{(l-1)}$ layer. 

To construct the $i$th \textit{key} vector for the source node $v_{s}$, we start with an input to the first layer by taking the transaction features of the source node $X_{\tau(v_s)}^{txn}$, its node type embedding $\tau(v_s)^{emb}$ and the edge type embedding $\phi(e)^{emb}$ to calculate
\begin{equation}
    K^i(v_s) = \text{K-Linear}^i_{\tau(v_s)} \\ \Big(X_{\tau(v_s)}^{txn}+\tau(v_s)^{emb}+\phi(e)^{emb}\Big) ,
\end{equation}
Then, for $H^{(l-1)}$, where $l \neq 1$, we compute
\begin{equation}
    K^i(v_s) = \text{K-Linear}^i_{\tau(v_s)}\Big(H^{(l-1)}[v_s]\Big) ,
\end{equation}
where $H^{(l-1)}[s]$ is the node representation of the node $v_s$ on the $H^{(l-1)}$ layer.

To construct the $i$th \textit{value} vector for source node $v_{s}$, we start with an input to the first layer by the taking transaction features of source node $X_{\tau(v_s)}^{txn}$, its node type embedding $\tau(v_s)^{emb}$ and the edge type embedding $\phi(e)^{emb}$ to calculate
\begin{equation}
     V^i(v_s) = \text{V-Linear}^i_{\tau(v_s)}\Big(X_{\tau(v_s)}^{txn}+\tau(v_s)^{emb}+\phi(e)^{emb}\Big) ,
\end{equation}
Then, for $H^{(l-1)}$, where $l \neq 1$, we compute
\begin{equation}
    V^i(v_s) = \text{V-Linear}^i_{\tau(v_s)}\Big(H^{(l-1)}[v_s]\Big) .  
\end{equation}

We adopt the multiheaded attention to control the randomness of initial weights. First, we compute the attention output of one attention head, denoted by $\alpha\text{-head}^i(v_s,e,v_t)$,  using this equation
\begin{equation}
   \alpha\text{-head}^i(v_s,e,v_t) =  \frac{\Big(K^i(v_s)W^{att}_{\tau(v_s)} + Q^i(v_t)W^{att}_{\tau(v_t)}\Big)}{\sqrt{d_k}} ,
\end{equation}
where $\sqrt{d_k}$ is the square root of the key vector's dimension. 

The heterogeneous mutual attention of the target node query vector $Q^i(v_t)$ and the source node key vector $K^i(v_s)$ is then computed by 
\begin{equation}
    \alpha(v_s,e,v_t) = \underset{\forall v_s \in N(v_t)}{\text{softmax}}\Big(\underset{i \in [1,h]}{\parallel} \alpha\text{-head}^i(v_s,e,v_t) \Big) ,
\end{equation}
where $N(v_t)$ represents the neighbors of $v_t$, $h$ the number of attention heads, $\parallel$ vector concatenation.

Finally, the message passing between $H^{(l)}$ and $H^{(l-1)}$ is given by 
\begin{equation}
    \text{msg}(v_s,e,v_t)= \underset{i \in [1,h]}{\parallel}\Bigg(V^i(v_s) \cdot  \text{dropout} \Big(\alpha\text{-head}^i(v_s,e,v_t)\Big)\Bigg) , 
\end{equation}
where the right hand side is a concatenation of all $\text{msg-head}^i$, the message passing of one attention head at the $i$th query vector.
\begin{figure*}[t!]
    \centering
    \includegraphics[width=0.75\textwidth]{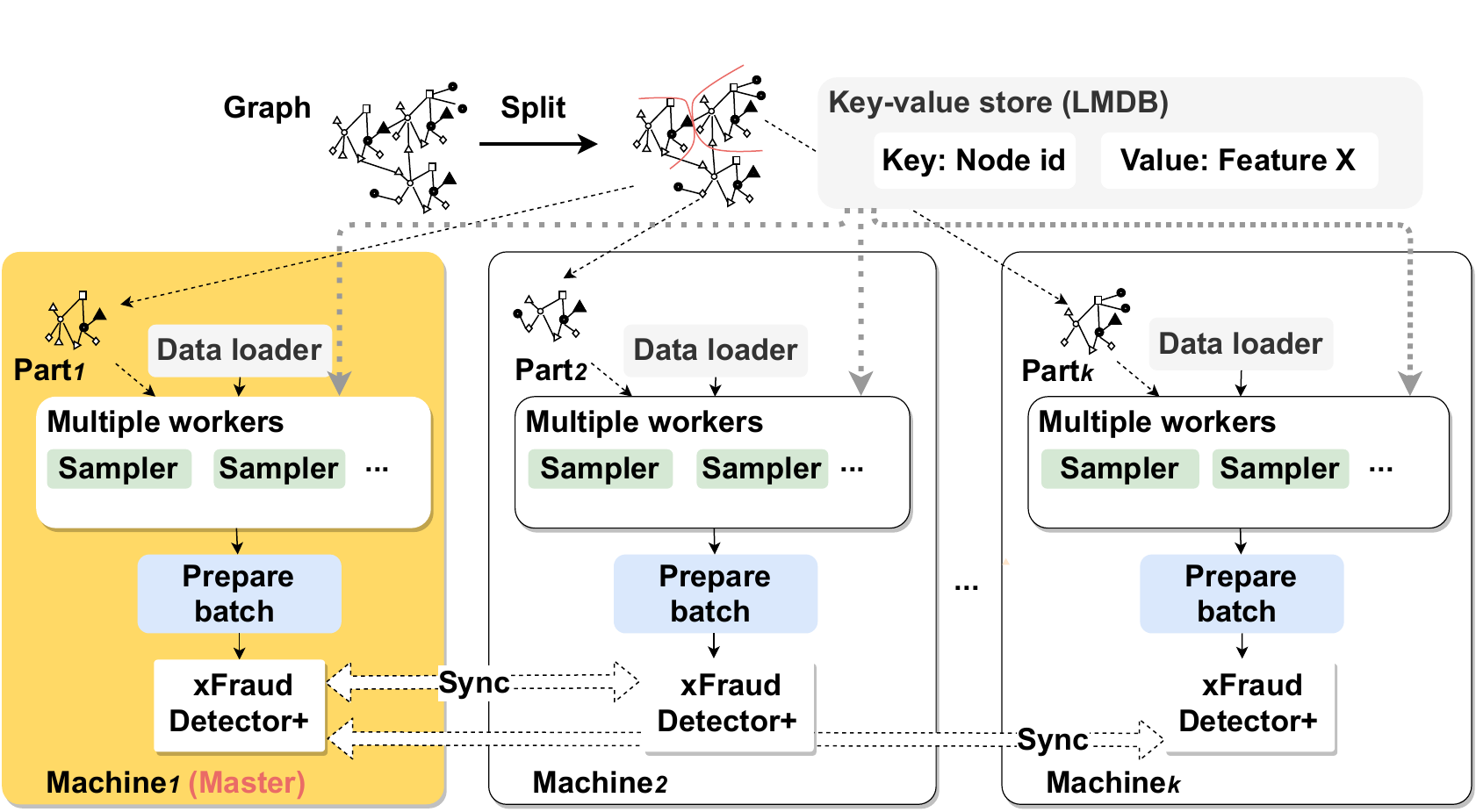}
    \caption{Architecture of Distributed xFraud Detector+.}
    \label{fig:multi-kv-distributed}
\end{figure*}
\subsubsection{\textbf{xFraud Detector+}: \textbf{An Improvement over HGT}}\label{sec:detector-plus}
We implement xFraud detector and detector+, whose difference lies in the sampler. In detector, we use the original HGT implementation\footnote{\url{https://github.com/acbull/pyHGT} (last accessed: Oct 18, 2020).} and empirically show that HGSampling is computationally costly (see the inference time in Figure~\ref{fig:inference-time}). Hence, we modify the sampling as in GraphSAGE and denote the efficient version of the xFraud detector as detector+. In detector+, the algorithm first samples $k$-hop neighborhood of a node and then aggregates feature information from neighbors and finally allows GNN to predict the label using aggregated information. In HGSampling, used by HGT, it tries to maintain a similar number of different $\tau(v)$ and $\phi(e)$ types after sampling and minimize the information loss and sample variance in the subgraph after sampling. 
However, in our datasets, the graph is much sparser (2.12 and 1.49 edges/node for \textit{eBay-small} and \textit{eBay-large}) compared with the Open Academic Graph (11.173 edges/node) used in HGT. Therefore,
HGSampling is more costly than GraphSAGE because it requires all types of nodes and edges to be of similar size in the sampled subgraph. In the following sections, we consider xFraud detector equivalent to HGT and mainly focus on the evaluation of xFraud detector+.


\subsection{Distributed xFraud Detector+}\label{sec:gnnexplainer}
To make xFraud detector+ scalable to industrial-scale datasets, we designed a distributed learning architecture (see Figure \ref{fig:multi-kv-distributed}). 
We briefly discuss its design and leave more details to Appendix~\ref{app:distributed}~\cite{xfraudAppendix}.

\subsubsection{\textbf{Graph Partitioning}}
We adopt the Power Iteration Clustering (PIC) algorithm \cite{lin2010power} to partition the graph according to pairwise similarities of edge properties. PIC is effective  for graph partition/clustering and well-suited to very large datasets due to its high efficiency.
Each worker takes charge of a different partition during
distributed training.

\subsubsection{\textbf{Distributed Learning}} We utilize the DistributedDataParallel (DDP) tool provided by the flexible package, \textit{PyTorch Ignite}~\cite{pytorch-ignite}, for distributed model learning. In terms of gradient synchronization, the gradients computed by different workers will be averaged following the default DDP gradient synchronization protocol. After that, parameters of the local model will be updated, and all models on different workers will be the same. 

\subsubsection{\textbf{Data Loading}} We use a lightweight 
KV-store to store all graph-related information. 
We choose to use Lightning Memory-Mapped Database (LMDB)
as it allows us to have multiple data loaders simultaneously, where each worker has its own data loader.
This alleviates the system bottleneck that we had when 
using LevelDB for the same purpose, which we found challenging 
to support multi-thread operations.
This design decision turns out significant in reducing the training and inference time.

\subsection{The Explainer of xFraud}\label{sec:gnnexplainer}

\begin{table}[!t]
\centering
\caption{Top$k$ hit rate computed on different explainability methods using various measures (on all 41 communities).} 
\label{tab:centraliy-vs-explainer}
\resizebox{\linewidth}{!}{
\begin{tabular}{llcrrrr}
\toprule
\textbf {} & \textbf{Measures to calculate hit rate}               & \textbf{$H_{Top5}$}  & \textbf{$H_{Top10}$} & \textbf{$H_{Top15}$} & \textbf{$H_{Top20}$} & \textbf{$H_{Top25}$} \\
\midrule
1& edge betweenness                      & \textbf{0.469} & 0.718 & 0.812 & \textbf{0.903} & 0.923 \\
2& edge load                            & 0.455 & 0.707 & 0.812 & 0.902 & 0.923 \\ \hline

3& approximate current flow betweenness & 0.450 & 0.690 & \textbf{0.821} & 0.899 & 0.923 \\
4& betweenness                          & 0.451 & \textbf{0.724} & 0.815 & 0.901 & 0.923 \\
5& closeness                                       & 0.464 & 0.719 & 0.816 & 0.901 & \textbf{0.924} \\
6& communicability betweenness          & 0.448 & 0.688 & 0.812 & 0.899 & 0.922 \\
7& current flow betweenness             & 0.446 & 0.700 & 0.820 & 0.900 & 0.922 \\
8& current flow closeness                & 0.441 & 0.691 & 0.815 & 0.900 & \textbf{0.924} \\
9& degree                                           & 0.464 & 0.716 & 0.815 & 0.901 & \textbf{0.924} \\
10& eigenvector                           & 0.443 & 0.714 & 0.811 & 0.901 & \textbf{0.924} \\
11& harmonic                            & 0.464 & 0.719 & 0.816 & 0.901 & \textbf{0.924} \\
12& load                                 & 0.452 & \textbf{0.724} & 0.815 & 0.901 & 0.923 \\
13& subgraph                              & 0.447 & 0.714 & 0.813 & 0.899 & 0.922 \\
\midrule
14& GNNExplainer weights                               & 0.445 & 0.692 & \textbf{0.821} & 0.898 & 0.921 \\
15& random weights                                  & 0.127 & 0.454 & 0.602 & 0.695 & 0.791 \\
\bottomrule
\end{tabular}}
\end{table}

\begin{figure*}[!t]

    \includegraphics[width=0.7\linewidth]{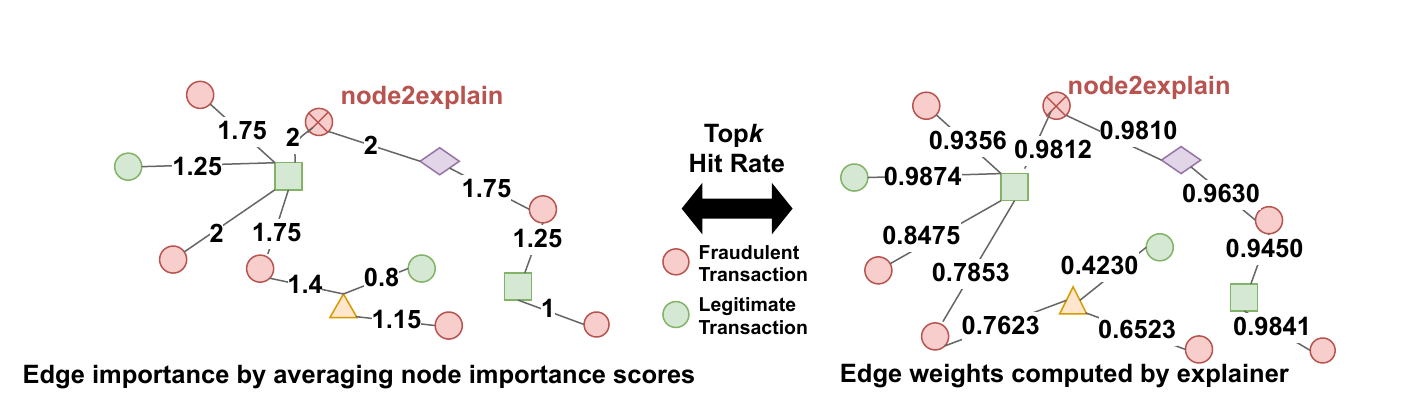}
    \caption{Edge importance scores (left) by human and edge weights by the explainer (right).}
    \label{fig:human-vs-explainer}
\end{figure*}

\begin{figure*}[t!]
    \includegraphics[width=0.75\linewidth]{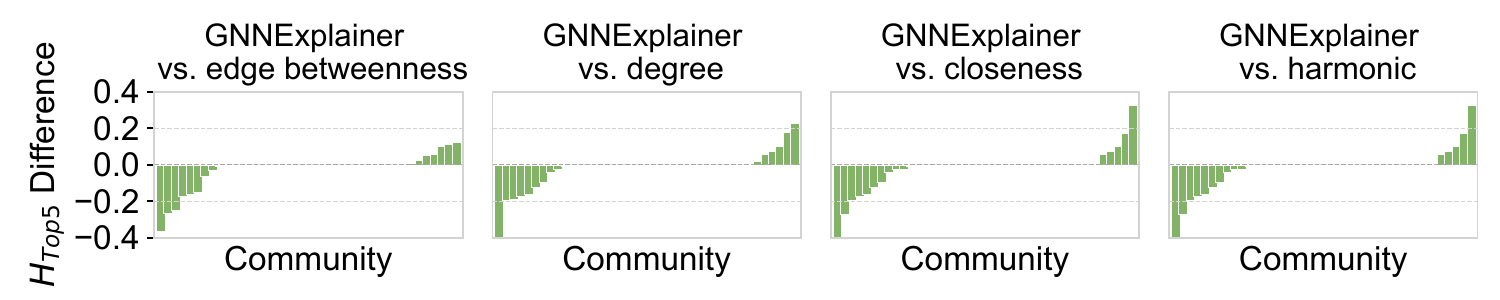}
    \caption{GNNExplainer and centrality measures work well for different communities, forming a trade-off.}
    \label{fig:barplot-diffH}
\end{figure*} 
We present a hybrid explainer (see Figure~\ref{fig:detector-explainer}) in xFraud based on a trade-off between the \textit{task-aware} GNNExplainer and the \textit{task-agnostic} centrality measures.

\subsubsection{\textbf{Trade-off between GNNExplainer and edge centrality}} GNNExplainer \cite{ying2019gnnexplainer} is a model-agnostic explainer which assigns edge weights during node prediction. But,
a fundamental question is --- \textit{Is GNNExplainer itself optimal for all scenarios? What if we replace the edge weights with other measures, such as random weights and edge centrality?} 
To answer this, we conduct a micro benchmark against other measures (see Table~\ref{tab:centraliy-vs-explainer}).
For conducting a fair comparison, we design a metric called Top{\em k} hit rate.

\textbf{Metric: Top\textit{k} hit rate.}
We create human annotations of edge importance and compare the explanation weights with the human annotations. The goal of this quantitative evaluation is to quantify the agreement between different edge importance measures and human annotations. Note that the average edge importance scores and edge weights are in different domains (see Figure~\ref{fig:human-vs-explainer} for an example): the former are discrete values $\in [0,2]$, and the latter continuous numbers $\in [0,1]$. We need a metric that reports the edge ranking of the most important ones in both domains. Hence, we compute the top\textit{k} hit rate $H$, defined as $\frac{\mathbb{N}_{human}^{+,k} \cap \mathbb{N}_{explainer}^{+,k}}{k}$, where $\mathbb{N_{\_}^{+,k}}$ denotes the set of edges ranked by human/explainer as top \textit{k}. Concretely, we count the common edges in their top \textit{k} selection and divide this count by \textit{k}. 

\textbf{Trade-off and Intuition.} Table~\ref{tab:centraliy-vs-explainer} and Figure~\ref{fig:barplot-diffH}
illustrate a trade-off between GNNExplainer and edge centrality measures---they work well on different ``communities'' (test examples) and none of them dominates the other.
GNNExplainer is developed to explain the predictions generated by a GNN network. GNNexplainer computes the importance scores of node features and assigns edge weights, with which we determine the most informative edges when a node prediction is made.
On the contrary, edge centrality measures are popular methods to quantify the edge importance in a network, which is task-agnostic for node prediction. 
Intuitively, we should combine these two measures to generate an explainer which attends to both task-aware and task-agnostic measures.


\subsubsection{\textbf{xFraud Hybrid Explainer}}
Based on the above analysis, we propose a hybrid explainer and
formulate a learning problem as follows.
First, we learn two coefficients, namely, the centrality coefficient $A$ and explainer coefficient $B$ to combine the weights 
from different explanation mechanisms: $A * w(c) + B * w(e)$.
We can learn these two coefficients by either 
Ridge regression or directly maximizing the hit rate on the training set.

\section{Experiments of xFraud Detector+}\label{sec:scalable-xfraud}

We conduct extensive experiments on real-world transaction datasets sampled from the eBay commerce platform to verify the efficacy and efficiency of xFraud detector+. The statistics of the datasets are summarized in Table \ref{tab:data}. 
The details on the graph construction process are in Appendix \ref{app:dataset}~\cite{xfraudAppendix}. We run end-to-end experiments on  \textit{eBay-xlarge} as it is a superset of \textit{eBay-large} and \textit{eBay-small}. 
Specifically, we run the distributed version of xFraud detector+ since \textit{eBay-xlarge} is too large to be fit into a single machine. 
In the graph partition process, we first split the whole graph into $128$ subgraphs using PIC. 
We then organize these $128$ subgraphs into $\kappa$ groups, where $\kappa$ is the number of workers\footnote{{We first order the $128$ subgraphs according to the total number of nodes in ascending order. Then, we put the first few subgraphs that cumulatively have $\lceil \frac{|\mathcal{V}|}{\kappa} \rceil$ nodes into the same group. We repeat this process until we get $\kappa$ groups. In this way, we ensure that each machine receives a graph partition of similar total number of nodes.}}. Different groups are handled by different workers.
After the end-to-end evaluation (Sec.~\ref{sec:end2end}), we conducted an ablation study (Sec.~\ref{sec:ablation}) on \textit{eBay-large} and \textit{eBay-small} to study the trade-off between xFraud detector (i.e., HGT) and xFraud detector+.

\begin{figure*}[!t]
    \centering
    \includegraphics[width=1\textwidth]{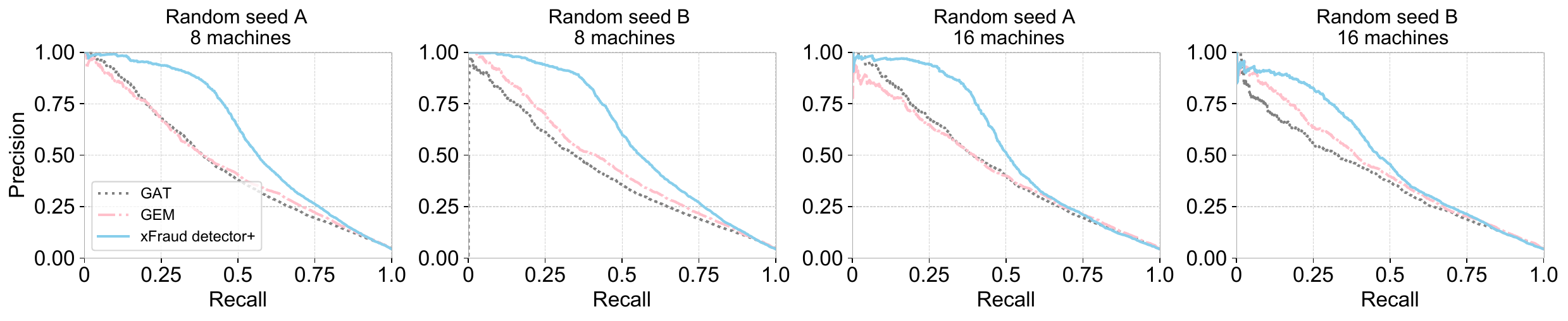}
    \caption{{Precision-recall curves using different settings (seeds and \# machines) on \textit{eBay-xlarge}.}}
    \label{fig:pr-curve-xlarge}
\end{figure*}


\begin{figure*}[!t]
\includegraphics[width=1\textwidth]{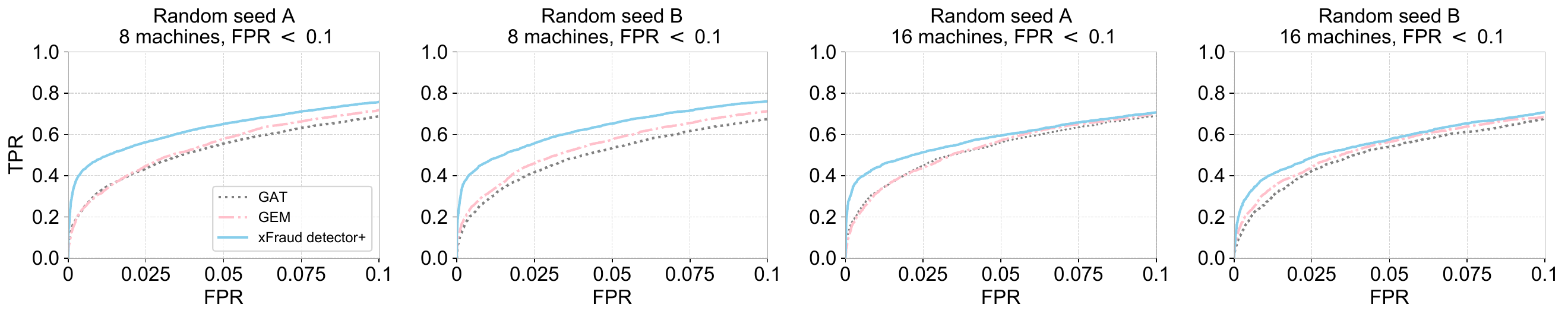}
\caption{{ROC curves using different settings (seeds and \# machines) on \textit{eBay-xlarge} for FPR < 0.1.}}
    \label{fig:roc-curve-xlarge-0.1}
\end{figure*}

\begin{figure}[!t]
\centering
\includegraphics[width=\linewidth]{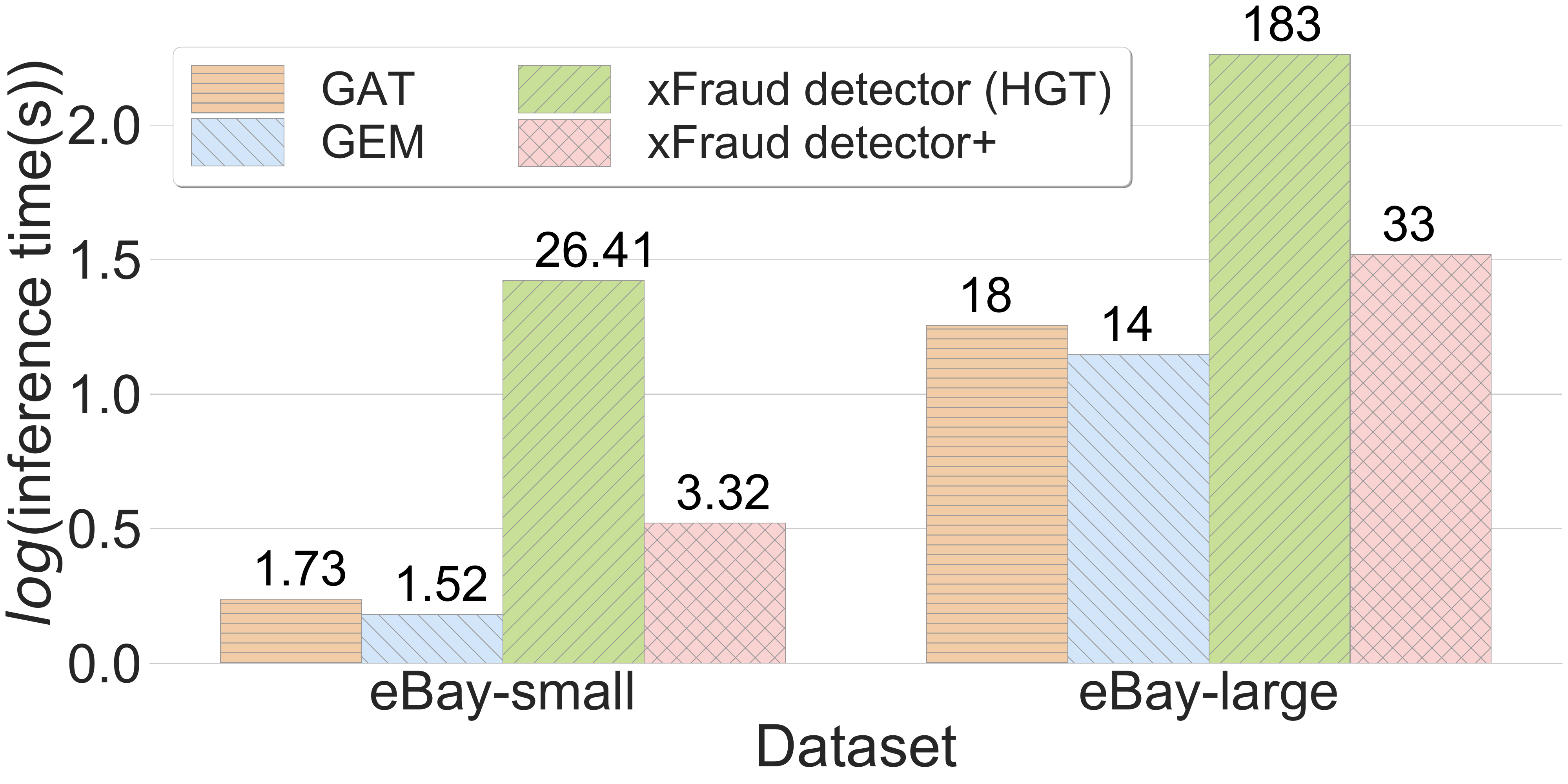}
\caption{{Total inference time (in $log$) on the test set of \textit{eBay-small} and \textit{eBay-large} (actual time in parentheses).}}
\label{fig:inference-time}
\end{figure}

{\subsection{End-to-end Experiments}\label{sec:end2end}}


{We report the end-to-end results on \textit{eBay-xlarge} in Table~\ref{tab:xlarge-detector-result-avg}.
In addition, we show precision-recall curve and ROC curve in Figure~\ref{fig:pr-curve-xlarge} and Figure~\ref{fig:roc-curve-xlarge-0.1} to further study
prediction performance as \textit{eBay-xlarge} is an extremely imbalanced graph dataset.}

{
{\bf End-to-end Results.}
From Table~\ref{tab:xlarge-detector-result-avg}, our detector+, achieves the best AUC (averaged across seeds)  using 8 machines w.r.t. GEM and GAT.
In terms of training efficiency, xFraud detector+ takes only slightly longer time per epoch compared to GEM
in an 8-machine setting. 
If we increase the number of machines to 16, the training time is reduced by 1.89$\times$,
1.84$\times$ and 1.82$\times$ for GAT, GEM, and our detector+, respectively.
Despite roughly linear speedups, we observe lowered AUC compared with 8 machines.
In our implementation, each machine only has a subgraph; therefore,
all three methods obtain a suboptimal model due to a
restrained field of neighbors and edges.
This phenomenon reveals a trade-off about our current way of  handling large-scale graphs ---
one can use more resources to accelerate the model training but might have to compromise the model performance.
It is an interesting future work to understand how to 
develop better distributed algorithms for training heterogeneous graph models.
For more details on the system implementation and results, see Appendix~\ref{app:distributed}~\cite{xfraudAppendix}.
GEM (8 machines) takes the shortest time to do inference over a batch of 640 nodes due to the simplicity of its convolution layers. GAT and xFraud have longer inference time because their implementations of attention mechanisms.
xFraud takes slightly longer than GAT due to its attention on heterogeneous types of nodes and edges. Since all methods take less than 0.1 second for a batch, all of them are practical to be deployed in production. 
Overall, xFraud is appealing in fraud detection,
as it achieves the best model quality with a reasonably fast inference speed.} {
Since xFraud detector+ is scalable to 16 machines, an online production scenario using xFraud can leverage historical and up-to-date transaction records to incrementally train a detector (Appendix~\ref{app:production}~\cite{xfraudAppendix}).}

\begin{table}[!t]
\centering
\caption{\centering Dataset summary (``B":billion;``M":million;``K":thousand) \hspace{10em} \small {*The ratio of frauds is only reported on the sampled datasets.}}
\label{tab:data}
\resizebox{\linewidth}{!}{
\begin{tabular}{cccrrc}
\toprule
\textbf{Dataset}            & \textbf{Features} & \textbf{Graph type} & \multicolumn{1}{c}{\textbf{\#Nodes}} & \multicolumn{1}{c}{\textbf{\#Edges}} & \textbf{Fraud\%*}       \\
\midrule
\textit{eBay-xlarge}                 & 480               & hetero              & 1.1B           & 3.7B            & 4.33\% \\  
\hline
\textit{eBay-small} & 114              &            hetero     & 289K                              & 613K                          & 4.30\% \\ 
                            
\textit{eBay-large}                  & 480               & hetero              & 8.9M                           & 13.2M                           & 3.57\%   \\
\bottomrule
\end{tabular}}
\end{table}

\begin{table}[!t]
\centering
\caption{End-to-end performance on the dataset \textit{eBay-xlarge} (epochs: 128). We report the \textbf{average} scores over two different seeds (A and B). }
\label{tab:xlarge-detector-result-avg}
\resizebox{\linewidth}{!}{
\begin{tabular}{ccccc}
\toprule
\textbf{\# machines} & \textbf{Model} & \textbf{AUC} & \begin{tabular}[c]{@{}c@{}}\textbf{Training time} \\

\textbf{(s/epoch)}\end{tabular} & \begin{tabular}[c]{@{}c@{}}\textbf{Inference time} \\

\textbf{(s/batch)}\end{tabular}  \\ 
\midrule

\multirow{3}{*}{8}      & GAT   & 0.8879       & 62.74                            & 0.0557 $\pm$ 0.1966                        \\ 
                        & GEM     & 0.8961       & \textbf{61.77}                            & \textbf{0.0167 $\pm$ 0.0054 }                       \\
                        & xFraud detector+ & \textbf{0.9074}       & 70.47                            & 0.0799 $\pm$ 0.1868                        \\ \hline
\multirow{3}{*}{16}     & GAT  & 0.8866       & \textbf{33.11} (1.89$\times$)                            & 0.0557 $\pm$ 0.1966                         \\ 
                        & GEM & \textbf{0.8938}       & 33.56 (1.84$\times$)                           & \textbf{0.0167 $\pm$ 0.0054}                        \\
                        & xFraud detector+  & 0.8892       & 38.72 (1.82$\times$)                            & 0.0799 $\pm$ 0.1868          \\ \bottomrule               
\end{tabular}}
\end{table}

{
{\bf Precision-recall Curve (P/R Curve).}}
{Figure~\ref{fig:pr-curve-xlarge} illustrates the P/R Curve using different settings. 
The trade-off of precision and recall is an ever-lasting goal for machine learning models; and
xFraud detector+ achieves a better balance between precision and recall compared to GAT and GEM, which means that our model can return more accurate results (higher precision) as well as most of the true fraudulent transactions being found (higher recall).}

{
{\bf ROC Curve.}
In \textit{eBay-xlarge}, the majority of transactions are benign and the ratio of fraud transactions is very low.
Besides, from the application perspective, the task of fraud detection is vulnerable to
false positive cases, when benign transactions are flagged as fraud. This would cause an overwhelming human verification and significantly worsen user experience.
Therefore, it is important to study the imbalance-aware metrics like true positive rate (FPR) and false positive rate (TPR).
Specifically, if we restrict the FPR being lower than 0.1 as in Figure~\ref{fig:roc-curve-xlarge-0.1} (even this small ratio could involve 85.7M transactions in \textit{eBay-xlarge}), xFraud significantly outperforms GAT and GEM when only a small FPR is allowed. We plot the full range of FPR in Figure~\ref{fig:roc-curve-xlarge} of Appendix~\ref{app:tpr-tnr-fpr-fnr}~\cite{xfraudAppendix}, where the three models have a similar area under ROC curve (i.e., AUC-ROC), xFraud's ROC curve is consistently beyond GAT and GEM. 
}

{\bf Discussion.} All results we report are on a dataset
after pre-filtering the fraudulent/benign transactions
with rule/ML-based filters and down-sampling 
the benign transactions. In 
Appendix~\ref{app:more-metrics-eval}~\cite{xfraudAppendix}
we discuss the implication of this pre-filtering step
and the production scenario of xFraud.
Even without downsampling benign transactions, xFraud achieves a reasonable 
precision and recall on industrial-scale data: from 3 fraud candidates investigated by the business unit, 1 will be a real fraud, with 0.1 of recall.

{\subsection{Ablation Study: xFraud detector+ vs. xFraud detector (i.e., HGT)} \label{sec:ablation} }

Here, we conduct an ablation study of xFraud detector (i.e., HGT) and xFraud detector+ to demonstrate the efficiency of the sampler. We run experiments on \textit{eBay-small} and \textit{eBay-large}, which are subsets of \textit{eBay-xlarge} because our \textit{eBay-xlarge} is too large such that xFraud detector (i.e., HGT) can no longer handle it. We report the inference time and AUC using a single machine in Figure~\ref{fig:inference-time}. Concretely, xFraud detector+ achieves a 5 $\times$ speedups in terms of inference time (during testing) on \textit{eBay-large} compared with xFraud detector (i.e., HGT), and the speedup on \textit{eBay-small} is even larger, i.e., up to 7 $\times$. 
Meanwhile, using a simplified yet efficient sampler (GraphSAGE) will not sacrifice the model AUC (\textit{eBay-small} vs. \textit{eBay-large}): 0.7248 vs. 0.8683 for HGT, and 0.7262 vs. 0.8690 for xFraud detector+.
Interestingly, we observe that the xFraud detector+ can even slightly outperform xFraud detector (i.e., HGT) on both datasets in terms of AUC.

\section{Experiments of xFraud Explainer}\label{sec:explainer-eval}
In this section, we discuss how we build an xFraud explainer on top of the detector. The main contribution of the explainer is to compute node features and edge masks of important features and nodes. As we see in the quantitative analysis (Sec.~\ref{sec:explainer-eval-quantitative}) and the case studies (Sec.~\ref{sec:explainer-eval-qualitative}), xFraud explainer ranks important edges with high agreement with expert human annotators and provides interesting insights for risk experts when analyzing risk propagation in a local heterogeneous graph. The output of the explainer carries the following meaning: node feature masks give high weights to the node feature dimensions influential in prediction; edge masks are the weights of edges in the subgraph, which indicate the strength of connectedness between pairs of nodes when flagging fraud. We visualize the subgraph structure with these outputs.

\subsection{Quantitative Analysis of xFraud Explainer: Top\textit{k} Hit Rate}
\label{sec:explainer-eval-quantitative}
First, we want to quantify the efficacy of xFraud explainer by studying the agreement between human perception and explainer output. Our approach is to compute the agreement between the edge importance scores based on human annotations and the edge weights generated by our {hybrid} explainer. 

\textbf{Sample}. In total, we randomly select 41 communities from our test set, among which 18 communities has a fraudulent transaction as seed (label 1), 23 a legitimate transaction as seed (label 0). The AUC score of this test sample is 81.88\%. A community is formed around a transaction seed node, where all connected nodes and edges are taken. In total, we have 1,591 nodes of five types (buyer, transaction, shipping address, email, and payment token), and 3,344 edges.\footnote{Note that explainer assumes directions and assigns two weights to bidirectional edges connecting a pair of nodes. Since human annotation is on the node level, and it is generally hard for annotators to consider directions, we remove directions in the explainer weights by taking the larger weight.} On average, there are 81.56 edges per community. 

\textbf{{Human annotations and edge importance score.}} {We have created the human evaluation of edge importance scores of all edges in these 41 communities. The annotation protocol and score calculations are listed in Appendix~\ref{app:human-annotation}~\cite{xfraudAppendix}. }
{Out of 41 communities, we take the first 21 communities as the training set, the last 20 as the test set. We trained two versions
of the hybrid explainer: (1) via ridge regression on the 
human annotations on the training set, and (2) via directly optimizing the 
average hit rate on the training set.}

\textbf{{Results.}}
{Table~\ref{tab:hybrid-explainer}
illustrates the result. We see that the hybrid explainer 
consistently outperforms both GNNExplainer and centrality measures.
This is not surprising, as shown in our previous discussion
of the tradeoff. It is an exciting future direction to 
come up with better ways to combine these different metrics together 
to form an even better explainer for graphs.
}

    

\begin{table}[!t]
\centering
\caption{{Top$k$ hit rate in the test communities. } }
\label{tab:hybrid-explainer}
\resizebox{\linewidth}{!}{
\begin{tabular}{lcccc}
\hline
\textbf{H(\_)} & 
\multicolumn{1}{c}{\begin{tabular}[c]{@{}c@{}}Edge \\ betweenness $H(c)$\end{tabular}}
 & 
 \multicolumn{1}{c}{\begin{tabular}[c]{@{}c@{}}GNNExplainer \\  $H(e)$\end{tabular}} &
 \multicolumn{1}{c}{\begin{tabular}[c]{@{}c@{}}Hybrid \\  (ridge) $H(h)$\end{tabular}} &  \multicolumn{1}{c}{\begin{tabular}[c]{@{}c@{}}Hybrid \\  (grid) $H(h)$\end{tabular}} \\ \hline
Top5           & 0.45540                                              & 0.44800                      & 0.44890                                            & \textbf{0.45550}              \\
Top10          & 0.78175                                              & 0.77580                      & \textbf{0.81115}                                   & 0.78700                       \\
Top15          & 0.87763                                              & 0.88473                      & 0.88963                                            & \textbf{0.89410}              \\
Top20          & 0.96205                                              & 0.95840                      & 0.96198                                            & \textbf{0.96275}              \\
Top25          & \textbf{0.96616}                                     & 0.95954                      & 0.96614                                            & 0.96614                       \\ \hline
\end{tabular}}
\end{table}

\begin{figure}[!t]

    \centering
\includegraphics[width=1\linewidth]{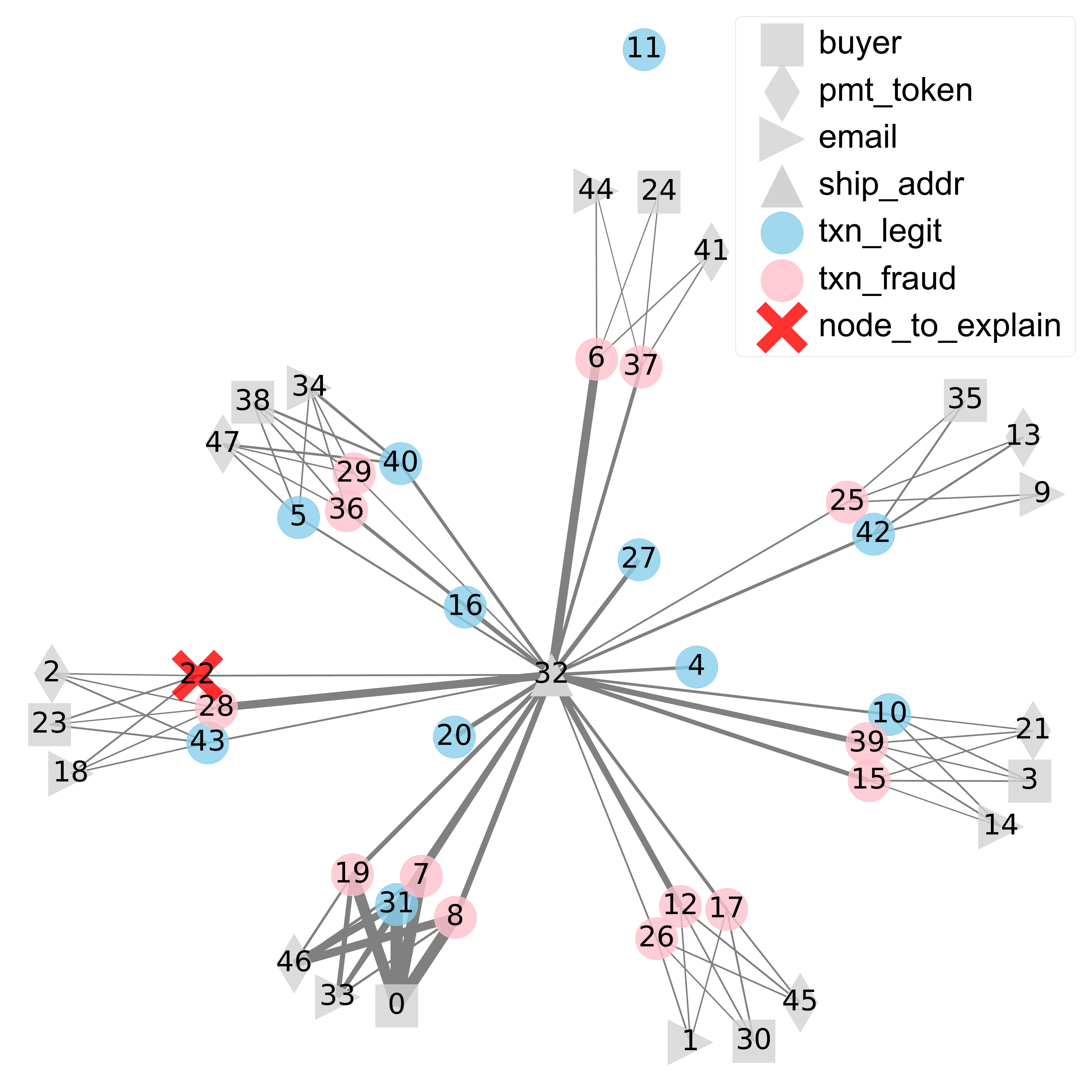}
\caption{{TP: xFraud helps to catch potential frauds.}}
\label{fig:tp-case}
\end{figure}

\subsection{Qualitative Analysis of xFraud Explainer: Case Studies}\label{sec:explainer-eval-qualitative}
\label{sec:explainer-eval-qualitative}

To visualize the subgraph for a certain node, we use the node index, edge indices and their masks, true labels of nodes as input. The thicker an edge is, the stronger the connection is. We visualize the connections with nondirectional edges and use the ground truth labels for transactions. 
{\textbf{True positive (TP): flagging frauds.} In Figure~\ref{fig:tp-case}, we see a generic shipping address (node 32, a warehouse) connected to both fraudulent/benign transactions related to various buyers using various payment tokens/emails. According to BU, one explanation for this pattern is there is often a lag between user chargebacks and when the frauds have taken place, not to mention it is possible that a card stolen claim might never be forwarded to eBay from some banks\footnote{This is out of control of eBay and is due to the inconsistency of reporting systems at some banks.}. As a result, we cannot fully trust the positive labels in such a case, where it is clearly unusual for such a community to have an extremely mixed benign/transactions across buyers. And it could also be a case where defaulters disguise their true purposes by "cultivating" some legitimate accounts to execute a few legit transactions. For the $2^{nd}$ assumption, the BU needs extra evidence to further examine.}  {Either way, xFraud is able to flag the node-to-predict as fraudulent by learning from the important edges (the thicker ones), and to inform the BU that this set of buyers are highly suspicious and should be under more detailed examinations. This shows the importance in detecting frauds on the transaction level as we propose in xFraud, instead of just on the account level as in GEM~\cite{liu2018heterogeneous}. } Currently, the BU is only using a rule based system\footnote{\textit{skope-rules} which perform rule mining on tabular data, \url{https://github.com/scikit-learn-contrib/skope-rules} (last accessed: Oct 18, 2020).} to filter the suspicious transactions stored in the tabular format. xFraud explainer is innovative to a traditional BU annotation routine, because it allows experts to combine graph level and feature level information.
{For extensive case studies on \textbf{false positive (FP): benign $\rightarrow$ fraud} and \textbf{false negative (FN): fraud $\rightarrow$ benign}, we discuss in Appendix~\ref{app:more-cases-hybridexplainer}~\cite{xfraudAppendix}, where we also discuss system limitations and potential solutions to improve xFraud.}
\section{Conclusion}\label{sec:conclusion}
In this paper, we propose xFraud, a system for detecting fraud transaction and explaining model prediction. Specifically, a heterogeneous graph is constructed and a self-attentive heterogeneous graph neural network is leveraged for risky transaction scoring. 
{We further design a learnable hybrid explainer that leverages both GNNExplainer and centrality measures to learn node- and edge-level explanations simultaneously.} 
{Through extensive experiments on real-world datasets, we show 
the proposed xFraud detector+ can efficiently process billion-scale heterogeneous graphs and
outperform the competitive baselines.}
More importantly, xFraud is the first work that quantifies a strong agreement between human perception and explainer outputs. Real-world case studies illustrate that with the hybrid xFraud explainer, we can generate convincing explanations to assist further decision-making of business units. 


\begin{acks}

Ce Zhang and the DS3Lab gratefully acknowledge the support from Swiss National Science Foundation (Project Number 200021\_184628 and 197485), Innosuisse/SNF BRIDGE Discovery (Project Number 40B2-0\_187132), European Union Horizon 2020 Research and Innovation Programme (DAPHNE, 957407), Botnar Research Centre for Child Health, Swiss Data Science Center, Alibaba, Cisco, eBay, Google Focused Research Awards, Kuaishou Inc., Oracle Labs, Zurich Insurance, and the Department of Computer Science at ETH Zurich. We also thank the anonymous reviewers for their constructive feedback and the PC chair’s coordination throughout the reviewing process. Special thanks go to  \href{https://scholar.google.com/citations?user=G_Hg-j0AAAAJ&hl=en}{Jiang Jiawei}, who is always generous in providing thoughtful comments and making efforts towards improving our manuscript. 
\end{acks}

\balance
\bibliographystyle{ACM-Reference-Format}
\bibliography{sample}


\begin{thebibliography}{54}


\ifx \showCODEN    \undefined \def \showCODEN     #1{\unskip}     \fi
\ifx \showDOI      \undefined \def \showDOI       #1{#1}\fi
\ifx \showISBNx    \undefined \def \showISBNx     #1{\unskip}     \fi
\ifx \showISBNxiii \undefined \def \showISBNxiii  #1{\unskip}     \fi
\ifx \showISSN     \undefined \def \showISSN      #1{\unskip}     \fi
\ifx \showLCCN     \undefined \def \showLCCN      #1{\unskip}     \fi
\ifx \shownote     \undefined \def \shownote      #1{#1}          \fi
\ifx \showarticletitle \undefined \def \showarticletitle #1{#1}   \fi
\ifx \showURL      \undefined \def \showURL       {\relax}        \fi
\providecommand\bibfield[2]{#2}
\providecommand\bibinfo[2]{#2}
\providecommand\natexlab[1]{#1}
\providecommand\showeprint[2][]{arXiv:#2}

\bibitem[\protect\citeauthoryear{Baesens, Van~Vlasselaer, and Verbeke}{Baesens
  et~al\mbox{.}}{2015}]%
        {baesens2015fraud}
\bibfield{author}{\bibinfo{person}{Bart Baesens}, \bibinfo{person}{Veronique
  Van~Vlasselaer}, {and} \bibinfo{person}{Wouter Verbeke}.}
  \bibinfo{year}{2015}\natexlab{}.
\newblock \bibinfo{booktitle}{\emph{Fraud analytics using descriptive,
  predictive, and social network techniques: a guide to data science for fraud
  detection}}.
\newblock \bibinfo{publisher}{John Wiley \& Sons}.
\newblock


\bibitem[\protect\citeauthoryear{Baldassarre and Azizpour}{Baldassarre and
  Azizpour}{2019}]%
        {baldassarre2019explainability}
\bibfield{author}{\bibinfo{person}{Federico Baldassarre} {and}
  \bibinfo{person}{Hossein Azizpour}.} \bibinfo{year}{2019}\natexlab{}.
\newblock \showarticletitle{Explainability techniques for graph convolutional
  networks}.
\newblock \bibinfo{journal}{\emph{arXiv preprint arXiv:1905.13686}}
  (\bibinfo{year}{2019}).
\newblock


\bibitem[\protect\citeauthoryear{Breuer, Eilat, and Weinsberg}{Breuer
  et~al\mbox{.}}{2020}]%
        {breuer2020friend}
\bibfield{author}{\bibinfo{person}{Adam Breuer}, \bibinfo{person}{Roee Eilat},
  {and} \bibinfo{person}{Udi Weinsberg}.} \bibinfo{year}{2020}\natexlab{}.
\newblock \showarticletitle{Friend or Faux: Graph-Based Early Detection of Fake
  Accounts on Social Networks}. In \bibinfo{booktitle}{\emph{Proceedings of The
  Web Conference 2020}}. \bibinfo{pages}{1287--1297}.
\newblock


\bibitem[\protect\citeauthoryear{Cao, Mao, Viidu, and Philip}{Cao
  et~al\mbox{.}}{2017}]%
        {cao2017hitfraud}
\bibfield{author}{\bibinfo{person}{Bokai Cao}, \bibinfo{person}{Mia Mao},
  \bibinfo{person}{Siim Viidu}, {and} \bibinfo{person}{S~Yu Philip}.}
  \bibinfo{year}{2017}\natexlab{}.
\newblock \showarticletitle{HitFraud: a broad learning approach for collective
  fraud detection in heterogeneous information networks}. In
  \bibinfo{booktitle}{\emph{2017 IEEE international conference on data mining
  (ICDM)}}. IEEE, \bibinfo{pages}{769--774}.
\newblock


\bibitem[\protect\citeauthoryear{Cao, Yang, Chen, Zhou, Li, and Qi}{Cao
  et~al\mbox{.}}{2019}]%
        {cao2019titant}
\bibfield{author}{\bibinfo{person}{Shaosheng Cao}, \bibinfo{person}{XinXing
  Yang}, \bibinfo{person}{Cen Chen}, \bibinfo{person}{Jun Zhou},
  \bibinfo{person}{Xiaolong Li}, {and} \bibinfo{person}{Yuan Qi}.}
  \bibinfo{year}{2019}\natexlab{}.
\newblock \showarticletitle{TitAnt: online real-time transaction fraud
  detection in Ant Financial}.
\newblock \bibinfo{journal}{\emph{Proceedings of the VLDB Endowment}}
  (\bibinfo{year}{2019}).
\newblock


\bibitem[\protect\citeauthoryear{Cen, Zou, Zhang, Yang, Zhou, and Tang}{Cen
  et~al\mbox{.}}{2019}]%
        {cen2019representation}
\bibfield{author}{\bibinfo{person}{Yukuo Cen}, \bibinfo{person}{Xu Zou},
  \bibinfo{person}{Jianwei Zhang}, \bibinfo{person}{Hongxia Yang},
  \bibinfo{person}{Jingren Zhou}, {and} \bibinfo{person}{Jie Tang}.}
  \bibinfo{year}{2019}\natexlab{}.
\newblock \showarticletitle{Representation learning for attributed multiplex
  heterogeneous network}. In \bibinfo{booktitle}{\emph{Proceedings of the 25th
  ACM SIGKDD International Conference on Knowledge Discovery \& Data Mining}}.
  \bibinfo{pages}{1358--1368}.
\newblock


\bibitem[\protect\citeauthoryear{Chang, Han, Tang, Qi, Aggarwal, and
  Huang}{Chang et~al\mbox{.}}{2015}]%
        {chang2015heterogeneous}
\bibfield{author}{\bibinfo{person}{Shiyu Chang}, \bibinfo{person}{Wei Han},
  \bibinfo{person}{Jiliang Tang}, \bibinfo{person}{Guo-Jun Qi},
  \bibinfo{person}{Charu~C Aggarwal}, {and} \bibinfo{person}{Thomas~S Huang}.}
  \bibinfo{year}{2015}\natexlab{}.
\newblock \showarticletitle{Heterogeneous network embedding via deep
  architectures}. In \bibinfo{booktitle}{\emph{Proceedings of the 21th ACM
  SIGKDD international conference on knowledge discovery and data mining}}.
  \bibinfo{pages}{119--128}.
\newblock


\bibitem[\protect\citeauthoryear{Dhawan, Gangireddy, Kumar, and
  Chakraborty}{Dhawan et~al\mbox{.}}{2019}]%
        {dhawan2019spotting}
\bibfield{author}{\bibinfo{person}{Sarthika Dhawan}, \bibinfo{person}{Siva
  Charan~Reddy Gangireddy}, \bibinfo{person}{Shiv Kumar}, {and}
  \bibinfo{person}{Tanmoy Chakraborty}.} \bibinfo{year}{2019}\natexlab{}.
\newblock \showarticletitle{Spotting collective behaviour of online frauds in
  customer reviews}.
\newblock \bibinfo{journal}{\emph{arXiv preprint arXiv:1905.13649}}
  (\bibinfo{year}{2019}).
\newblock


\bibitem[\protect\citeauthoryear{Dong, Chawla, and Swami}{Dong
  et~al\mbox{.}}{2017}]%
        {dong2017metapath2vec}
\bibfield{author}{\bibinfo{person}{Yuxiao Dong}, \bibinfo{person}{Nitesh~V
  Chawla}, {and} \bibinfo{person}{Ananthram Swami}.}
  \bibinfo{year}{2017}\natexlab{}.
\newblock \showarticletitle{metapath2vec: Scalable representation learning for
  heterogeneous networks}. In \bibinfo{booktitle}{\emph{Proceedings of the 23rd
  ACM SIGKDD international conference on knowledge discovery and data mining}}.
  \bibinfo{pages}{135--144}.
\newblock


\bibitem[\protect\citeauthoryear{Emmerich, Pudelko, Gallenm{\"u}ller, and
  Carle}{Emmerich et~al\mbox{.}}{2017}]%
        {emmerich2017flowscope}
\bibfield{author}{\bibinfo{person}{Paul Emmerich}, \bibinfo{person}{Maximilian
  Pudelko}, \bibinfo{person}{Sebastian Gallenm{\"u}ller}, {and}
  \bibinfo{person}{Georg Carle}.} \bibinfo{year}{2017}\natexlab{}.
\newblock \showarticletitle{FlowScope: Efficient packet capture and storage in
  100 Gbit/s networks}. In \bibinfo{booktitle}{\emph{2017 IFIP Networking
  Conference (IFIP Networking) and Workshops}}. IEEE, \bibinfo{pages}{1--9}.
\newblock


\bibitem[\protect\citeauthoryear{Eswaran, G{\"u}nnemann, Faloutsos, Makhija,
  and Kumar}{Eswaran et~al\mbox{.}}{2017}]%
        {eswaran2017zoobp}
\bibfield{author}{\bibinfo{person}{Dhivya Eswaran}, \bibinfo{person}{Stephan
  G{\"u}nnemann}, \bibinfo{person}{Christos Faloutsos}, \bibinfo{person}{Disha
  Makhija}, {and} \bibinfo{person}{Mohit Kumar}.}
  \bibinfo{year}{2017}\natexlab{}.
\newblock \showarticletitle{Zoobp: Belief propagation for heterogeneous
  networks}.
\newblock \bibinfo{journal}{\emph{Proceedings of the VLDB Endowment}}
  \bibinfo{volume}{10}, \bibinfo{number}{5} (\bibinfo{year}{2017}),
  \bibinfo{pages}{625--636}.
\newblock


\bibitem[\protect\citeauthoryear{Fomin, Anmol, Desroziers, Kriss, and
  Tejani}{Fomin et~al\mbox{.}}{2020}]%
        {pytorch-ignite}
\bibfield{author}{\bibinfo{person}{V. Fomin}, \bibinfo{person}{J. Anmol},
  \bibinfo{person}{S. Desroziers}, \bibinfo{person}{J. Kriss}, {and}
  \bibinfo{person}{A. Tejani}.} \bibinfo{year}{2020}\natexlab{}.
\newblock \bibinfo{title}{High-level library to help with training neural
  networks in PyTorch}.
\newblock \bibinfo{howpublished}{\url{https://github.com/pytorch/ignite}}.
\newblock


\bibitem[\protect\citeauthoryear{Fu, Zhang, Meng, and King}{Fu
  et~al\mbox{.}}{2020}]%
        {fu2020magnn}
\bibfield{author}{\bibinfo{person}{Xinyu Fu}, \bibinfo{person}{Jiani Zhang},
  \bibinfo{person}{Ziqiao Meng}, {and} \bibinfo{person}{Irwin King}.}
  \bibinfo{year}{2020}\natexlab{}.
\newblock \showarticletitle{Magnn: Metapath aggregated graph neural network for
  heterogeneous graph embedding}. In \bibinfo{booktitle}{\emph{Proceedings of
  The Web Conference 2020}}. \bibinfo{pages}{2331--2341}.
\newblock


\bibitem[\protect\citeauthoryear{Hamilton, Ying, and Leskovec}{Hamilton
  et~al\mbox{.}}{2017}]%
        {hamilton2017inductive-sage}
\bibfield{author}{\bibinfo{person}{Will Hamilton}, \bibinfo{person}{Zhitao
  Ying}, {and} \bibinfo{person}{Jure Leskovec}.}
  \bibinfo{year}{2017}\natexlab{}.
\newblock \showarticletitle{Inductive representation learning on large graphs}.
  In \bibinfo{booktitle}{\emph{Advances in neural information processing
  systems}}. \bibinfo{pages}{1024--1034}.
\newblock


\bibitem[\protect\citeauthoryear{Hooi, Song, Beutel, Shah, Shin, and
  Faloutsos}{Hooi et~al\mbox{.}}{2016}]%
        {hooi2016fraudar}
\bibfield{author}{\bibinfo{person}{Bryan Hooi}, \bibinfo{person}{Hyun~Ah Song},
  \bibinfo{person}{Alex Beutel}, \bibinfo{person}{Neil Shah},
  \bibinfo{person}{Kijung Shin}, {and} \bibinfo{person}{Christos Faloutsos}.}
  \bibinfo{year}{2016}\natexlab{}.
\newblock \showarticletitle{Fraudar: Bounding graph fraud in the face of
  camouflage}. In \bibinfo{booktitle}{\emph{Proceedings of the 22nd ACM SIGKDD
  International Conference on Knowledge Discovery and Data Mining}}.
  \bibinfo{pages}{895--904}.
\newblock


\bibitem[\protect\citeauthoryear{Hu, Fang, and Shi}{Hu et~al\mbox{.}}{2019a}]%
        {hu2019adversarial}
\bibfield{author}{\bibinfo{person}{Binbin Hu}, \bibinfo{person}{Yuan Fang},
  {and} \bibinfo{person}{Chuan Shi}.} \bibinfo{year}{2019}\natexlab{a}.
\newblock \showarticletitle{Adversarial learning on heterogeneous information
  networks}. In \bibinfo{booktitle}{\emph{Proceedings of the 25th ACM SIGKDD
  International Conference on Knowledge Discovery \& Data Mining}}.
  \bibinfo{pages}{120--129}.
\newblock


\bibitem[\protect\citeauthoryear{Hu, Zhang, Shi, Zhou, Li, and Qi}{Hu
  et~al\mbox{.}}{2019b}]%
        {hu2019cash}
\bibfield{author}{\bibinfo{person}{Binbin Hu}, \bibinfo{person}{Zhiqiang
  Zhang}, \bibinfo{person}{Chuan Shi}, \bibinfo{person}{Jun Zhou},
  \bibinfo{person}{Xiaolong Li}, {and} \bibinfo{person}{Yuan Qi}.}
  \bibinfo{year}{2019}\natexlab{b}.
\newblock \showarticletitle{Cash-out user detection based on attributed
  heterogeneous information network with a hierarchical attention mechanism}.
  In \bibinfo{booktitle}{\emph{Proceedings of the AAAI Conference on Artificial
  Intelligence}}, Vol.~\bibinfo{volume}{33}. \bibinfo{pages}{946--953}.
\newblock


\bibitem[\protect\citeauthoryear{Hu, Dong, Wang, and Sun}{Hu
  et~al\mbox{.}}{2020}]%
        {hu2020hgt}
\bibfield{author}{\bibinfo{person}{Ziniu Hu}, \bibinfo{person}{Yuxiao Dong},
  \bibinfo{person}{Kuansan Wang}, {and} \bibinfo{person}{Yizhou Sun}.}
  \bibinfo{year}{2020}\natexlab{}.
\newblock \showarticletitle{Heterogeneous graph transformer}. In
  \bibinfo{booktitle}{\emph{Proceedings of The Web Conference 2020}}.
  \bibinfo{pages}{2704--2710}.
\newblock


\bibitem[\protect\citeauthoryear{Huang, Yamada, Tian, Singh, Yin, and
  Chang}{Huang et~al\mbox{.}}{2020}]%
        {huang2020graphlime}
\bibfield{author}{\bibinfo{person}{Qiang Huang}, \bibinfo{person}{Makoto
  Yamada}, \bibinfo{person}{Yuan Tian}, \bibinfo{person}{Dinesh Singh},
  \bibinfo{person}{Dawei Yin}, {and} \bibinfo{person}{Yi Chang}.}
  \bibinfo{year}{2020}\natexlab{}.
\newblock \showarticletitle{GraphLIME: Local interpretable model explanations
  for graph neural networks}.
\newblock \bibinfo{journal}{\emph{arXiv preprint arXiv:2001.06216}}
  (\bibinfo{year}{2020}).
\newblock


\bibitem[\protect\citeauthoryear{Kaghazgaran, Caverlee, and
  Squicciarini}{Kaghazgaran et~al\mbox{.}}{2018}]%
        {kaghazgaran2018combating}
\bibfield{author}{\bibinfo{person}{Parisa Kaghazgaran}, \bibinfo{person}{James
  Caverlee}, {and} \bibinfo{person}{Anna Squicciarini}.}
  \bibinfo{year}{2018}\natexlab{}.
\newblock \showarticletitle{Combating crowdsourced review manipulators: A
  neighborhood-based approach}. In \bibinfo{booktitle}{\emph{Proceedings of the
  Eleventh ACM International Conference on Web Search and Data Mining}}.
  \bibinfo{pages}{306--314}.
\newblock


\bibitem[\protect\citeauthoryear{Kumar, Hooi, Makhija, Kumar, Faloutsos, and
  Subrahmanian}{Kumar et~al\mbox{.}}{2018}]%
        {kumar2018rev2}
\bibfield{author}{\bibinfo{person}{Srijan Kumar}, \bibinfo{person}{Bryan Hooi},
  \bibinfo{person}{Disha Makhija}, \bibinfo{person}{Mohit Kumar},
  \bibinfo{person}{Christos Faloutsos}, {and} \bibinfo{person}{VS
  Subrahmanian}.} \bibinfo{year}{2018}\natexlab{}.
\newblock \showarticletitle{Rev2: Fraudulent user prediction in rating
  platforms}. In \bibinfo{booktitle}{\emph{Proceedings of the Eleventh ACM
  International Conference on Web Search and Data Mining}}.
  \bibinfo{pages}{333--341}.
\newblock


\bibitem[\protect\citeauthoryear{Li, Qin, Liu, Yang, and Li}{Li
  et~al\mbox{.}}{2019}]%
        {li2019spam}
\bibfield{author}{\bibinfo{person}{Ao Li}, \bibinfo{person}{Zhou Qin},
  \bibinfo{person}{Runshi Liu}, \bibinfo{person}{Yiqun Yang}, {and}
  \bibinfo{person}{Dong Li}.} \bibinfo{year}{2019}\natexlab{}.
\newblock \showarticletitle{Spam review detection with graph convolutional
  networks}. In \bibinfo{booktitle}{\emph{Proceedings of the 28th ACM
  International Conference on Information and Knowledge Management}}.
  \bibinfo{pages}{2703--2711}.
\newblock


\bibitem[\protect\citeauthoryear{Li, Zhang, Xi, and Zhu}{Li
  et~al\mbox{.}}{2018}]%
        {li2018hgsuspector}
\bibfield{author}{\bibinfo{person}{Xiang Li}, \bibinfo{person}{Wen Zhang},
  \bibinfo{person}{Jiuzhou Xi}, {and} \bibinfo{person}{Hao Zhu}.}
  \bibinfo{year}{2018}\natexlab{}.
\newblock \showarticletitle{HGsuspector: Scalable Collective Fraud Detection in
  Heterogeneous Graphs}.
\newblock  (\bibinfo{year}{2018}).
\newblock


\bibitem[\protect\citeauthoryear{Liang, Liu, Liu, Zhou, Li, Yang, and Qi}{Liang
  et~al\mbox{.}}{2019}]%
        {liang2019uncovering}
\bibfield{author}{\bibinfo{person}{Chen Liang}, \bibinfo{person}{Ziqi Liu},
  \bibinfo{person}{Bin Liu}, \bibinfo{person}{Jun Zhou},
  \bibinfo{person}{Xiaolong Li}, \bibinfo{person}{Shuang Yang}, {and}
  \bibinfo{person}{Yuan Qi}.} \bibinfo{year}{2019}\natexlab{}.
\newblock \showarticletitle{Uncovering Insurance Fraud Conspiracy with Network
  Learning}. In \bibinfo{booktitle}{\emph{Proceedings of the 42nd International
  ACM SIGIR Conference on Research and Development in Information Retrieval}}.
  \bibinfo{pages}{1181--1184}.
\newblock


\bibitem[\protect\citeauthoryear{Lin and Cohen}{Lin and Cohen}{2010}]%
        {lin2010power}
\bibfield{author}{\bibinfo{person}{Frank Lin} {and} \bibinfo{person}{William~W
  Cohen}.} \bibinfo{year}{2010}\natexlab{}.
\newblock \showarticletitle{Power iteration clustering}. In
  \bibinfo{booktitle}{\emph{ICML}}.
\newblock


\bibitem[\protect\citeauthoryear{Liu, Hooi, and Faloutsos}{Liu
  et~al\mbox{.}}{2017}]%
        {liu2017holoscope}
\bibfield{author}{\bibinfo{person}{Shenghua Liu}, \bibinfo{person}{Bryan Hooi},
  {and} \bibinfo{person}{Christos Faloutsos}.} \bibinfo{year}{2017}\natexlab{}.
\newblock \showarticletitle{Holoscope: Topology-and-spike aware fraud
  detection}. In \bibinfo{booktitle}{\emph{Proceedings of the 2017 ACM on
  Conference on Information and Knowledge Management}}.
  \bibinfo{pages}{1539--1548}.
\newblock


\bibitem[\protect\citeauthoryear{Liu, Chen, Li, Zhou, Li, Song, and Qi}{Liu
  et~al\mbox{.}}{2019}]%
        {liu2019geniepath}
\bibfield{author}{\bibinfo{person}{Ziqi Liu}, \bibinfo{person}{Chaochao Chen},
  \bibinfo{person}{Longfei Li}, \bibinfo{person}{Jun Zhou},
  \bibinfo{person}{Xiaolong Li}, \bibinfo{person}{Le Song}, {and}
  \bibinfo{person}{Yuan Qi}.} \bibinfo{year}{2019}\natexlab{}.
\newblock \showarticletitle{Geniepath: Graph neural networks with adaptive
  receptive paths}. In \bibinfo{booktitle}{\emph{Proceedings of the AAAI
  Conference on Artificial Intelligence}}, Vol.~\bibinfo{volume}{33}.
  \bibinfo{pages}{4424--4431}.
\newblock


\bibitem[\protect\citeauthoryear{Liu, Chen, Yang, Zhou, Li, and Song}{Liu
  et~al\mbox{.}}{2018}]%
        {liu2018heterogeneous}
\bibfield{author}{\bibinfo{person}{Ziqi Liu}, \bibinfo{person}{Chaochao Chen},
  \bibinfo{person}{Xinxing Yang}, \bibinfo{person}{Jun Zhou},
  \bibinfo{person}{Xiaolong Li}, {and} \bibinfo{person}{Le Song}.}
  \bibinfo{year}{2018}\natexlab{}.
\newblock \showarticletitle{Heterogeneous graph neural networks for malicious
  account detection}. In \bibinfo{booktitle}{\emph{Proceedings of the 27th ACM
  International Conference on Information and Knowledge Management}}.
  \bibinfo{pages}{2077--2085}.
\newblock


\bibitem[\protect\citeauthoryear{Liu, Dou, Yu, Deng, and Peng}{Liu
  et~al\mbox{.}}{2020}]%
        {liu2020alleviating}
\bibfield{author}{\bibinfo{person}{Zhiwei Liu}, \bibinfo{person}{Yingtong Dou},
  \bibinfo{person}{Philip~S Yu}, \bibinfo{person}{Yutong Deng}, {and}
  \bibinfo{person}{Hao Peng}.} \bibinfo{year}{2020}\natexlab{}.
\newblock \showarticletitle{Alleviating the Inconsistency Problem of Applying
  Graph Neural Network to Fraud Detection}.
\newblock \bibinfo{journal}{\emph{arXiv preprint arXiv:2005.00625}}
  (\bibinfo{year}{2020}).
\newblock


\bibitem[\protect\citeauthoryear{Lv, Ding, Liu, Chen, Feng, He, Zhou, Jiang,
  Dong, and Tang}{Lv et~al\mbox{.}}{2021}]%
        {lv2021we}
\bibfield{author}{\bibinfo{person}{Qingsong Lv}, \bibinfo{person}{Ming Ding},
  \bibinfo{person}{Qiang Liu}, \bibinfo{person}{Yuxiang Chen},
  \bibinfo{person}{Wenzheng Feng}, \bibinfo{person}{Siming He},
  \bibinfo{person}{Chang Zhou}, \bibinfo{person}{Jianguo Jiang},
  \bibinfo{person}{Yuxiao Dong}, {and} \bibinfo{person}{Jie Tang}.}
  \bibinfo{year}{2021}\natexlab{}.
\newblock \showarticletitle{Are we really making much progress? Revisiting,
  benchmarking, and refining heterogeneous graph neural networks}.
\newblock  (\bibinfo{year}{2021}).
\newblock


\bibitem[\protect\citeauthoryear{Ma, Zhang, Wang, Zhang, and Pozdnoukhov}{Ma
  et~al\mbox{.}}{2018}]%
        {ma2018graphrad}
\bibfield{author}{\bibinfo{person}{Jun Ma}, \bibinfo{person}{Danqing Zhang},
  \bibinfo{person}{Yun Wang}, \bibinfo{person}{Yan Zhang}, {and}
  \bibinfo{person}{Alexey Pozdnoukhov}.} \bibinfo{year}{2018}\natexlab{}.
\newblock \showarticletitle{GraphRAD: A Graph-based Risky Account Detection
  System}.
\newblock  (\bibinfo{year}{2018}).
\newblock


\bibitem[\protect\citeauthoryear{Min, Tang, Zhu, Dai, Wei, and Zhang}{Min
  et~al\mbox{.}}{2018}]%
        {min2018behavior}
\bibfield{author}{\bibinfo{person}{Wei Min}, \bibinfo{person}{Zhengyang Tang},
  \bibinfo{person}{Min Zhu}, \bibinfo{person}{Yuxi Dai}, \bibinfo{person}{Yan
  Wei}, {and} \bibinfo{person}{Ruinan Zhang}.} \bibinfo{year}{2018}\natexlab{}.
\newblock \showarticletitle{Behavior language processing with graph based
  feature generation for fraud detection in online lending}. In
  \bibinfo{booktitle}{\emph{Proceedings of Workshop on Misinformation and
  Misbehavior Mining on the Web}}.
\newblock


\bibitem[\protect\citeauthoryear{Nilforoshan and Shah}{Nilforoshan and
  Shah}{2019}]%
        {nilforoshan2019slicendice}
\bibfield{author}{\bibinfo{person}{Hamed Nilforoshan} {and}
  \bibinfo{person}{Neil Shah}.} \bibinfo{year}{2019}\natexlab{}.
\newblock \showarticletitle{SliceNDice: Mining Suspicious Multi-Attribute
  Entity Groups with Multi-View Graphs}. In \bibinfo{booktitle}{\emph{2019 IEEE
  International Conference on Data Science and Advanced Analytics (DSAA)}}.
  IEEE, \bibinfo{pages}{351--363}.
\newblock


\bibitem[\protect\citeauthoryear{Rao, Zhang, Han, Zhang, Min, Chen, Shan, Zhao,
  and Zhang}{Rao et~al\mbox{.}}{2021}]%
        {xfraudAppendix}
\bibfield{author}{\bibinfo{person}{Susie~Xi Rao}, \bibinfo{person}{Shuai
  Zhang}, \bibinfo{person}{Zhichao Han}, \bibinfo{person}{Zitao Zhang},
  \bibinfo{person}{Wei Min}, \bibinfo{person}{Zhiyao Chen},
  \bibinfo{person}{Yinan Shan}, \bibinfo{person}{Yang Zhao}, {and}
  \bibinfo{person}{Ce Zhang}.} \bibinfo{year}{2021}\natexlab{}.
\newblock \bibinfo{title}{Appendix for xFraud: Explainable Fraud Transaction
  Detection}.
\newblock
  \bibinfo{howpublished}{\url{https://github.com/eBay/xFraud/blob/master/documents/Appendix_XFraud_VLDB.pdf}}.
\newblock


\bibitem[\protect\citeauthoryear{Ren, Zhu, ZHang, Dai, and Bo}{Ren
  et~al\mbox{.}}{2019}]%
        {ren2019ensemfdet}
\bibfield{author}{\bibinfo{person}{Yuxiang Ren}, \bibinfo{person}{Hao Zhu},
  \bibinfo{person}{Jiawei ZHang}, \bibinfo{person}{Peng Dai}, {and}
  \bibinfo{person}{Liefeng Bo}.} \bibinfo{year}{2019}\natexlab{}.
\newblock \showarticletitle{EnsemFDet: An Ensemble Approach to Fraud Detection
  based on Bipartite Graph}.
\newblock \bibinfo{journal}{\emph{arXiv preprint arXiv:1912.11113}}
  (\bibinfo{year}{2019}).
\newblock


\bibitem[\protect\citeauthoryear{Shi, Hu, Zhao, and Philip}{Shi
  et~al\mbox{.}}{2018b}]%
        {shi2018heterogeneous}
\bibfield{author}{\bibinfo{person}{Chuan Shi}, \bibinfo{person}{Binbin Hu},
  \bibinfo{person}{Wayne~Xin Zhao}, {and} \bibinfo{person}{S~Yu Philip}.}
  \bibinfo{year}{2018}\natexlab{b}.
\newblock \showarticletitle{Heterogeneous information network embedding for
  recommendation}.
\newblock \bibinfo{journal}{\emph{IEEE Transactions on Knowledge and Data
  Engineering}} \bibinfo{volume}{31}, \bibinfo{number}{2}
  (\bibinfo{year}{2018}), \bibinfo{pages}{357--370}.
\newblock


\bibitem[\protect\citeauthoryear{Shi, Han, He, He, Yang, Luo, and Han}{Shi
  et~al\mbox{.}}{2018a}]%
        {shi2018mvn2vec}
\bibfield{author}{\bibinfo{person}{Yu Shi}, \bibinfo{person}{Fangqiu Han},
  \bibinfo{person}{Xinwei He}, \bibinfo{person}{Xinran He},
  \bibinfo{person}{Carl Yang}, \bibinfo{person}{Jie Luo}, {and}
  \bibinfo{person}{Jiawei Han}.} \bibinfo{year}{2018}\natexlab{a}.
\newblock \showarticletitle{mvn2vec: Preservation and collaboration in
  multi-view network embedding}.
\newblock \bibinfo{journal}{\emph{arXiv preprint arXiv:1801.06597}}
  (\bibinfo{year}{2018}).
\newblock


\bibitem[\protect\citeauthoryear{Shu, Mahudeswaran, Wang, and Liu}{Shu
  et~al\mbox{.}}{2020}]%
        {shu2020hierarchical}
\bibfield{author}{\bibinfo{person}{Kai Shu}, \bibinfo{person}{Deepak
  Mahudeswaran}, \bibinfo{person}{Suhang Wang}, {and} \bibinfo{person}{Huan
  Liu}.} \bibinfo{year}{2020}\natexlab{}.
\newblock \showarticletitle{Hierarchical propagation networks for fake news
  detection: Investigation and exploitation}. In
  \bibinfo{booktitle}{\emph{Proceedings of the International AAAI Conference on
  Web and Social Media}}, Vol.~\bibinfo{volume}{14}. \bibinfo{pages}{626--637}.
\newblock


\bibitem[\protect\citeauthoryear{Vaswani, Shazeer, Parmar, Uszkoreit, Jones,
  Gomez, Kaiser, and Polosukhin}{Vaswani et~al\mbox{.}}{2017}]%
        {vaswani2017attention}
\bibfield{author}{\bibinfo{person}{Ashish Vaswani}, \bibinfo{person}{Noam
  Shazeer}, \bibinfo{person}{Niki Parmar}, \bibinfo{person}{Jakob Uszkoreit},
  \bibinfo{person}{Llion Jones}, \bibinfo{person}{Aidan~N Gomez},
  \bibinfo{person}{{\L}ukasz Kaiser}, {and} \bibinfo{person}{Illia
  Polosukhin}.} \bibinfo{year}{2017}\natexlab{}.
\newblock \showarticletitle{Attention is all you need}. In
  \bibinfo{booktitle}{\emph{Advances in neural information processing
  systems}}. \bibinfo{pages}{5998--6008}.
\newblock


\bibitem[\protect\citeauthoryear{Wang, Zhou, Wu, Dang, Zhu, and Wang}{Wang
  et~al\mbox{.}}{2018}]%
        {wang2018deep}
\bibfield{author}{\bibinfo{person}{Haibo Wang}, \bibinfo{person}{Chuan Zhou},
  \bibinfo{person}{Jia Wu}, \bibinfo{person}{Weizhen Dang},
  \bibinfo{person}{Xingquan Zhu}, {and} \bibinfo{person}{Jilong Wang}.}
  \bibinfo{year}{2018}\natexlab{}.
\newblock \showarticletitle{Deep structure learning for fraud detection}. In
  \bibinfo{booktitle}{\emph{2018 IEEE International Conference on Data Mining
  (ICDM)}}. IEEE, \bibinfo{pages}{567--576}.
\newblock


\bibitem[\protect\citeauthoryear{Wang, Wen, Wu, Huang, and Xion}{Wang
  et~al\mbox{.}}{2019b}]%
        {wang2019fdgars}
\bibfield{author}{\bibinfo{person}{Jianyu Wang}, \bibinfo{person}{Rui Wen},
  \bibinfo{person}{Chunming Wu}, \bibinfo{person}{Yu Huang}, {and}
  \bibinfo{person}{Jian Xion}.} \bibinfo{year}{2019}\natexlab{b}.
\newblock \showarticletitle{Fdgars: Fraudster detection via graph convolutional
  networks in online app review system}. In \bibinfo{booktitle}{\emph{Companion
  Proceedings of The 2019 World Wide Web Conference}}.
  \bibinfo{pages}{310--316}.
\newblock


\bibitem[\protect\citeauthoryear{Wang, Ji, Shi, Wang, Ye, Cui, and Yu}{Wang
  et~al\mbox{.}}{2019a}]%
        {wang2019han}
\bibfield{author}{\bibinfo{person}{Xiao Wang}, \bibinfo{person}{Houye Ji},
  \bibinfo{person}{Chuan Shi}, \bibinfo{person}{Bai Wang},
  \bibinfo{person}{Yanfang Ye}, \bibinfo{person}{Peng Cui}, {and}
  \bibinfo{person}{Philip~S Yu}.} \bibinfo{year}{2019}\natexlab{a}.
\newblock \showarticletitle{Heterogeneous graph attention network}. In
  \bibinfo{booktitle}{\emph{The World Wide Web Conference}}.
  \bibinfo{pages}{2022--2032}.
\newblock


\bibitem[\protect\citeauthoryear{Weber, Domeniconi, Chen, Weidele, Bellei,
  Robinson, and Leiserson}{Weber et~al\mbox{.}}{2019}]%
        {weber2019anti}
\bibfield{author}{\bibinfo{person}{Mark Weber}, \bibinfo{person}{Giacomo
  Domeniconi}, \bibinfo{person}{Jie Chen}, \bibinfo{person}{Daniel Karl~I
  Weidele}, \bibinfo{person}{Claudio Bellei}, \bibinfo{person}{Tom Robinson},
  {and} \bibinfo{person}{Charles~E Leiserson}.}
  \bibinfo{year}{2019}\natexlab{}.
\newblock \showarticletitle{Anti-money laundering in bitcoin: Experimenting
  with graph convolutional networks for financial forensics}.
\newblock \bibinfo{journal}{\emph{arXiv preprint arXiv:1908.02591}}
  (\bibinfo{year}{2019}).
\newblock


\bibitem[\protect\citeauthoryear{Wen, Wang, Wu, and Xiong}{Wen
  et~al\mbox{.}}{2020}]%
        {wen2020asa}
\bibfield{author}{\bibinfo{person}{Rui Wen}, \bibinfo{person}{Jianyu Wang},
  \bibinfo{person}{Chunming Wu}, {and} \bibinfo{person}{Jian Xiong}.}
  \bibinfo{year}{2020}\natexlab{}.
\newblock \showarticletitle{ASA: Adversary Situation Awareness via
  Heterogeneous Graph Convolutional Networks}. In
  \bibinfo{booktitle}{\emph{Companion Proceedings of the Web Conference 2020}}.
  \bibinfo{pages}{674--678}.
\newblock


\bibitem[\protect\citeauthoryear{X.~Li and Veloso}{X.~Li and Veloso}{2020}]%
        {li2020gnnexplainer-fin}
\bibfield{author}{\bibinfo{person}{P.~Reddy X.~Li, J.~Saude} {and}
  \bibinfo{person}{M. Veloso}.} \bibinfo{year}{2020}\natexlab{}.
\newblock \showarticletitle{Classifying and understanding financial data using
  graph neural network}. In \bibinfo{booktitle}{\emph{AAAI}}.
\newblock


\bibitem[\protect\citeauthoryear{Yang, Xiao, Zhang, Sun, and Han}{Yang
  et~al\mbox{.}}{2020}]%
        {yang2020heterogeneous}
\bibfield{author}{\bibinfo{person}{Carl Yang}, \bibinfo{person}{Yuxin Xiao},
  \bibinfo{person}{Yu Zhang}, \bibinfo{person}{Yizhou Sun}, {and}
  \bibinfo{person}{Jiawei Han}.} \bibinfo{year}{2020}\natexlab{}.
\newblock \showarticletitle{Heterogeneous network representation learning: A
  unified framework with survey and benchmark}.
\newblock \bibinfo{journal}{\emph{IEEE Transactions on Knowledge and Data
  Engineering}} (\bibinfo{year}{2020}).
\newblock


\bibitem[\protect\citeauthoryear{Ying, Bourgeois, You, Zitnik, and
  Leskovec}{Ying et~al\mbox{.}}{2019}]%
        {ying2019gnnexplainer}
\bibfield{author}{\bibinfo{person}{Zhitao Ying}, \bibinfo{person}{Dylan
  Bourgeois}, \bibinfo{person}{Jiaxuan You}, \bibinfo{person}{Marinka Zitnik},
  {and} \bibinfo{person}{Jure Leskovec}.} \bibinfo{year}{2019}\natexlab{}.
\newblock \showarticletitle{Gnnexplainer: Generating explanations for graph
  neural networks}. In \bibinfo{booktitle}{\emph{Advances in neural information
  processing systems}}. \bibinfo{pages}{9244--9255}.
\newblock


\bibitem[\protect\citeauthoryear{Yuan, Tang, Hu, and Ji}{Yuan
  et~al\mbox{.}}{2020}]%
        {yuan2020xgnn}
\bibfield{author}{\bibinfo{person}{Hao Yuan}, \bibinfo{person}{Jiliang Tang},
  \bibinfo{person}{Xia Hu}, {and} \bibinfo{person}{Shuiwang Ji}.}
  \bibinfo{year}{2020}\natexlab{}.
\newblock \showarticletitle{XGNN: Towards Model-Level Explanations of Graph
  Neural Networks}.
\newblock \bibinfo{journal}{\emph{arXiv preprint arXiv:2006.02587}}
  (\bibinfo{year}{2020}).
\newblock


\bibitem[\protect\citeauthoryear{Yun, Jeong, Kim, Kang, and Kim}{Yun
  et~al\mbox{.}}{2019}]%
        {yun2019graph}
\bibfield{author}{\bibinfo{person}{Seongjun Yun}, \bibinfo{person}{Minbyul
  Jeong}, \bibinfo{person}{Raehyun Kim}, \bibinfo{person}{Jaewoo Kang}, {and}
  \bibinfo{person}{Hyunwoo~J Kim}.} \bibinfo{year}{2019}\natexlab{}.
\newblock \showarticletitle{Graph transformer networks}.
\newblock \bibinfo{journal}{\emph{Advances in Neural Information Processing
  Systems}}  \bibinfo{volume}{32} (\bibinfo{year}{2019}),
  \bibinfo{pages}{11983--11993}.
\newblock


\bibitem[\protect\citeauthoryear{Zhang, Song, Huang, Swami, and Chawla}{Zhang
  et~al\mbox{.}}{2019b}]%
        {zhang2019hetgnn}
\bibfield{author}{\bibinfo{person}{Chuxu Zhang}, \bibinfo{person}{Dongjin
  Song}, \bibinfo{person}{Chao Huang}, \bibinfo{person}{Ananthram Swami}, {and}
  \bibinfo{person}{Nitesh~V Chawla}.} \bibinfo{year}{2019}\natexlab{b}.
\newblock \showarticletitle{Heterogeneous graph neural network}. In
  \bibinfo{booktitle}{\emph{Proceedings of the 25th ACM SIGKDD International
  Conference on Knowledge Discovery \& Data Mining}}.
  \bibinfo{pages}{793--803}.
\newblock


\bibitem[\protect\citeauthoryear{Zhang, Fan, Ye, Zhao, and Shi}{Zhang
  et~al\mbox{.}}{2019a}]%
        {zhang2019key}
\bibfield{author}{\bibinfo{person}{Yiming Zhang}, \bibinfo{person}{Yujie Fan},
  \bibinfo{person}{Yanfang Ye}, \bibinfo{person}{Liang Zhao}, {and}
  \bibinfo{person}{Chuan Shi}.} \bibinfo{year}{2019}\natexlab{a}.
\newblock \showarticletitle{Key Player Identification in Underground Forums
  over Attributed Heterogeneous Information Network Embedding Framework}. In
  \bibinfo{booktitle}{\emph{Proceedings of the 28th ACM International
  Conference on Information and Knowledge Management}}.
  \bibinfo{pages}{549--558}.
\newblock


\bibitem[\protect\citeauthoryear{Zhao, Bai, Wu, Wang, Zhang, Yang, and
  Nie}{Zhao et~al\mbox{.}}{2020}]%
        {zhao2020deep}
\bibfield{author}{\bibinfo{person}{Kai Zhao}, \bibinfo{person}{Ting Bai},
  \bibinfo{person}{Bin Wu}, \bibinfo{person}{Bai Wang}, \bibinfo{person}{Youjie
  Zhang}, \bibinfo{person}{Yuanyu Yang}, {and} \bibinfo{person}{Jian-Yun Nie}.}
  \bibinfo{year}{2020}\natexlab{}.
\newblock \showarticletitle{Deep adversarial completion for sparse
  heterogeneous information network embedding}. In
  \bibinfo{booktitle}{\emph{Proceedings of the Web Conference 2020}}.
  \bibinfo{pages}{508--518}.
\newblock


\bibitem[\protect\citeauthoryear{Zhong, Liu, Ao, Hu, Feng, Tang, and He}{Zhong
  et~al\mbox{.}}{2020}]%
        {zhong2020financial}
\bibfield{author}{\bibinfo{person}{Qiwei Zhong}, \bibinfo{person}{Yang Liu},
  \bibinfo{person}{Xiang Ao}, \bibinfo{person}{Binbin Hu},
  \bibinfo{person}{Jinghua Feng}, \bibinfo{person}{Jiayu Tang}, {and}
  \bibinfo{person}{Qing He}.} \bibinfo{year}{2020}\natexlab{}.
\newblock \showarticletitle{Financial Defaulter Detection on Online Credit
  Payment via Multi-view Attributed Heterogeneous Information Network}. In
  \bibinfo{booktitle}{\emph{Proceedings of The Web Conference 2020}}.
  \bibinfo{pages}{785--795}.
\newblock


\bibitem[\protect\citeauthoryear{Zhu, Xi, Song, Zhuang, Chen, Gu, and He}{Zhu
  et~al\mbox{.}}{2020}]%
        {zhu2020modeling}
\bibfield{author}{\bibinfo{person}{Yongchun Zhu}, \bibinfo{person}{Dongbo Xi},
  \bibinfo{person}{Bowen Song}, \bibinfo{person}{Fuzhen Zhuang},
  \bibinfo{person}{Shuai Chen}, \bibinfo{person}{Xi Gu}, {and}
  \bibinfo{person}{Qing He}.} \bibinfo{year}{2020}\natexlab{}.
\newblock \showarticletitle{Modeling Users’ Behavior Sequences with
  Hierarchical Explainable Network for Cross-domain Fraud Detection}. In
  \bibinfo{booktitle}{\emph{Proceedings of The Web Conference 2020}}.
  \bibinfo{pages}{928--938}.
\newblock


\end{thebibliography}

\clearpage
\nobalance
\appendix

\section{heterogeneous datasets in the literature}
\label{app:survey-hetero}

{We list the heterogeneous datasets used in the literature in the past six years in the following table. }

\begin{table}[!h]
\centering
\caption{{A survey of the existing heterogeneous graph data sets (``B":billion;``M":million;``K":thousand).}}
\label{tab:survey-hetero}
\resizebox{1.0\linewidth}{!}{
\begin{tabular}{cccrrr}
\toprule
\textbf{Year}                   & \textbf{Paper}                          & \textbf{Dataset}      & \textbf{\# Node}      & \textbf{\# Edge}         & \textbf{\# Edge/\# Node} \\
\midrule
2015                   & HNE (2015) \cite{chang2015heterogeneous}                    & BlogCatalog  & 5,196      & 171,743       & 33.05     \\ \hline
\multirow{5}{*}{2017}  & \multirow{5}{*}{MVE \cite{shi2018heterogeneous}}           & PPI          & 16,545     & 1,098,711     & 66.41     \\
                       &                                & DBLP         & 69,110     & 1,884,236     & 27.26     \\
                       &                                & Youtube      & 14,901     & 13,552,130    & 909.48    \\
                       &                                & Twitter      & 304,692    & 131,151,083   & 430.44    \\ 
                       &                     & Flickr       & 35,314     & 6,548,830     & 185.45    \\
                       \midrule
\multirow{9}{*}{2018}  & GEM     \cite{liu2018heterogeneous}                       & GEM-graph    & 8M  & 10M    & 1.67      \\ \cline{2-6}
                       & \multirow{3}{*}{HERec \cite{shi2018heterogeneous}}         & Yelp         & 95,110     & 488,120       & 5.13      \\
                       &                                & Douban Book  & 138,423    & 1,026,046     & 7.41      \\
                       &                                & Douban Movie & 90,241     & 1,714,941     & 19.00     \\ \cline{2-6}
                       & \multirow{2}{*}{metapath2vec \cite{dong2017metapath2vec}}  & DBIS         & 78,366     & 326,481       & 4.17      \\
                       &                                & AMiner CS    & 12,522,027 & 14,215,558    & 1.14      \\ \cline{2-6}
                       & \multirow{3}{*}{mvn2vec \cite{shi2018mvn2vec}}       & Twitter      & 116,408    & 183,341       & 1.57      \\
                       &                                & Youtube      & 14,900     & 7,977,881     & 535.43    \\
                       &                                & Snapchat     & 7,406,859  & 131,729,903   & 17.78     \\ \midrule
\multirow{16}{*}{2019} & \multirow{5}{*}{GATNE \cite{cen2019representation}}         & Alibaba-S    & 6,163      & 17,865        & 2.90      \\
                       &                                & Amazon-GATNE & 312,320    & 7,500,100     & 24.01     \\
                       &                                & YouTube      & 15,088     & 13,628,895    & 903.29    \\
                       &                                & Twitter      & 456,626    & 15,367,315    & 33.65     \\
                       &                                & Alibaba      & 41,991,048 & 571,892,183   & 13.62     \\ \cline{2-6}
                       & GTN   \cite{yun2019graph}                         & DBLP         & 26128      & 239,566       & 9.17      \\ \cline{2-6}
                       & \multirow{2}{*}{HAN \cite{wang2019han}}           & IMDB         & 21420      & 86,642        & 4.04      \\
                       &                                & ACM          & 10942      & 547,872       & 50.07     \\ \cline{2-6}
                       & \multirow{4}{*}{HeGAN \cite{hu2019adversarial}}         & Yelp         & 3,913      & 38,680        & 9.88      \\
                       &                                & DBLP         & 37,791     & 170,794       & 4.52      \\
                       &                                & Aminer       & 312,776    & 599,951       & 1.92      \\
                       &                                & Movielens    & 10,038     & 1,014,164     & 101.03    \\ \cline{2-6}
                       & \multirow{4}{*}{HetGNN \cite{zhang2019hetgnn}}        & Academic II  & 49,708     & 137,286       & 2.76      \\
                       &                                & Academic I   & 272,272    & 544,976       & 2.00      \\
                       &                                & CDs Review   & 123,736    & 555,050       & 4.49      \\
                       &                                & Movie Review & 74,701     & 629,125       & 8.42      \\ \midrule
\multirow{9}{*}{2020}  & HGT \cite{hu2020hgt}                           & ogbn-mag     & 179M  & 2B & 11.17     \\ \cline{2-6}
                       & HNE \cite{yang2020heterogeneous}                            & PubMed       & 63,109      & 244,986       & 3.88      \\ \cline{2-6}
                       & MAGNN \cite{fu2020magnn}                          & LastFM-r     & 71,689      & 3,034,763     & 42.33     \\ \cline{2-6}
                       & \multirow{6}{*}{MV-ACM \cite{zhao2020deep}}        & Amazon       & 10,099     & 113,637       & 11.25     \\
                       &                                & Alibaba      & 40,324     & 149,587       & 3.71      \\
                       &                                & Twitter      & 40,000     & 1,028,364     & 25.71     \\
                       &                                & PPI          & 15,005     & 1,044,541     & 69.61     \\
                       &                                & Youtube      & 2,000      & 1,114,025     & 557.01    \\
                       &                                & Aminer       & 178,385    & 5,935,349     & 33.27     \\ \midrule
\multirow{9}{*}{2021}  & \multirow{6}{*}{HGB \cite{lv2021we}}           & LastFM       & 20,612      & 141,521       & 6.87      \\
                       &                                & Amazon       & 10,099      & 148,659       & 14.72     \\
                       &                                & Freebase     & 180,098     & 148,659       & 0.83      \\
                       &                                & Movielens    & 43,567      & 539,300       & 12.38     \\
                       &                                & Amazon-book  & 95,594      & 846,434       & 8.85      \\
                       &                                & Yelp-2018    & 91,457      & 1,183,610     & 12.94     \\ \cline{2-6}
                       & \multirow{3}{*}{xFraud (ours)} & eBay-small   & 288,853    & 612,904       & 2.12      \\
                       &                                & eBay-large   & 8,857,866  & 13,158,984    & 1.49      \\
                       &                                & eBay-xlarge  & 1.1B & 3.7B & 3.36    \\ \bottomrule  
\end{tabular}}
\end{table}

\section{Dataset construction}\label{app:dataset}

In the transaction records, there exist a rich set of relations when two transactions share some linkage entities, e.g., executed by the same buyer, using the same payment instrument, shipped to the same address. 

\begin{table}[!t]
    \centering
    \caption{{Node type counts $|\mathcal{V}_t|$ in heterogeneous graphs.}}
    \label{tab:typecount}
    \resizebox{0.8\linewidth}{!}{
    
    \begin{tabular}{ccrr}
    \toprule
\textbf{Dataset}             & \textbf{Node type} & \multicolumn{1}{c}{\textbf{\#Count}} & \multicolumn{1}{c}{\textbf{Node type\%}} \\

                             \midrule
\multirow{5}{*}{\textit{eBay-xlarge}} & txn                & 857M            & 77\%                                     \\
                             & pmt                & 81M             & 7\%                                      \\
                             & email              & 72M             & 6\%                                      \\
                             & addr               & 62M             & 5\%                                      \\
                             & buyer              & 69M             & 5\%    \\
                             \midrule

\multirow{5}{*}{\textit{eBay-large}}  & txn                & 3,752,225                            & 42.40\%                                  \\
                             & pmt                & 1,180,114                            & 13.30\%                                  \\
                             & email              & 1,307,179                            & 14.80\%                                  \\
                             & addr               & 1,316,251                            & 14.90\%                                  \\
                             & buyer              & 1,302,097                            & 14.60\%                                  \\ \midrule
\multirow{5}{*}{\textit{eBay-small}}  & txn                & 207,749                              & 71.90\%                                  \\
                             & pmt                & 22,273                               & 7.70\%                                   \\
                             & email              & 25,878                               & 9.00\%                                   \\
                             & addr               & 7,138                                & 2.40\%                                   \\
                             & buyer              & 25,815                               & 9.00\%                                   \\
                             
                             \bottomrule
\end{tabular}
    
}
\end{table}


Following the same graph construction protocol, we construct \textit{heterogeneous graphs} out of the transaction records in different scales --- \textit{eBay-xlarge}, \textit{eBay-large}, and \textit{eBay-small}.  Note that \textit{eBay-large} and \textit{eBay-small} are subsets of \textit{eBay-xlarge}. 
The graph construction protocol is as follows. Both transaction and linkage entities are treated as nodes. If an entity is used in a transaction, we create an edge between the transaction node and that entity (see Sec.~\ref{sec:hetG-constructor}). Only transaction node has input features. 

We construct three datasets from the historical transactions at eBay. Historical transaction records were sampled to generate {three} datasets with different scales. Note that the ratio of fraudulent transactions is usually much smaller than that of the legitimate ones; hence, we try to down sample the benign transactions to alleviate the class imbalance problem. 

{\paragraph{\textit{\textbf{eBay-xlarge.}}} It is a billion-scale dataset of 1.1 billion nodes and 3.7 billion edges, which contains seven months of selected transaction records. The dimension of transaction features is 480.}

\paragraph{\textit{\textbf{eBay-large.}}} All the historical transaction records spanning a given period in {\em eBay-xlarge} are sampled. The dimension of transaction features is 480. The linking entities whose transaction numbers below a predefined threshold are removed to maintain graph connectivity, which means that there is no isolated transaction in the graph.

\paragraph{\textbf{\textit{eBay-small.}}} We take a subset of {\textit{eBay-large} to construct \textit{eBay-small}} by shrinking the transaction spanning period, and the dimension of transaction features is 114. 

{For all three datasets}, to further reduce the graph size while preserving graph connectivity, we adopt a graph sampling strategy: (1) All fraudulent transactions and randomly sampled benign transactions are selected as seeds. (2) Each seed is expanded to its {$k$}-hop neighbors, and at each hop, no more than {$N$} neighbors are picked. (3) The neighborhoods around the seeds with transaction numbers less than five are filtered out. {We have a similar graph sampling across all datasets: for \textit{eBay-xlarge}, $k = 8, N = 512$, for \textit{eBay-large} and \textit{eBay-small} $k = 3, N = 32$.}

In addition, for the \textit{eBay-xlarge} dataset, we produce the training/testing labels via two filtering and sampling steps: 
\begin{enumerate}[wide=0pt]
    \item The original data stream has a fraud rate = 0.016\%. 
    \item We then use some simple rules to filter out certain low-risk transactions. These rules are already implemented in the eBay transaction platforms to filter transactions. This is similar to what GEM~\cite{liu2018heterogeneous} does in graph preprocessing and is also consistent with how this model will be used in practice. After this step, we have a fraud rate = 0.043\%.
    \item To create the training and testing labels, we sampled all fraud transactions, and 1\% of non-fraud transactions. Note that the other transactions are still in the graph, but without supervised labels. This gives us the \textit{eBay-xlarge} dataset with a fraud rate = 4.33\%.
\end{enumerate}

The ratio of transaction frauds is 4.33\%, 3.57\%, and 4.30\%, in \textit{eBay-xlarge}, \textit{eBay-large}, and \textit{eBay-small}. respectively. We further summarize the distribution of each node type in Table~\ref{tab:typecount}.

\begin{figure}[t]
    \centering
    \includegraphics[width=0.45\textwidth]{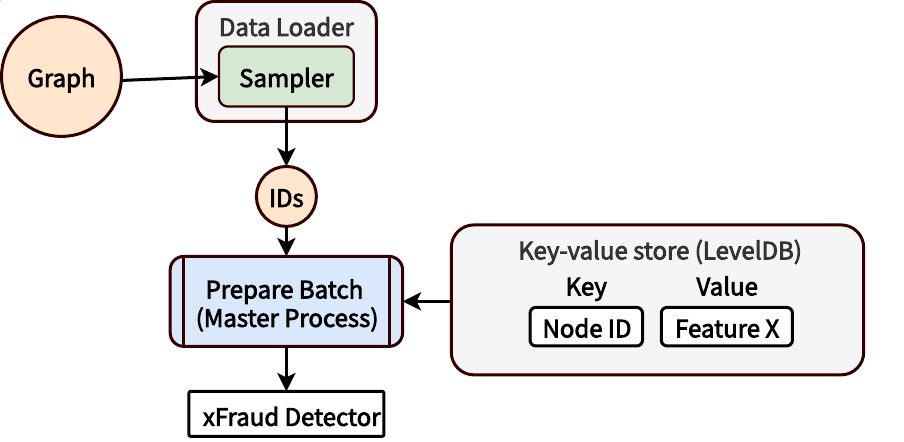}
    \caption{(Previous implementation) Using a single threaded KVStore to interact with GNN on a single machine.}
    \label{fig:single-kv}
\end{figure}

\begin{figure}[t]
    \centering
    \includegraphics[width=0.45\textwidth]{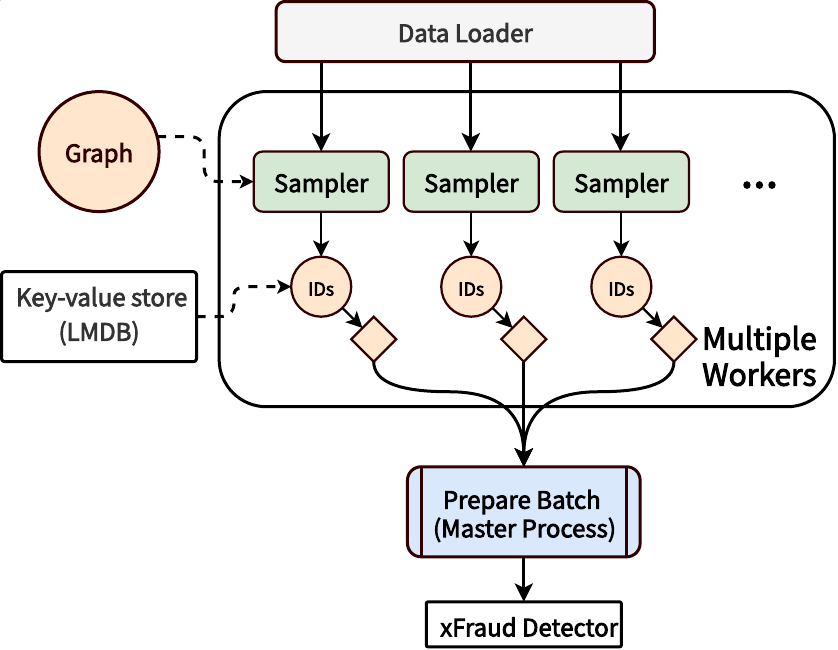}
    \caption{(New implementation) Using a multi threaded KVStore to interact with GNN on a single machine.}
    \label{fig:multi-kv}
\end{figure}
\begin{table*}[!t]
\centering
\caption{Model performance in distributed settings on \textit{eBay-xlarge} (\# epochs: 128). The best score of a column is in \textbf{bold}.}
\label{tab:xlarge-detector-result}
\resizebox{\linewidth}{!}{
\begin{tabular}{cccccccc}
\toprule
\textbf{Model}                    & \textbf{\# of machines} & \textbf{Seed} & \textbf{Accuracy} & \textbf{AP} & \textbf{AUC} & \textbf{Training time (s/epoch)} & \textbf{Inference time (s/batch)} \\ \hline

\multirow{4}{*}{GAT}              & \multirow{2}{*}{8}      & A          & 0.9334 &	0.4478 &	0.8894
       & 62.62                            & \multirow{4}{*}{0.0557 $\pm$ 0.1966}       \\
                                  &                         & B          & 0.9309 &	0.4120	& 0.8863
      & 62.85                            &                             \\ \cline{2-7}
                                  & \multirow{2}{*}{16}     & A          & 0.9348 &	0.4481 &	0.8894
       & {\bf 32.08}                          &       \\
                                  &                         & B          & 0.9425 &	0.3932	& 0.8838
       & 34.14                            &                             \\ \hline
\multirow{4}{*}{GEM}              & \multirow{2}{*}{8}      & A          & 0.9554 &	0.4513 &	0.8960
       & 61.67                            & \multirow{4}{*}{\bf 0.0167$\pm$ 0.0054}       \\
                                  &                         & B          & 0.9578 &	0.4613 &	0.8962
       & 61.86                            &                             \\  \cline{2-7}  
                                  & \multirow{2}{*}{16}     & A          & 0.9552 &	0.4349 &	0.8949
       & 33.40                            &       \\ 
                                  &                         & B          & 0.9574 &	0.4363 &	0.8926
       & 33.71                            &                             \\ \hline
\multirow{4}{*}{xFraud detector+} & \multirow{2}{*}{8}      & A          & 0.9701 &	\textbf{0.5942} &	\textbf{0.9074}
       & 71.79                            & \multirow{4}{*}{0.0799 $\pm$ 0.1868}       \\
                                  &                         & B          & \textbf{0.9703} &	0.5931 &	0.9073
      & 69.15                            &                             \\ \cline{2-7}
                                  & \multirow{2}{*}{16}     & A          & 0.9694 &	0.5483 &	0.8900
     & 37.35                            &       \\
                                  &                         & B          & 0.9650 &	0.4932 &	0.8883
     & 40.09                            &                            \\                               
\bottomrule
\end{tabular}}
\end{table*}

\section{More Details on the Distributed xFraud detector}\label{app:distributed}
We introduce the experimental protocols and an extensive set of results on our billion-scale dataset \textit{eBay-xlarge}.

{\paragraph{\textbf{Implementation of KVStores}} We have implemented two types of KVStores when optimizing xFraud detector+. The multi threaded KVStore solution is used in the distributed setting. We show the detailed implementation of the interaction between GNN and the single threaded KVstore and the multi threaded KVStore in Figure \ref{fig:single-kv} and \ref{fig:multi-kv}, respectively. With this new optimization, we manage to reduce the training (incl. data loading) on the \textit{eBay-large} to 1 minute per epoch. In comparison, a single-threaded KVStore on the same dataset would take 45 min/epoch. }

\paragraph{\textbf{Further experimental details for detector.}}
\begin{itemize} [wide=0pt]
    \item {\textit{Machines}.} We conduct the {single-machine} experiments on a Linux server with 16 Intel Xeon Gold 6230 CPU, one Nvidia Tesla V100 32 GB GPU, and 256 GB memory. {For the distributed settings, we use a set of machines, each of which has 4 Intel Xeon Gold 6230 CPU, one Nvidia Tesla V100 32 GB GPU, and 32 GB memory.}
    \item {\textit{Hyperparameters}.} The baselines and xFraud detector+ are trained with this set of hyperparameters: \text{\textit{n\_hid}} = 400, \text{\textit{n\_heads}} = 8, \text{\textit{n\_layers}} = 6, \text{\textit{dropout}} = 0.2, \text{\textit{optimizer}} = \text{\textit{``adamw"}}, \text{\textit{clip}} = 0.25,     \text{\textit{max\_epochs}} = 128, \text{\textit{patience}} = 32. We keep the same set of hyperparameters for all datasets.
\end{itemize}

\paragraph{\textbf{{Further experimental results.}}} 
{We present the complete results in Table~\ref{tab:xlarge-detector-result} on two different seeds (A and B) using 8 machines and 16 machines. }

\begin{figure*}[]
    \centering
      \begin{tabular}  {ccc}
    \includegraphics[width=.311\linewidth]{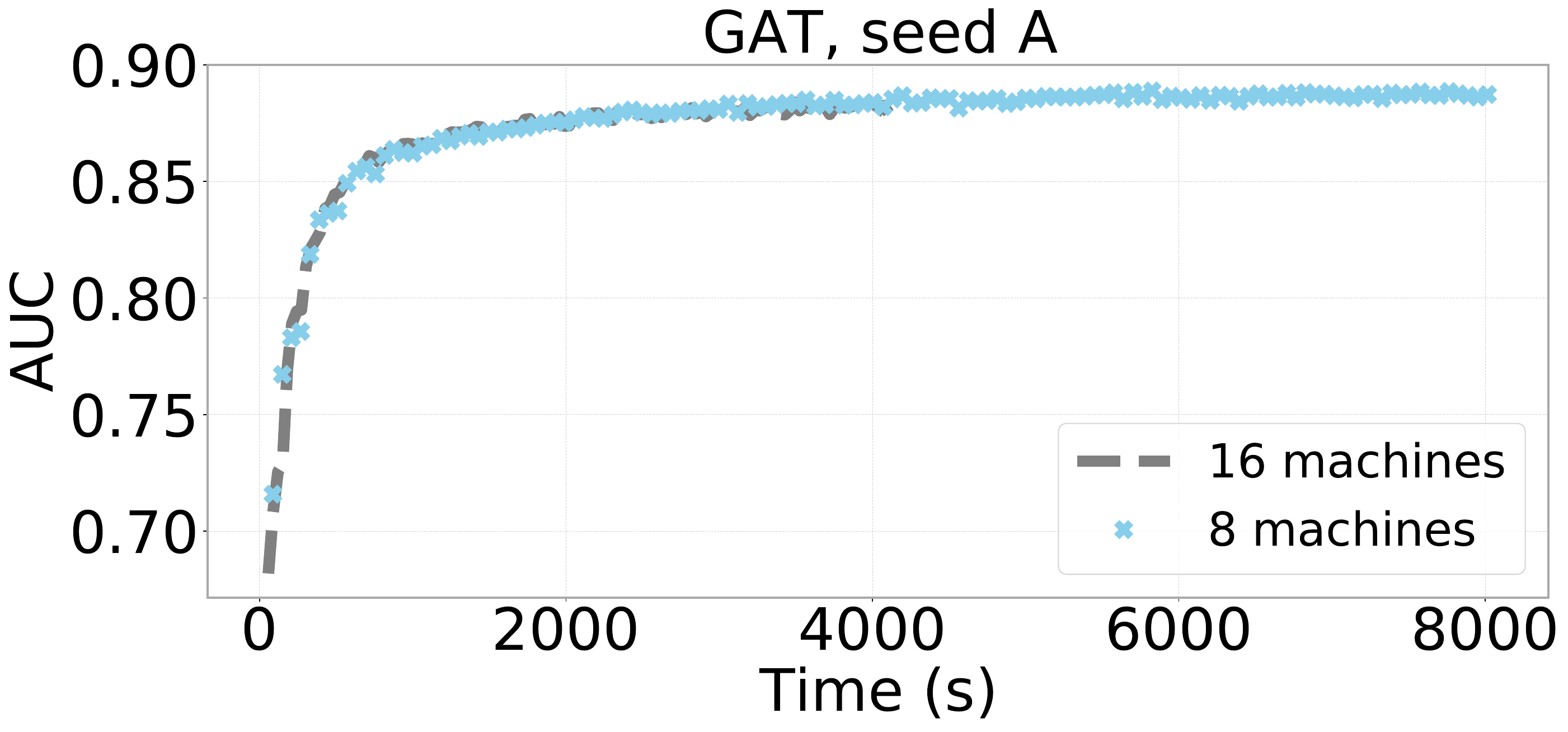}  & 
    \includegraphics[width=.311\linewidth]{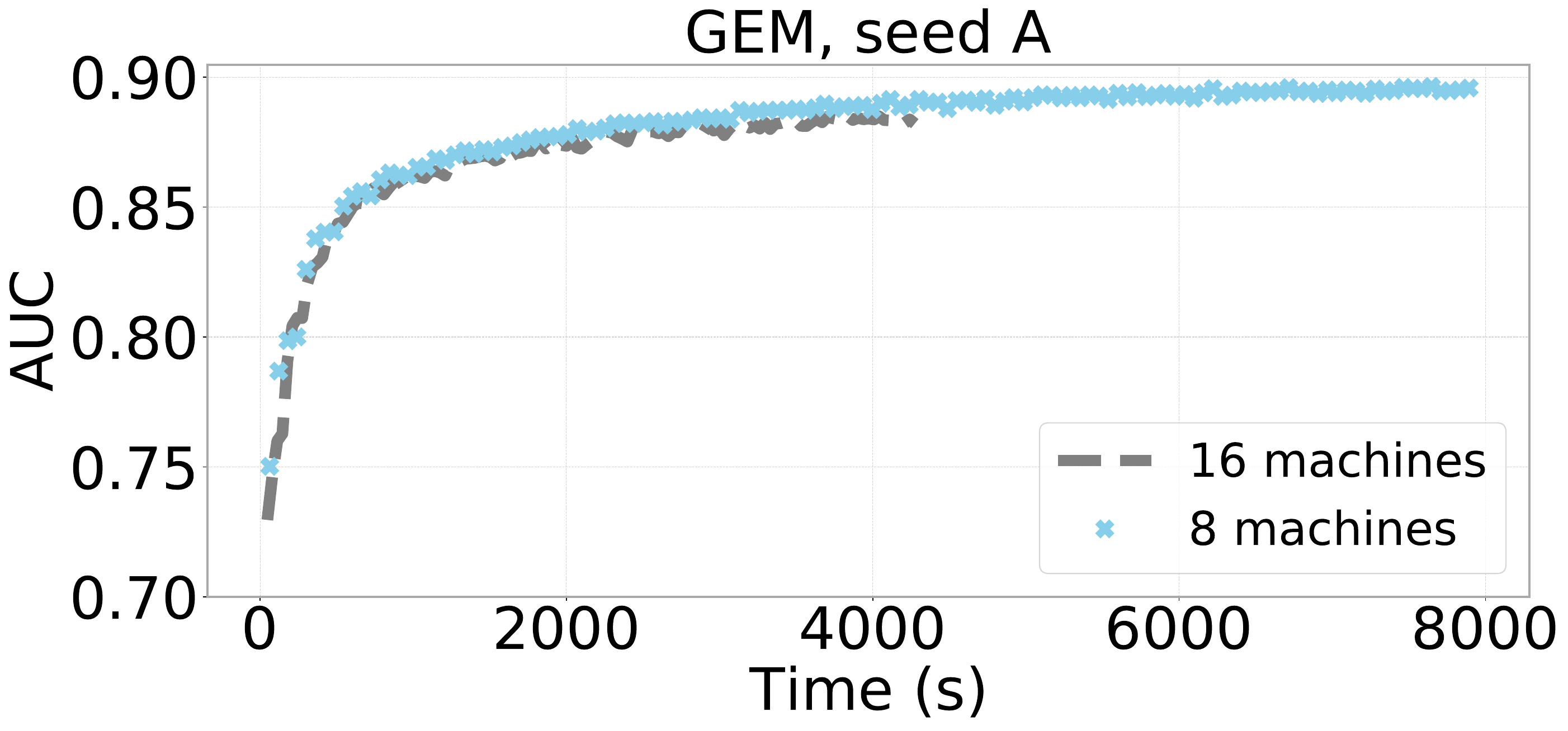} &
    \includegraphics[width=.311\linewidth]{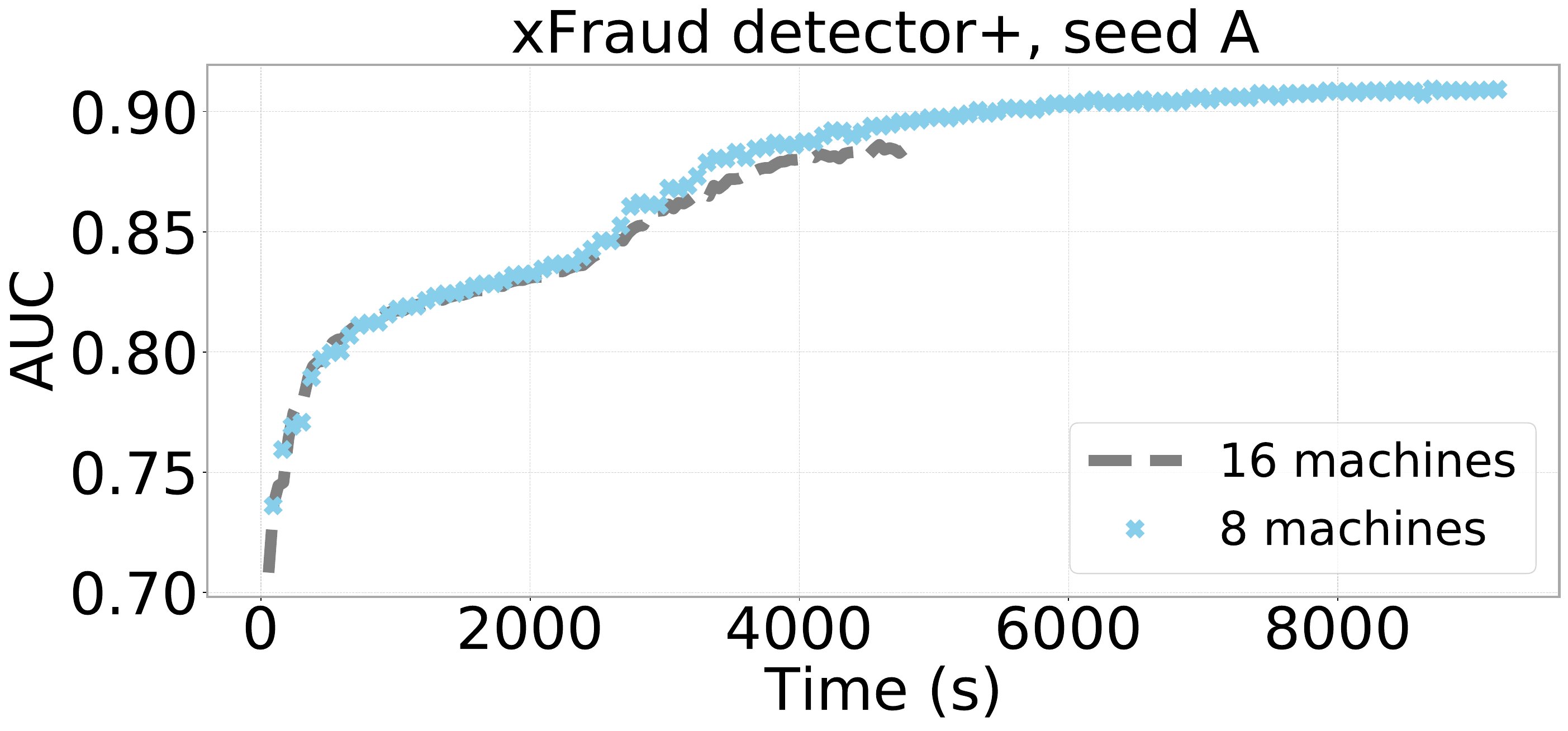} \\
    
    \includegraphics[width=.311\linewidth]{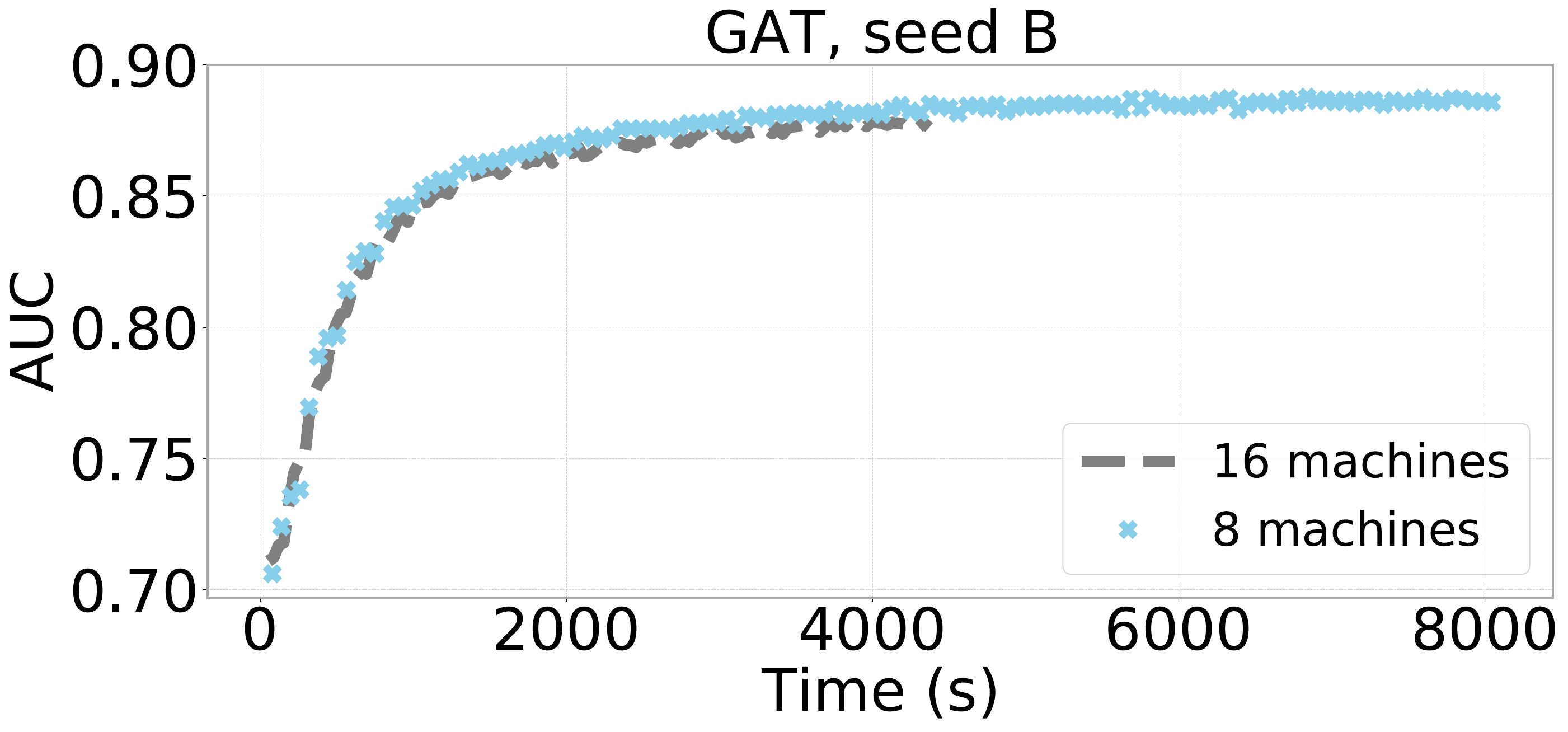} 
    &
    \includegraphics[width=.311\linewidth]{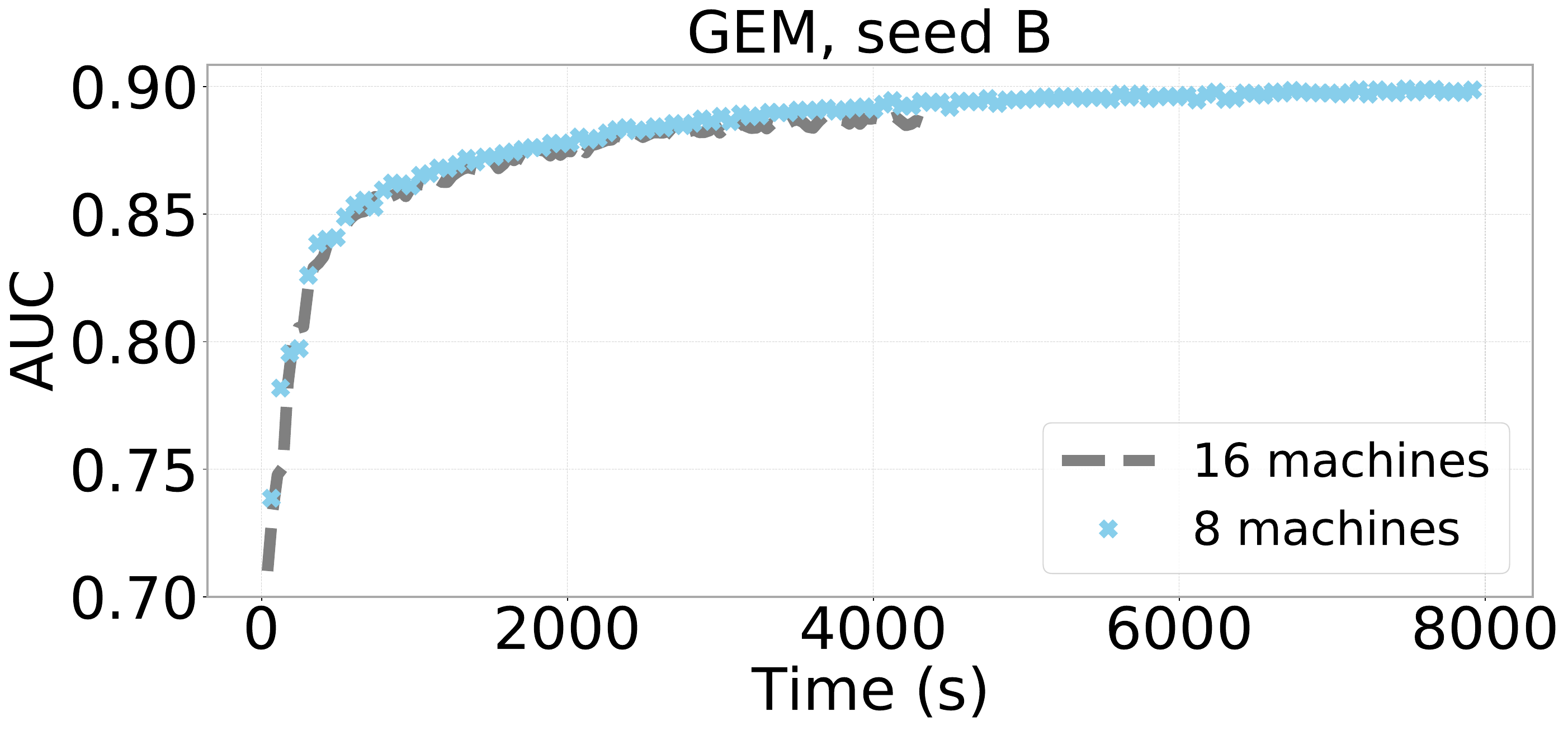}
    &
    \includegraphics[width=.311\linewidth]{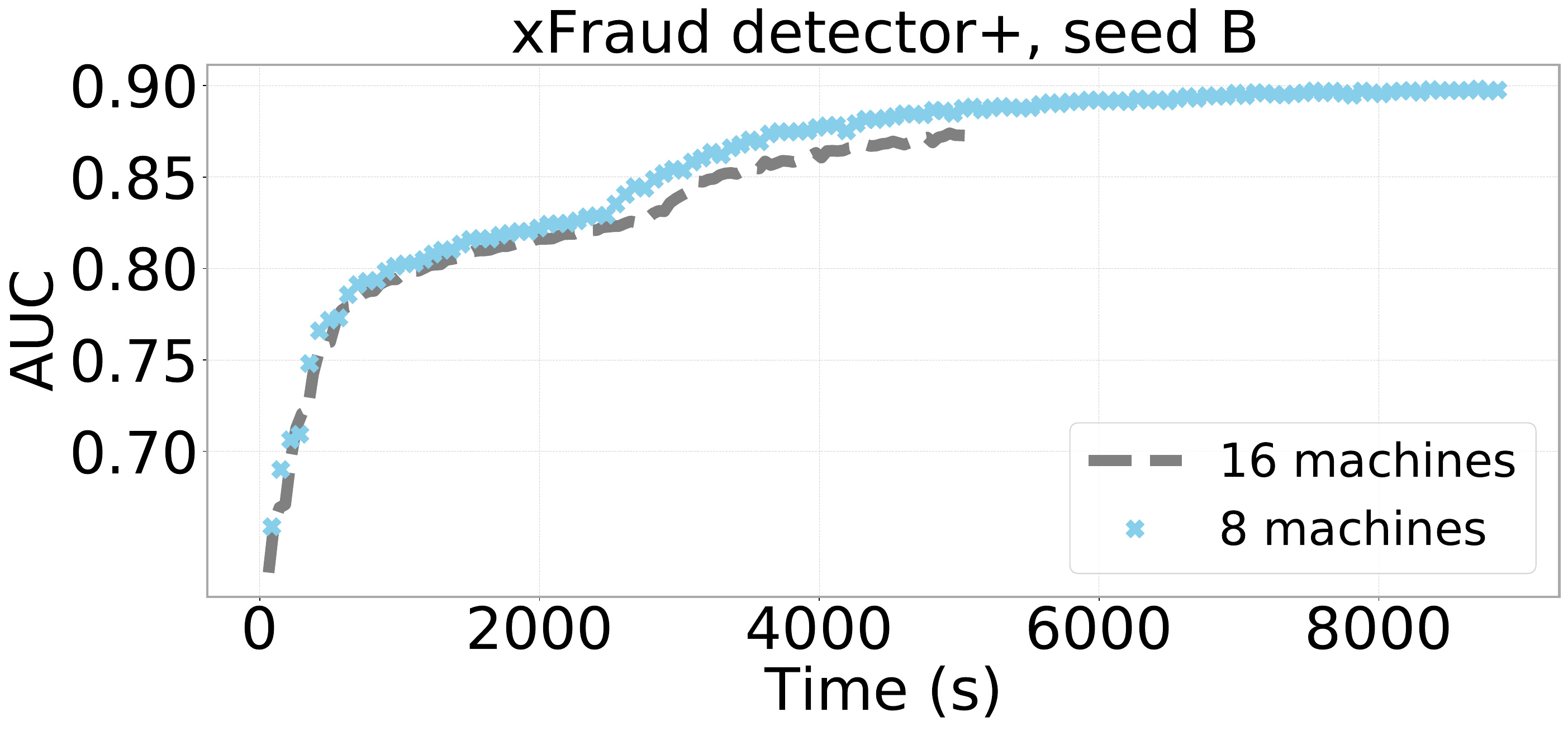}\\

\end{tabular}
    \caption{Convergence of distributed training (8 vs.~16 machines) on GAT, GEM, and xFraud detector+, measured by AUC (on two seeds).}
    \label{fig:convergence-all}
\end{figure*}

\paragraph{\textbf{{Convergence on eBay-xlarge}}}
{In this paragraph, we discuss the model convergence under different settings. 
In Figure~\ref{fig:convergence-all}, the models trained on 16 machines do not converge faster than that on 8 machines. Also, the final AUCs on 16 machines are worse compared to those on 8 machines. 
In this case, training using 8 machines is a better choice across different models using the current graph partitioning strategy as described in Sec.~\ref{sec:scalable-xfraud}.}


\section{Experimental details of a modified GNNExplainer} \label{app:gnnexplainer}
We implement GNNExplainer using the trained xFraud detector, transaction features, node index (of node-to-explain), edge indices, edge types, and node types as input. We obtain node feature masks of the subgraph as output, as well as the edge masks. 
The hyperparameters of GNNExplainer are: $\text{\textit{epochs}} = 100$, $\text{\textit{lr}} = 0.01$, $\beta_{\text{edge\_size}} = 0.005$, $\beta_{\text{node\_feature\_size}} = 1$, $\beta_{\text{edge\_entropy}} = 1$, $\beta_{\text{node\_feature\_size}} = 0.1$. The xFraud detector is not retrained during the explanation process. 

We extend the vanilla GNNExplainer \cite{ying2019gnnexplainer} so that it generates node feature masks for all the nodes, which enables a node-level feature explanation to a prediction. {Learning the feature importance of all node features was not possible with the vanilla GNNExplainer.} The training of explainer is initialized with a random edge mask $1 \times |\mathcal{E}|$ and a random node feature mask $|\mathcal{V}| \times F$ ($F$ the size of node features), as well as four random coefficients of edge size $\beta_{\text{es}}$, node feature size $\beta_{\text{nfs}}$, edge entropy $\beta_{\text{ee}}$, and node feature entropy $\beta_{\text{nfe}}$. It then takes the node index of node-to-explain, transaction features, node types, edge types and edge indices as input. The explainer takes the weights trained in the detector and uses only the detector model in the evaluation mode, as Figure~\ref{fig:detector-explainer} demonstrates (right). The loss function of explainer has to include edge entropy and node feature entropy so that the explainer knows which important nodes, node features and edges to attend in predicting the current node. The goal of explainer loss is to jointly minimize the detector loss with the entropy of node features and edges. 

To compute the explainer loss, we first compute the edge mask and node feature mask of the subgraph $S$ in $G$ that contributes to the node-to-explain. We use $M_{E_S}$ to denote the edge mask in the subgraph $S$ ($S$ a subgraph of $G$), and we have $M_{E_S} = \sigma (E_S)$, $E_S$ a random initialization of edge parameters. Likewise, we denote the node feature mask as $M_{V_S} = \sigma (V_S)$, $V_S$ a random initialization of node feature parameters. Now we calculate the explainer loss in three parts, detector loss (eq.~\ref{eq:detector-loss}), edge entropy (eq.~\ref{eq:edge-entropy}), and node feature entropy (eq.~\ref{eq:node-feature-entropy}). We finally sum them up to form the explainer loss:

\begin{equation}\label{eq:detector-loss}
    \sum_i C_i \log(S_i),
\end{equation}
\begin{equation}\label{eq:edge-entropy}
\resizebox{0.8\linewidth}{!}{
$\begin{aligned}
    & \beta_{\text{es}}\sum M_{E_S} + \\ 
    & \beta_{\text{ee}}\frac{\sum\Big(-M_{E_S}\log(M_{E_S}+\epsilon)-(1-M_{E_S})\log(1-M_{E_S}+\epsilon)\Big)}{|\mathcal{V_E}|},
\end{aligned}$
}
\end{equation}

\begin{equation}\label{eq:node-feature-entropy}
\resizebox{0.8\linewidth}{!}{
$
\begin{aligned}
    & \beta_{\text{nfs}}\frac{\sum M_{V_S}}{|\mathcal{V_S}|} + \\
    & \beta_{\text{nfe}}\frac{\sum\Big(-M_{V_S}\log(M_{V_S}+\epsilon)-(1-M_{V_S})\log(1-M_{V_S}+\epsilon)\Big)}{|\mathcal{V_S}|},
\end{aligned}$
}
\end{equation}
where $i \in \{\text{labeled transactions}\}$, $C_i$ is the true label of a transaction, $S_i$ the output of the \textit{softmax} layer in DNN, $|\mathcal{V_S}|$ and $|\mathcal{E_S}|$ the number of nodes and edges in the subgraph $S$, and $\epsilon$ a small constant to prevent $\log(0)$. 
By back propagation, the new loss is used to compute the new network weights in explainer and thus the network has the capacity to learn the subgraph nodes and their features that most importantly contribute to certain predictions made by the detector. 

\section{Annotation, node importance, random vs. GNNExplainer}\label{app:human-annotation}
\paragraph{\textbf{Annotation protocol.}}Now we introduce our annotation protocol of human ground truth. The goal of annotation is to assign a node importance score to each node in terms of how important the node is when the seed node prediction is made. We do not perform annotations on the edge level because the annotators only have access to node-level information (e.g., node features and representations). We have five expert annotators to ensure that each node is at least annotated by two annotators. There are three scales of node importance the annotators can assign: 0, 1, 2 for low, medium, and high importance, respectively. Then, we calculate the average inter-annotator agreement (IAA) score of annotator pairs. The average IAA score is 0.532 among 10 pairs of annotators, with the highest IAA 0.773 and lowest 0.314. But how do we know if our annotations are of high quality? Assuming the human annotators were randomly assigning scores, we employ random integer generators that assign 0, 1, 2 to replace all the annotations by human, and we calculate the average IAA score among the random annotators. Also, we repeat this random experiment 10 times and calculate the mean IAA score. The random IAA score is -0.006 which is significantly worse compared with 0.525, the IAA score of human annotators.

\begin{table}[!t]
\centering
    \caption{Top\textit{k} hit rate between explainer edge weights and human annotations: GNNExplainer vs.~ random.} 
\label{tab:eval}
\resizebox{\linewidth}{!}{\begin{tabular}{cccccc}
\toprule
\textbf{Top\textit{k} hit rate}     & \textbf{Top5} & \textbf{Top10} & \textbf{Top15} & \textbf{Top20} & \textbf{Top25} \\
\midrule
Random            & 0.13 & 0.45  & 0.60  & 0.70  & 0.79  \\
GNNExplainer              & 0.45 & 0.69  & 0.82  & 0.90  & 0.92  \\
$\Delta$(GNNExplainer-Random) & 0.32 & 0.24  & 0.22  & 0.20  & 0.13 \\
\bottomrule
\end{tabular}
}

\end{table}

\paragraph{\textbf{Node importance score.}} After we have obtained the human annotations by five annotators, we first calculate the average node importance score for a node $v$ by
\(\frac{\sum annotation_i}{5}\), where an annotator \(i \in [1,2,3,4,5]\). We then use these node importance scores to compute the edge importance scores. There are three strategies to calculate the edge importance based on its incident nodes --- by averaging (``avg") or summing (``sum") the node importance scores, or by taking the minimal node importance score (``min"). {There is no theoretical or empirical evidence of which aggregation method is the best, we therefore have run all the aggregation methods and select the one that performs the best.}
\paragraph{\textbf{Top\textit{k} hit rate.}}
In Figure \ref{fig:human-vs-explainer}, we demonstrate two graphs with edge importance scores and edge weights, respectively. We have to choose a meaningful set of \textit{k}. Note that the average count of total edges across the communities is 81.56, among which the average count of edges with the largest edge importance is 20.90. This means that as long as $k$ is smaller than or around 20, the top\textit{k} hit rate reflects a meaningful agreement score between the explainer and the annotators in ranking the important edges. 
\paragraph{\textbf{Random vs.~GNNExplainer.}}
Another caveat of the top$k$ edge selection is that we need to break the tie among the largest edge importance scores. To tackle this, we randomly draw 100 samples when selecting the top\textit{k} edges based on the highest importance score. Then, we take the average hit rate of these 100 draws. We have also run experiments with 10,000 draws, which renders similar results reported in Table \ref{tab:eval}. Here, we report the hit rates computed as the mean of random 100 draws. 

In Table \ref{tab:eval} we report the top\textit{k} hit rate with $k \in [5,10,15,20,25]$. Our hit rate is already 45\% when inspecting the top 5 edges and increases as $k$ increases. To draw the conclusion that the explainer agrees strongly with our human annotators, we look at a random baseline. The random baseline assumes that the explainer randomly assigns edge weights during prediction. We let a random number generator assign edge weights to each edge and we compute the top\textit{k} hit rate between random edge weights and edge importance scores. We repeat the random experiment 10 times, and finally obtain the average top\textit{k} hit rate from these 10 random draws. The random baseline is also reported in Table \ref{tab:eval}. Taking the difference between our hit rate and the random one, we see the largest discrepancy is at the top 5 hit rate and gradually drops as \textit{k} increases. This demonstrates the strong agreement between the actual explainer weights and our human annotators.\footnote{Comparing results from three aggregation methods (``sum" vs.~``min" vs.~``avg"), there is no significant difference in the distribution of our hit rates and $\Delta(GNNExplainer-Random)$. We hence report only the results of ``avg". In addition, there is no substantial difference between $\Delta(GNNExplainer-Random)$ of communities labeled 1 and 0. {For more details of the results on ``min" and ``sum" and their performances on communities labeled 1 and 0.}} 

\paragraph{\textbf{{Comparing aggregation methods in analyzing explainer \\ weights}}}
\label{app:explainer-agg}
{There are three different aggregation strategies, ranging from node importance (human) to edge importance (human). 
In Tables~\ref{tab:eval-avg}, \ref{tab:eval-min} and \ref{tab:eval-sum}, there is no substantial difference in results generated by averaging, summing, or minimizing the weights. Also, in communities with a benign (0)/fraudulent (1) node-to-predict, there is no substantial difference. 
We also ran a paired-\textit{t}-test to see if the distributions between these two types of communities are identical: the test results do not allow us to reject this null hypothesis on a significance level of 0.05. Therefore, in the follow-up experiments of hybrid explainer, we always use ``averaging" as the aggregation method to compute the edge importance score (human).}

\begin{table}[!h]
\centering
    \caption{{Top\textit{k} hit rate between explainer edge weights (``avg") and human annotations: GNNExplainer vs.~ random. $c1$: a community labeled 1, $c0$: a community labeled 0.} }
\label{tab:eval-avg}

\resizebox{\linewidth}{!}{\begin{tabular}{cccccc}
\toprule
\textbf{Top\textit{k} hit rate}     & \textbf{Top5} & \textbf{Top10} & \textbf{Top15} & \textbf{Top20} & \textbf{Top25} \\
\midrule
Random            & 0.13 & 0.45  & 0.60  & 0.70  & 0.79  \\
GNNExplainer              & 0.45 & 0.69  & 0.82  & 0.90  & 0.92  \\
$\Delta$(GNNExplainer-Random) & 0.32 & 0.24  & 0.22  & 0.20  & 0.13 \\ \hline
$Random_{c0}$ & 0.31 & 0.20 & 0.21 & 0.20 & 0.15 \\
$GNNExplainer_{c0}$ & 0.47 & 0.77 & 0.90 & 0.97 & 1.00 \\
$\Delta_{c0}$(GNNExplainer-Random) & 0.16 & 0.57 & 0.69 & 0.27 & 0.85 \\  \hline
$Random_{c1}$ & 0.33 & 0.17 & 0.23 & 0.21 & 0.11 \\
$GNNExplainer_{c1}$ &0.41& 0.59 & 0.72 & 0.81 & 0.82 \\
$\Delta_{c1}$(GNNExplainer-Random) & 0.08 & 0.42 & 0.49 & 0.60 & 0.71 \\ 
\bottomrule
\end{tabular}}

\end{table}

\begin{table}[!h]
\centering
    \caption{{Top\textit{k} hit rate between explainer edge weights (``min") and human annotations: GNNExplainer vs.~ random. $c1$: a community labeled 1, $c0$: a community labeled 0.} }
\label{tab:eval-min}

\resizebox{\linewidth}{!}{\begin{tabular}{cccccc}
\toprule
\textbf{Top\textit{k} hit rate}     & \textbf{Top5} & \textbf{Top10} & \textbf{Top15} & \textbf{Top20} & \textbf{Top25} \\
\midrule
Random                      & 0.13 & 0.45  & 0.60  & 0.70  & 0.79  \\
GNNExplainer                        & 0.45 & 0.69  & 0.82  & 0.90 & 0.92  \\
$\Delta$(GNNExplainer-Random)         & 0.32 & 0.24  & 0.22  & 0.20  & 0.13 \\ \hline
$Random_{c0}$               & 0.16 & 0.48 & 0.69 & 0.77 & 0.85 \\
$GNNExplainer_{c0}$                 & 0.47 & 0.77 & 0.90 & 0.97 & 1.00 \\
$\Delta_{c0}$(GNNExplainer-Random)  & 0.31 & 0.29 & 0.21 & 0.20 & 0.15 \\ \hline
$Random_{c1}$               & 0.08 & 0.42 & 0.49 & 0.60 & 0.71 \\
$GNNExplainer_{c1}$                 & 0.41 & 0.59 & 0.72 & 0.81 & 0.82 \\
$\Delta_{c1}$(GNNExplainer-Random)  & 0.33 & 0.17 & 0.23 & 0.21 & 0.11 \\ 
\bottomrule
\end{tabular}}
\end{table}

\begin{table}[!h]
\centering
    \caption{{Top\textit{k} hit rate between explainer edge weights (``sum") and human annotations: GNNExplainer vs.~ random. $c1$: a community labeled 1, $c0$: a community labeled 0.} }
\label{tab:eval-sum}

\resizebox{\linewidth}{!}{\begin{tabular}{cccccc}
\toprule
\textbf{Top\textit{k} hit rate}     & \textbf{Top5} & \textbf{Top10} & \textbf{Top15} & \textbf{Top20} & \textbf{Top25} \\
\midrule
Random            & 0.10 & 0.38  & 0.54  & 0.63  & 0.77  \\
GNNExplainer              & 0.38 & 0.63  & 0.80  & 0.88 & 0.90  \\
$\Delta$(GNNExplainer-Random) & 0.28 & 0.25  & 0.26  & 0.25  & 0.13 \\ \hline
$Random_{c0}$ & 0.17 & 0.40 & 0.61 & 0.70 & 0.85 \\
$GNNExplainer_{c0}$ & 0.41 & 0.70 & 0.90 & 0.97 & 1.00 \\
$\Delta_{c0}$(GNNExplainer-Random) & 0.24 & 0.30 & 0.29 & 0.27 & 0.15 \\ \hline
$Random_{c1}$              & 0.03 & 0.37 & 0.45 & 0.56 & 0.67 \\
$GNNExplainer_{c1}$                & 0.36 & 0.55 & 0.68 & 0.77 & 0.78 \\
$\Delta_{c1}$(GNNExplainer-Random) & 0.33 & 0.18 & 0.23 & 0.21 & 0.11 \\ 
\bottomrule
\end{tabular}}
\end{table}

\section{{More Details on the Hybrid Explainer}}
\paragraph{\textbf{{Centrality measures as edge weights.}}}\label{app:hybrid-explainer}
{We use the following two ways to compute edge weights using centrality measures:}
\begin{enumerate}[wide=0pt]
    \item {Compute the edge weights based on the node annotations using edge centrality measures such as {\bf edge betweenness} and \textbf{edge load centrality}. 
    Edge betweenness measures the number of the shortest paths between vertices that contain the edge, normalized by $\frac{2}{n(n-1)}$ in an undirected graph.  Edge load counts the number of the shortest paths which cross each edge. Then, we calculate the top$k$ hit rate between the edge centrality scores and the human annotations.} 
    \item {Compute the edge weights based on the node annotations after transforming the original graph into a line graph. We compute various types of \textbf{node centralities} (e.g., closeness, eigenvector centrality, degree, etc.) measures in the line graph. }
\end{enumerate}

{We report the results of representing edge weights with 13 centrality measures in Table~\ref{tab:centraliy-vs-explainer}. We compute the centrality measures using the \textit{networkx}\footnote{\url{https://networkx.org/documentation/stable/reference/algorithms/centrality.html} (last accessed: Sep 1, 2021).} package in Python. }

\paragraph{\textbf{{Observations on comparing explainer weights and centrality measures. }}}
{We observe that the hit rates computed by explainer weights $H(e)$ and by different centrality measures $H(c)$ are close. The similar hit rates show that GNNExplainer and centrality measures both manage to learn the most important edges when filtering the fraudulent transactions. For each rank, we mark the highest hit rate with bold: there is not a centrality measure that consistently outperforms the remaining measures. Even among various centrality measures, the $\Delta(H(e) - H(c))$ on different ranks vary. As $k$ increases, the differences between the hit rates decrease. }

{We pick the best four centrality measures according to the hit rate, i.e., edge betweenness, degree, edge load, closeness and harmonic. If we examine the differences of hit rate $\Delta(H(e) - H(c))$ across communities in Figure~\ref{fig:barplot-diffH}, we notice that there is no clear winner between $H(e)$ and $H(c)$ in detecting the most important edges when comparing to human annotations. More importantly, a trade-off between $H(e)$ and $H(c)$ can be observed. 
This motivates us to learn a hybrid explainer using both explainer weights and centrality measures. }



\paragraph{\textbf{{A hybrid learner that incorporates both explainer weights and centrality measures.}}}

{We formulate the learning problem as follows. 
First, we learn two coefficients --- centrality coefficient $A$ and explainer coefficient $B$, which maximizes $avg(H(h))$ in the training communities via $avg(A * w(c) + B * w(e))$, where $w(c)$ denotes the centrality measures, and $w(e)$ the explainer weights. Since there are 41 communities, we take the first 21 communities as the training set and the last 20 as the test set. Then, we calculate $avg(H(h))$ in the test communities using the $A$ and $B$ learned over the training set. 
There are various techniques to optimize $avg(A * w(c) + B * w(e))$ and search for the optimal $A$ and $B$, which maximize the average $H(h)$ in the training set. 
By taking a centrality measure that generates the best $H(c)_{Top5}$ in Table~\ref{tab:centraliy-vs-explainer} (edge betweenness), we have run these three sets of experiments to optimize $A$ and $B$:}

\begin{enumerate}[wide=0pt]
    
    \item {Fit polynomial functions where we find the best feature degree to maximize the average $H(h)$ in the training communities; }
    
    \item {Grid searches for $A$, where $B = 1-A$; }
    
    \item {Train Ridge linear regressions on $avg(A * w(c) + B * w(e))$, where we also optimize the regularization hyperparameter $\alpha$. }
    
\end{enumerate}
    
    {We run (1) to find the best degree $d$ among \{1, 2, ..., 9\} and obtain $d=1$ being the best fit, i.e., a linear combination of $A * w(c) + B * w(e)$. 
    We illustrate the results of hybrid explainers constructed by (2) and (3) in Table~\ref{tab:hybrid-explainer-app}. 
    For the grid search in (2), we tune $A$ in \{0.00, 0.01, ..., 1.00\}, where $A$ maximizes the average hit rate in the training communities. 
    For (3), the best $\alpha$ is 0.99 among \{0.01, 0.02, ..., 0.99\}, with $A = -0.1097, B = 0.1064$. If we compare the results of $H(h)$, $H(c)$, and $H(e)$ at all ranks, the hybrid learner has achieved at least a result as good as $H(c)$ if not better. } 
    
    {The results in Table~\ref{tab:centraliy-vs-explainer} and Figure~\ref{fig:barplot-diffH} validate that our proposed hybrid explainer can achieve a trade-off between the centrality measures and explainer weights. 
    Both measures agree with human annotators, and we can leverage weights learned by explainer and centrality to construct a better explainer. 
    This hybrid explainer incorporates both the topological features of the community and the message passing learned via GNNExplainer.}
    
   { Moreover, we have also noticed that the centrality measures assign identical weights to many edges in the community.
    In contrast, the vanilla GNNExplainer always assigns a total order of weights to the edges, with the edges closer to the node-to-explain getting the highest weights. GNNExplainer offers a local explanation of the prediction, while centrality measures attend to the global structure of the community. 
    Intuitively, by combining the best of both worlds, the hybrid explainer is beneficial since it considers both local and global structures of communities.}


\paragraph{\textbf{{More Results on the Hybrid Explainer}.}} {We report the hit rate of top 5 edges across GNNExplainer, edge betweenness centrality, and the hybrid explainer trained using Ridge and grid search in Table~\ref{tab:hybrid-explainer-app}.}


\begin{table}[!t]
\centering
\caption{{Top$k$ hit rate in the train and test communities by the hybrid explainer. $A$ is the coefficient of centrality weights (edge betweenness) in the hybrid explainer $A * w(c) + B * w(e)$, where $B = 1-A$.}} 
\label{tab:hybrid-explainer-app}

\resizebox{\linewidth}{!}{
\begin{tabular}{lccccccccc}
\toprule
\textbf{$H(\_)$}  &
\multicolumn{2}{c}{\begin{tabular}[c]{@{}c@{}}Edge \\ betweenness $H(c)$\end{tabular}}
 & 
 \multicolumn{2}{c}{\begin{tabular}[c]{@{}c@{}}GNNExplainer \\  $H(e)$\end{tabular}} &
 \multicolumn{2}{c}{\begin{tabular}[c]{@{}c@{}}Hybrid \\  (ridge) $H(h)$\end{tabular}} &  \multicolumn{3}{c}{\begin{tabular}[c]{@{}c@{}}Hybrid \\  (grid) $H(h)$\end{tabular}}                                           \\ \cline{2-10}
                           & \multicolumn{1}{c}{Train} & \multicolumn{1}{c}{Test} & \multicolumn{1}{c}{Train} & \multicolumn{1}{c}{Test} & \multicolumn{1}{c}{Train} & \multicolumn{1}{c}{Test} & \multicolumn{1}{c}{Train} & \multicolumn{1}{c}{Test} & \multicolumn{1}{c}{$A_{Train}$} \\ \hline
Top5                       & 0.4817                    & 0.45540                  & 0.44257                   & 0.44800                  & 0.44648                   & 0.44890                  & 0.45971                   & \textbf{0.45550}         & 0.75                  \\
Top10                      & 0.6581                    & 0.78175                  & 0.61210                   & 0.77580                  & 0.60767                   & \textbf{0.81115}                  & 0.61881                   & 0.78700                  & 0.94                  \\
Top15                      & 0.7497                    & 0.87763                  & 0.75981                   & 0.88473                  & 0.74838                   & 0.89210                 & 0.76375                   & \textbf{0.89410}         & 0.91                  \\
Top20                      & 0.8459                    & 0.96205                  & 0.84126                   & 0.95840                  & 0.83505                   & 0.96198                  & 0.85595                   & \textbf{0.96275}         & 1.00                  \\
Top25                      & 0.8813                    & \textbf{0.96616}                  & 0.88350                   & 0.95954                  & 0.88286                   & 0.96614                  & 0.88379                   & 0.96614                  & 0.64                  \\
Top30                      & 0.8868                    & 0.96705                  & 0.88778                   & 0.95893                  & 0.88790                   & \textbf{0.96780}                  & 0.88938                   & \textbf{0.96780}         & 0.65                  \\
Top35                      & 0.8920                    & \textbf{0.95780}                 & 0.89388                   & 0.94731                  & 0.89339                   & 0.95761                  & 0.89444                   & 0.95771                  & 0.51                  \\
Top40                      & 0.8976                    & 0.93659                  & 0.89860                   & 0.92608                  & 0.89893                   & 0.93673                  & 0.89917                   & \textbf{0.93764}         & 0.68                  \\
Top45                      & 0.9026                    & 0.91598                  & 0.90244                   & 0.90678                  & 0.90443                   & \textbf{0.91608}                 & 0.90472                   & 0.91607                  & 0.68                  \\
\bottomrule
\end{tabular}
}

\end{table}

\section{Case studies using Hybrid Explainer}\label{app:more-cases-hybridexplainer}

{In this paper, a fraudulent transaction is marked as 1 (positive) and a benign one as 0. 
When we report true positives (TP), true negatives (TN), false positives (FP), and false negatives (FN), these four cases correspond to the following detection scenarios: "TP" -- fraudulent transactions being flagged correctly,  "TN" -- benign transactions being flagged correctly, "FN" -- we have missed frauds in detection, "FP" -- benign transactions wrongly flagged as fraud. }

\subsection{More true positive cases}\label{app:tp-cases}
\begin{figure*}[!]
  \begin{tabular}  {ccc}
    \includegraphics[width=.35\linewidth]{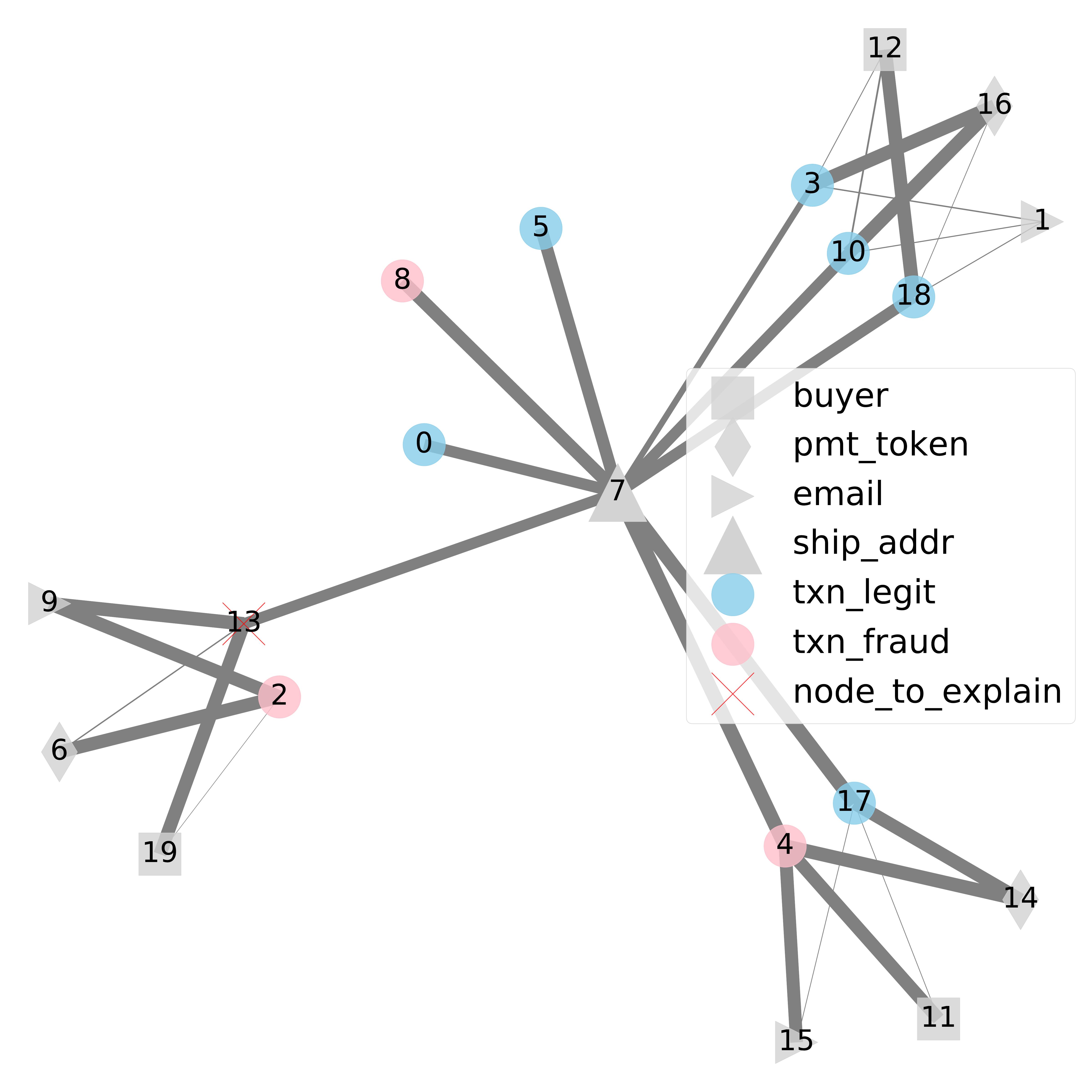}  &    \includegraphics[width=.35\linewidth]{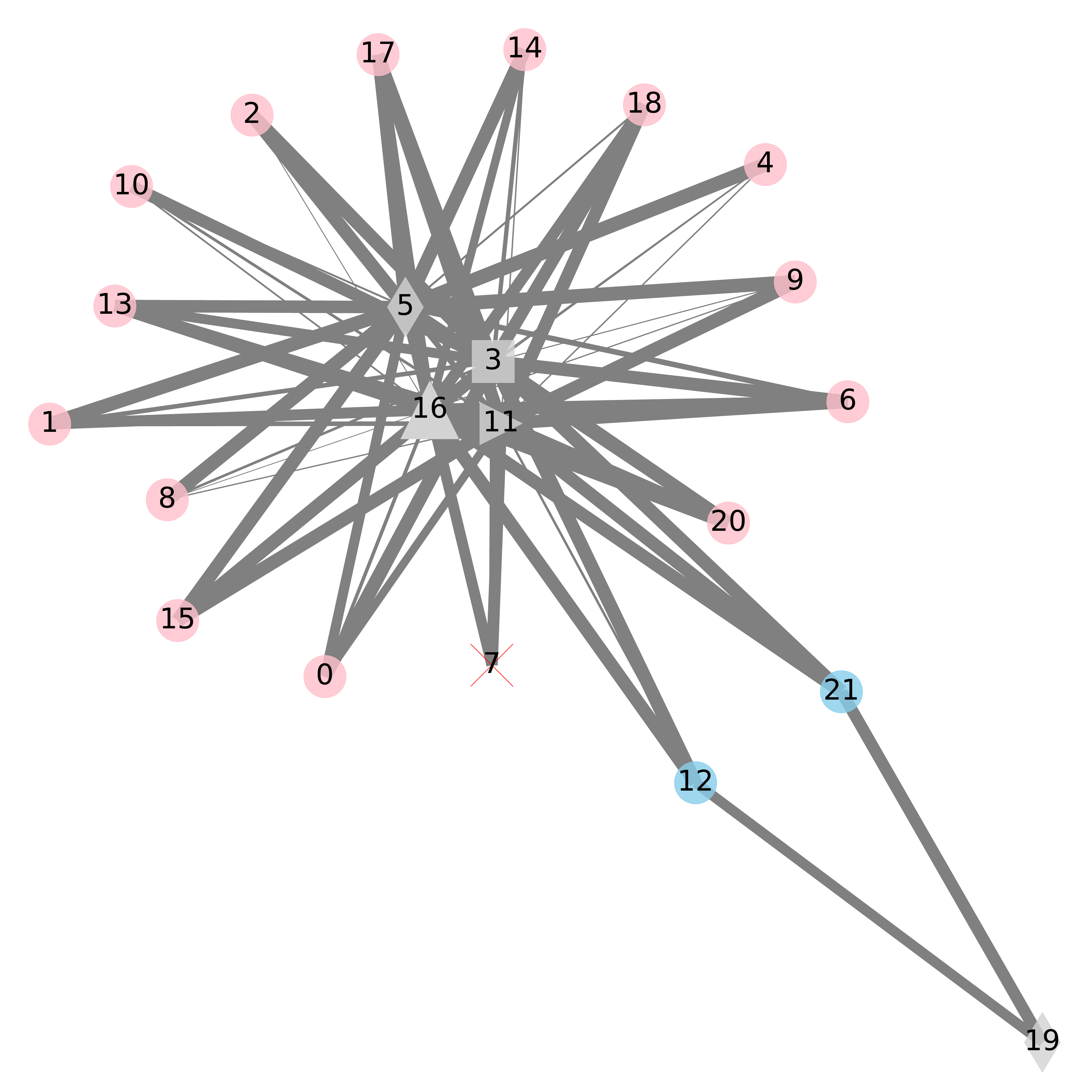} \\
    (a) & (d) \\[6pt]

   \includegraphics[width=.35\linewidth]{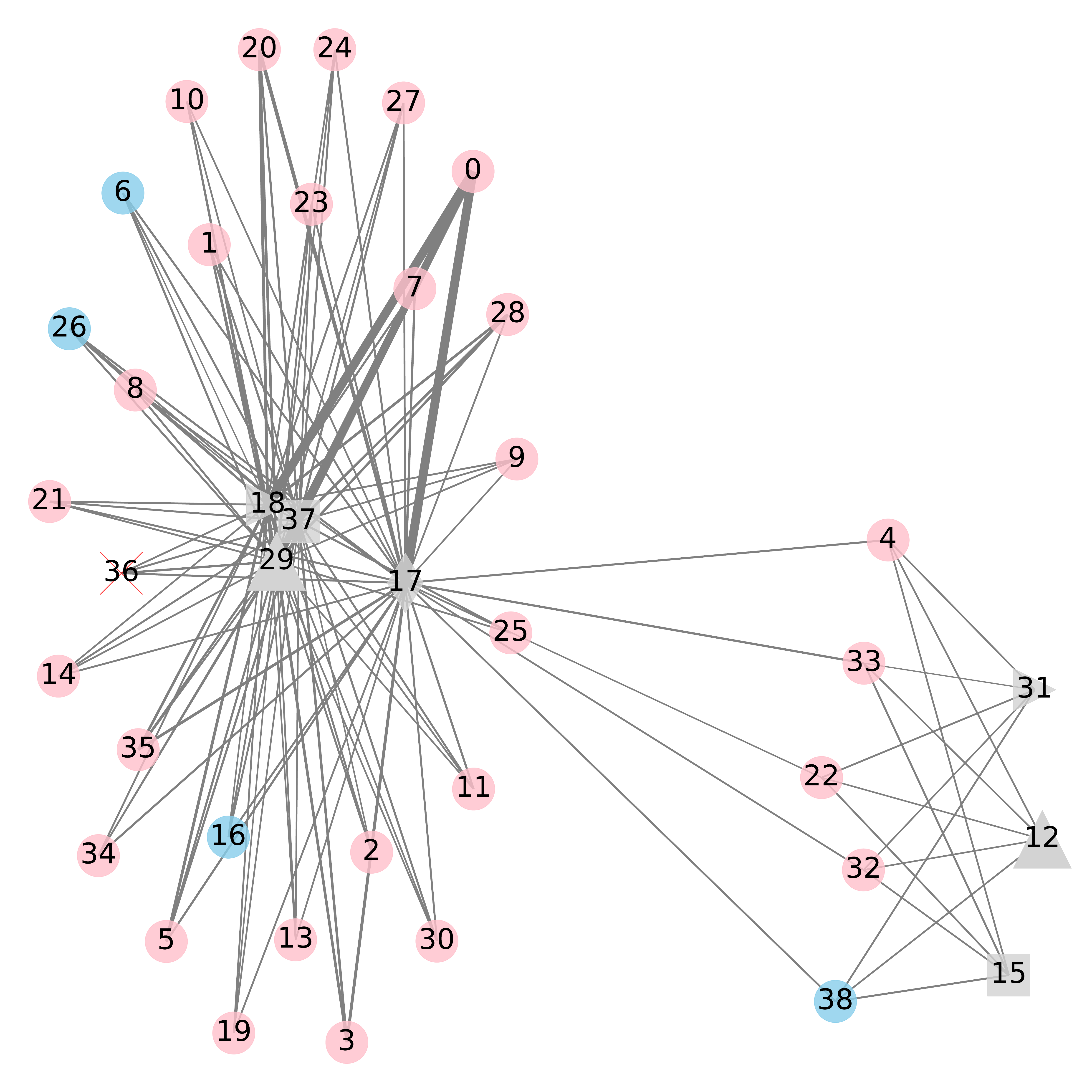} &     \includegraphics[width=.35\linewidth]{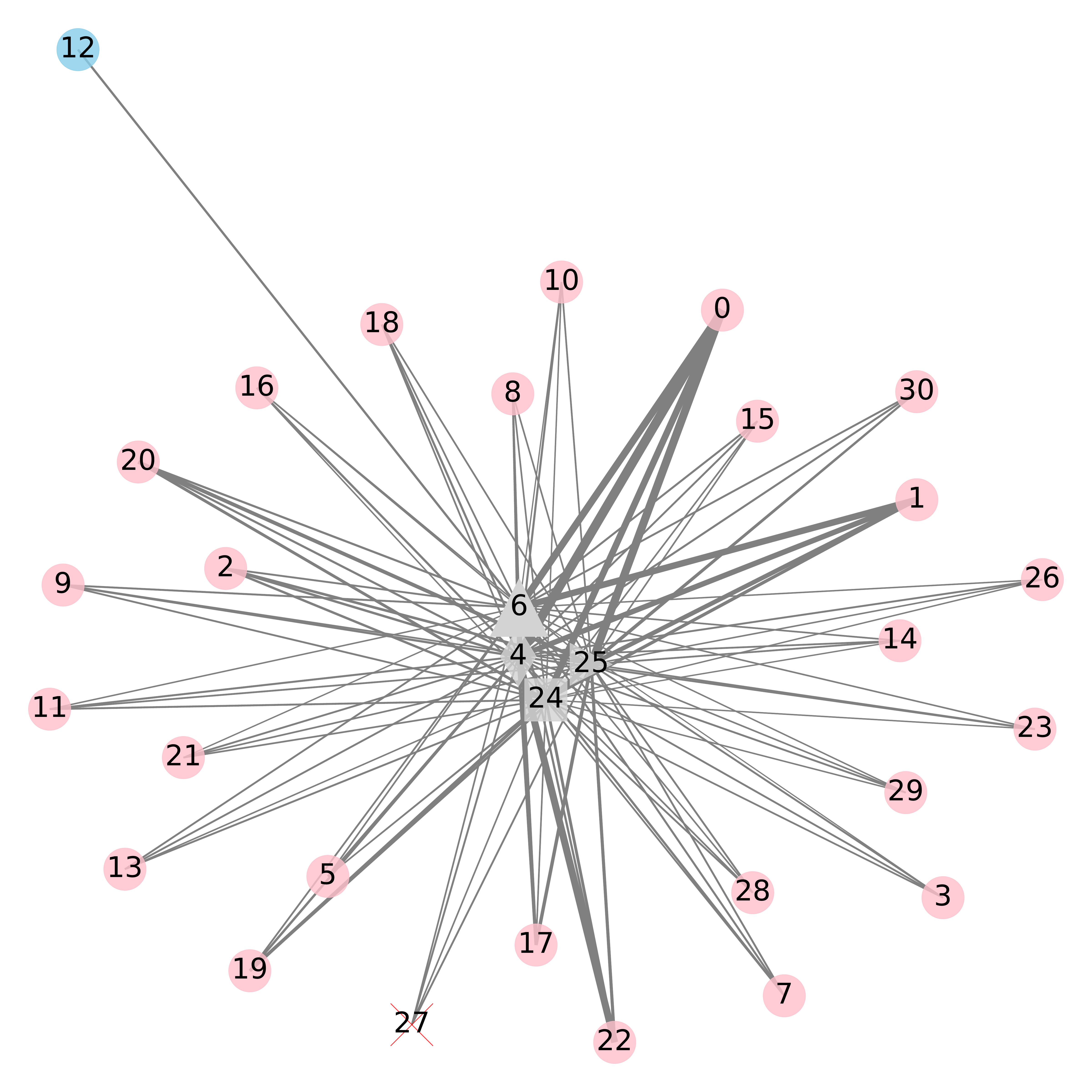} \\
    (b)  & (e) \\[6pt]

    \includegraphics[width=.35\linewidth]{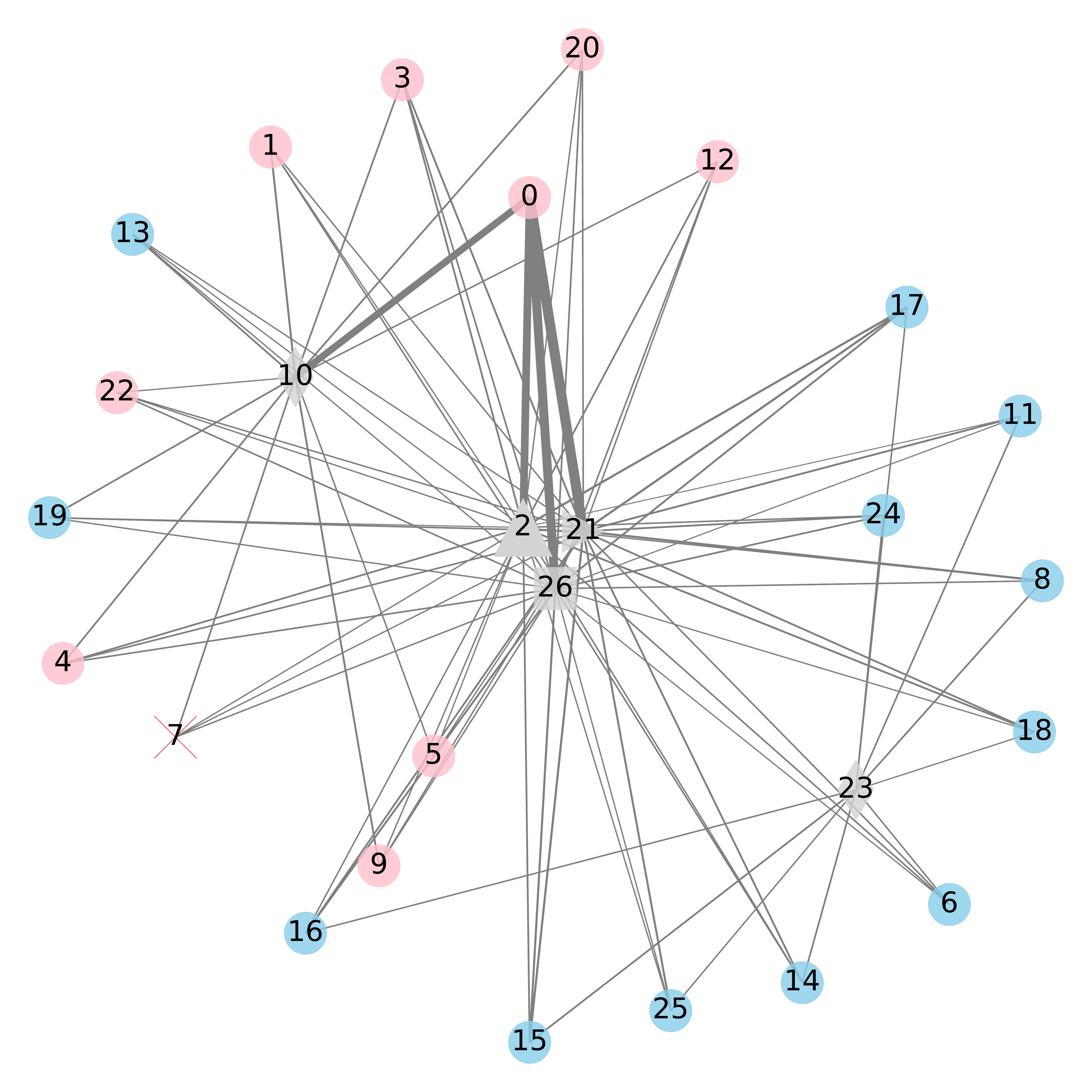} &     \includegraphics[width=.35\linewidth]{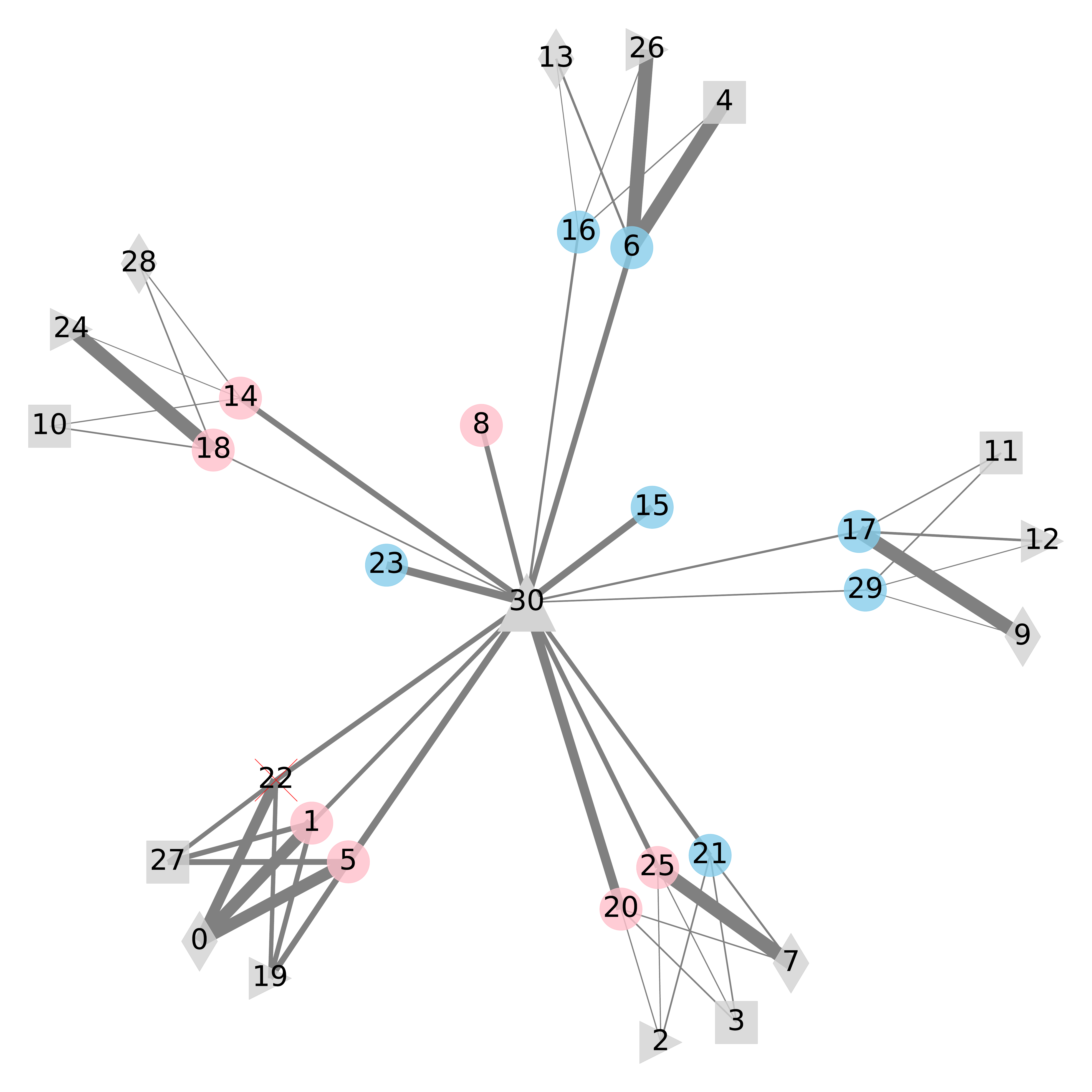}\\
        (c) & (f) \\[6pt]
\end{tabular}
    \caption{{Case studies using hybrid learner weights: true positives (TP).}}
    \label{fig:tp-more}
\end{figure*}

{Except the case we show in Figure~\ref{fig:tp-more}, we present six more TP cases in our 41 communities, visualized using weights learned in the hybrid explainer. We see from the visualization in (a), (c), and (f), xFraud does a good job in detecting suspicious transaction, even if the frequencies of fraudulent/benign cases are close. xFraud hybrid explainer not only manages to learn from the incident edges of the node-to-predicts, but also learns the path where the risk propagates. For instance, in (b), it learns that transaction 0 connects linking entities 17 (payment token) and 18 (email) with node 37 (buyer), which influences strongly the prediction of node 36 to be fraud. This type of risk propagation is useful in learning the embedding representations of linking entities. Similar to (b), we also observe a clear risk propagation path in (e). In case (d), we observe a typical fraudulent cluster which could be due to the fact that a valid payment token (node 19) was used at the beginning to gain the trust of the platform, then the defaulter conducts many fraudulent transactions. It could also be that the defaulter has hacked one payment token (node 5) and uses that payment token to conduct many transactions in a short time. In either case, the xFraud can assist BU to identify this kind of fraud cluster. }
\begin{figure*}[!t]
\includegraphics[width=0.9\textwidth]{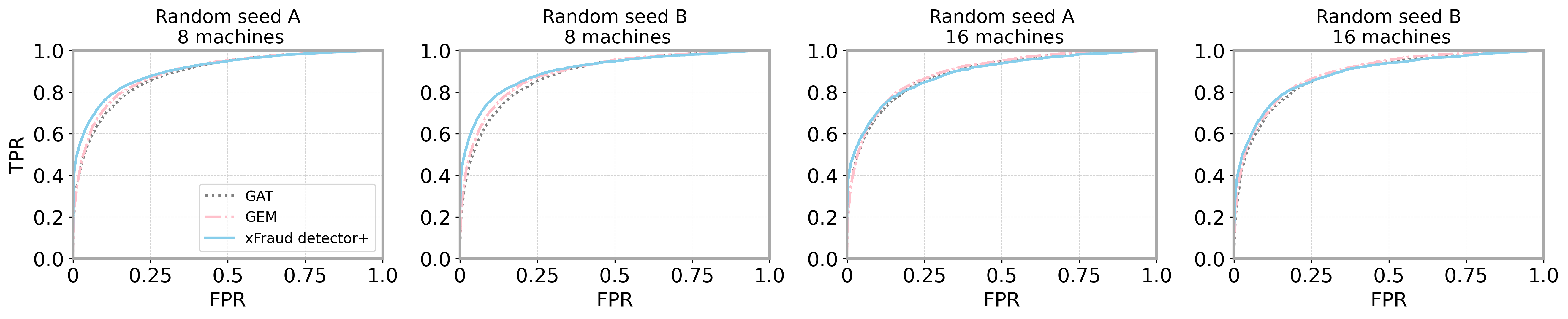}
\caption{{ROC curves using different settings (seeds and \# machines) on \textit{eBay-xlarge}.}}
    \label{fig:roc-curve-xlarge}
\end{figure*}

\subsection{More case studies: false positives and false negatives} \label{app:fp-fn-cases}

{We investigate several case studies for the conditions "FP" and "FN" and share our insights. }

\begin{figure*}[!]
  
  \begin{tabular}  {ccc}
    \includegraphics[width=.35\linewidth]{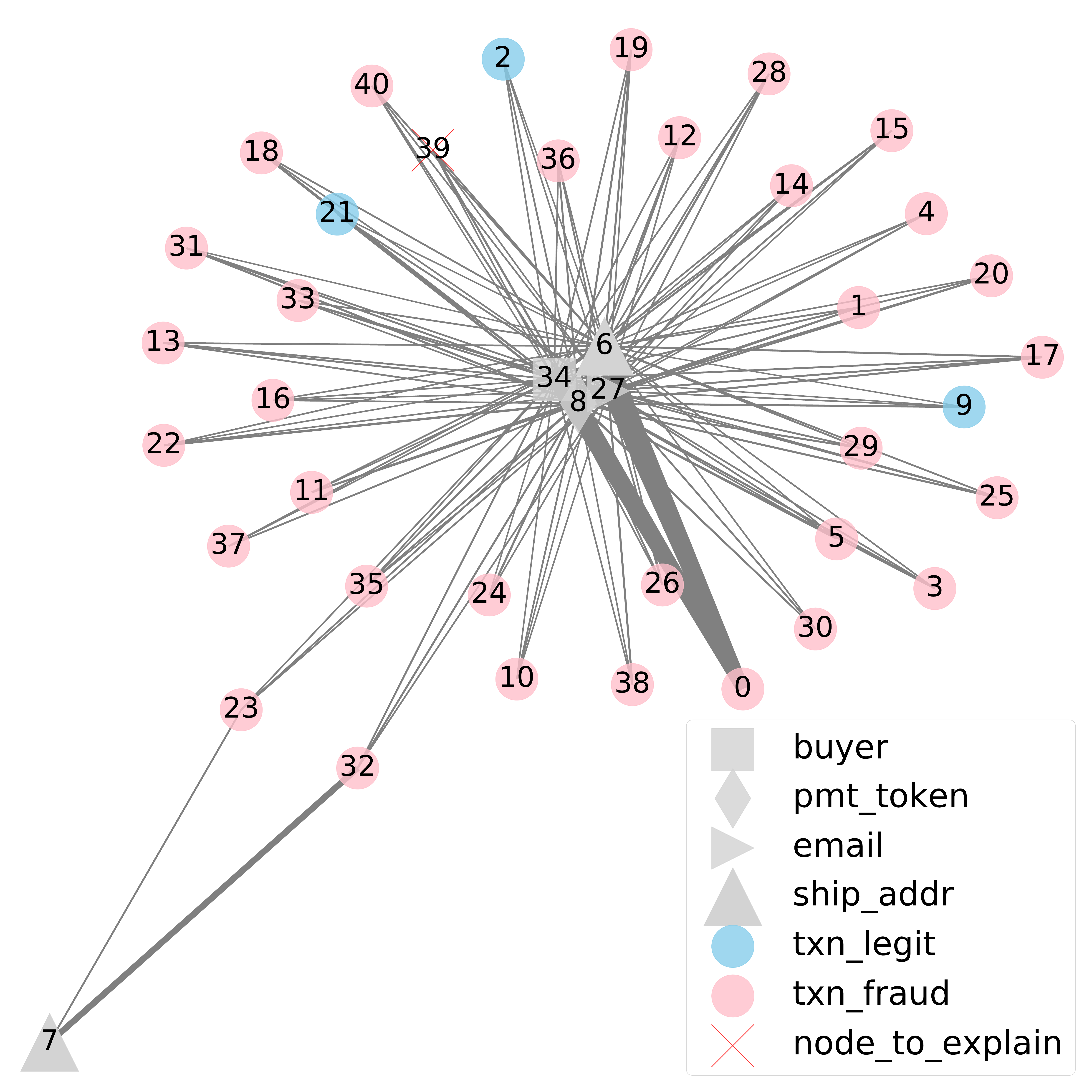}  &    \includegraphics[width=.35\linewidth]{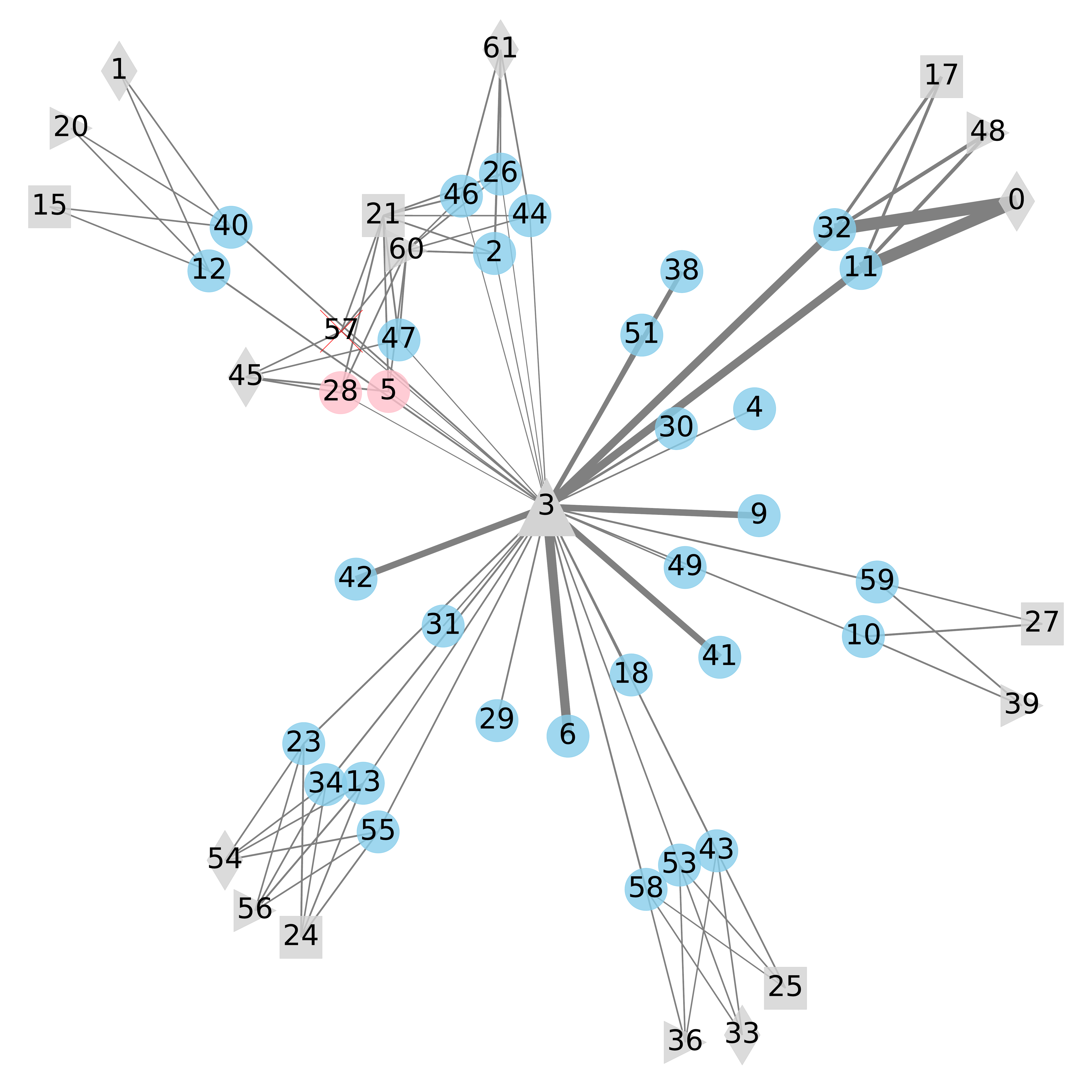} \\
    (a) FP: benign $\rightarrow$ fraudulent (case 1) & (d) FN: fraudulent $\rightarrow$ benign (case 1)\\[6pt]

   \includegraphics[width=.35\linewidth]{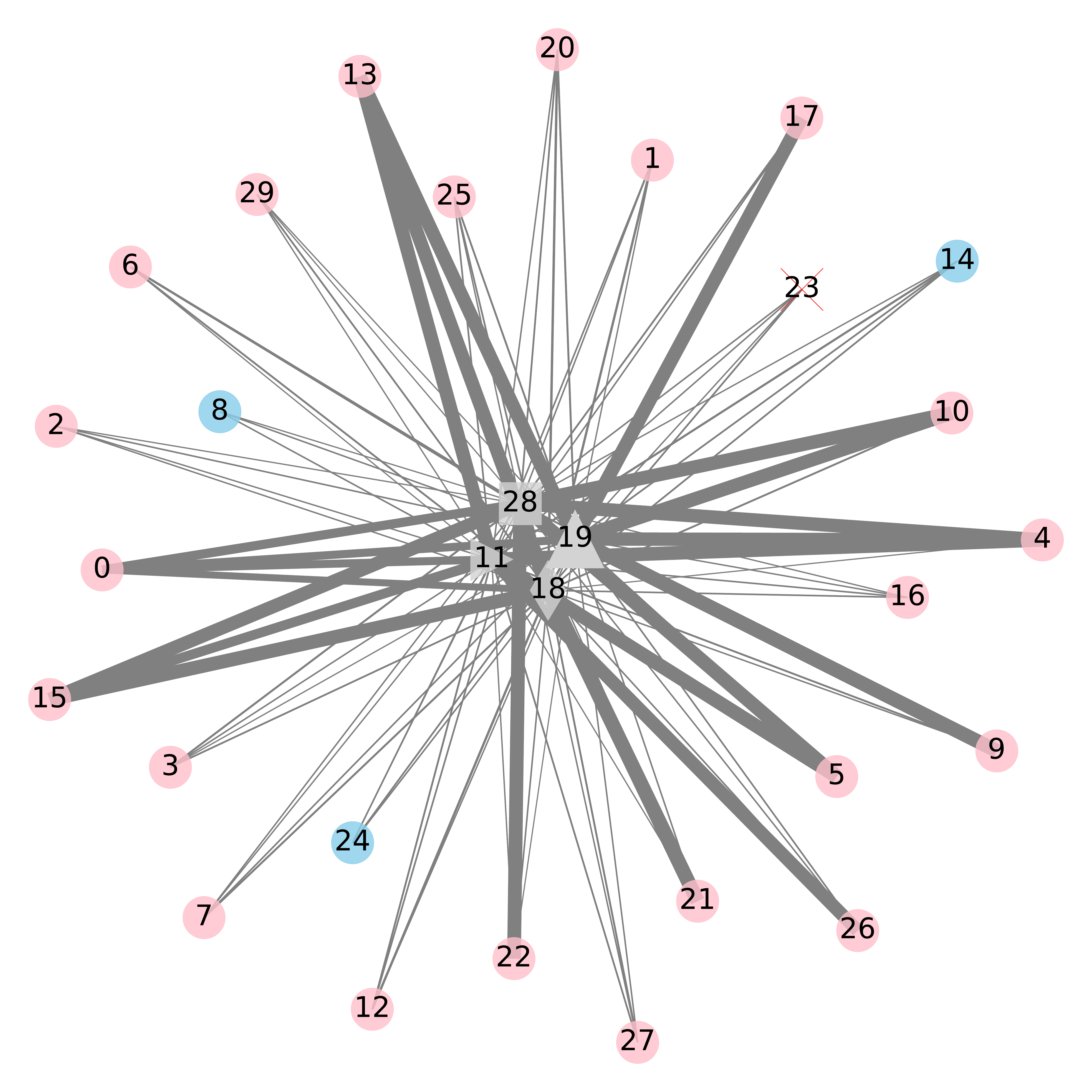} &     \includegraphics[width=.35\linewidth]{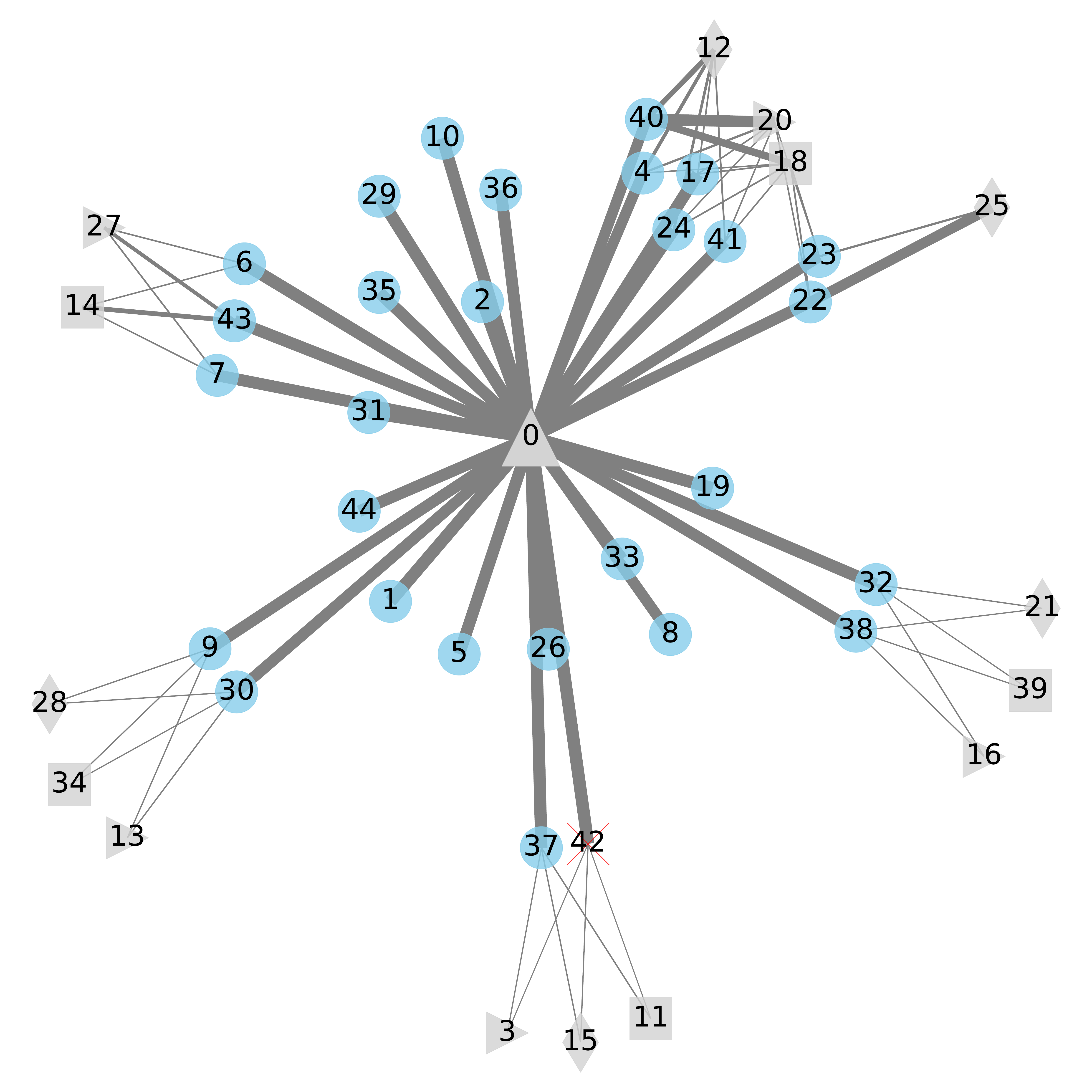}\\
    (b) FP: benign $\rightarrow$ fraudulent (case 2) & (e) FN: fraudulent $\rightarrow$ benign (case 2)\\[6pt]

    \includegraphics[width=.35\linewidth]{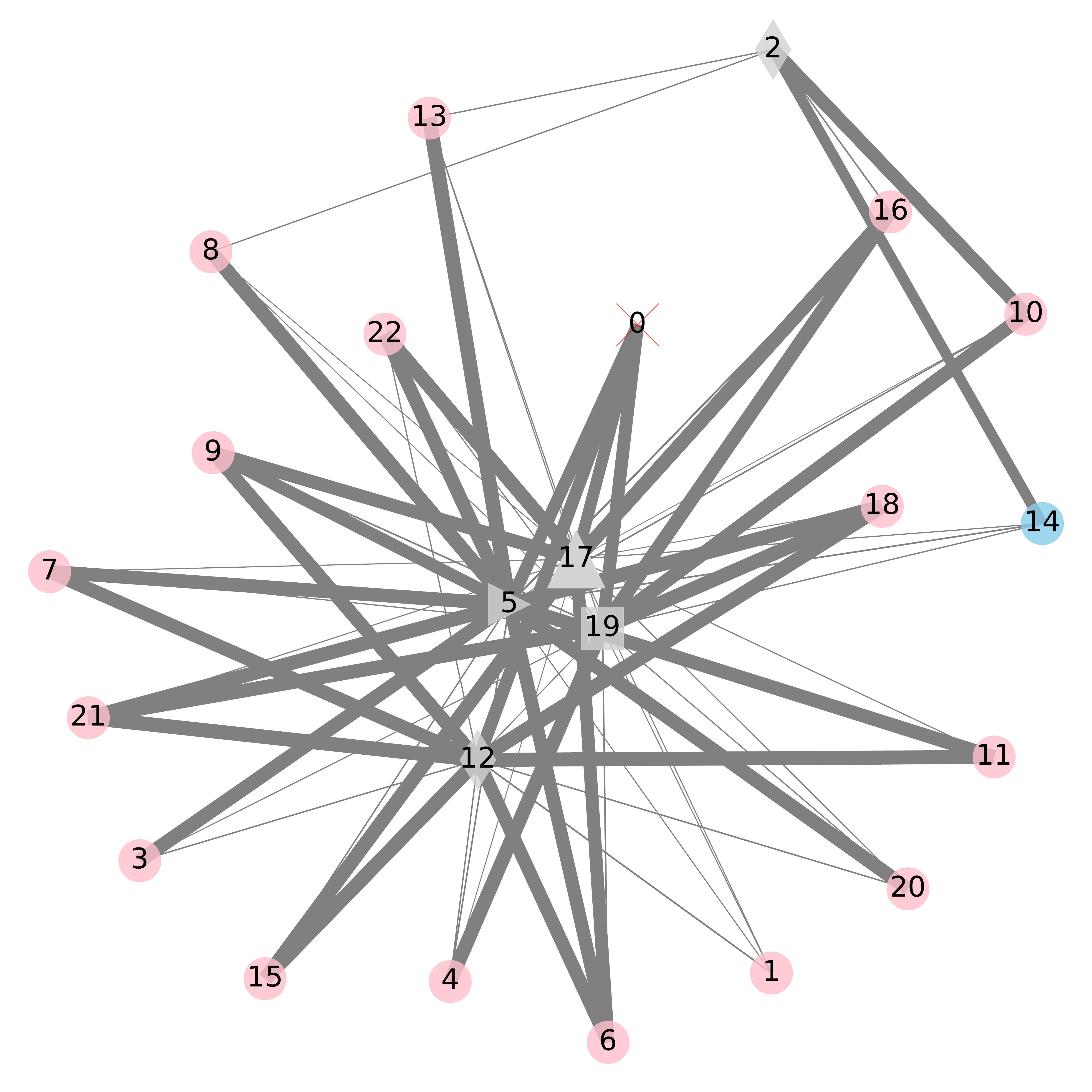} &     \includegraphics[width=.35\linewidth]{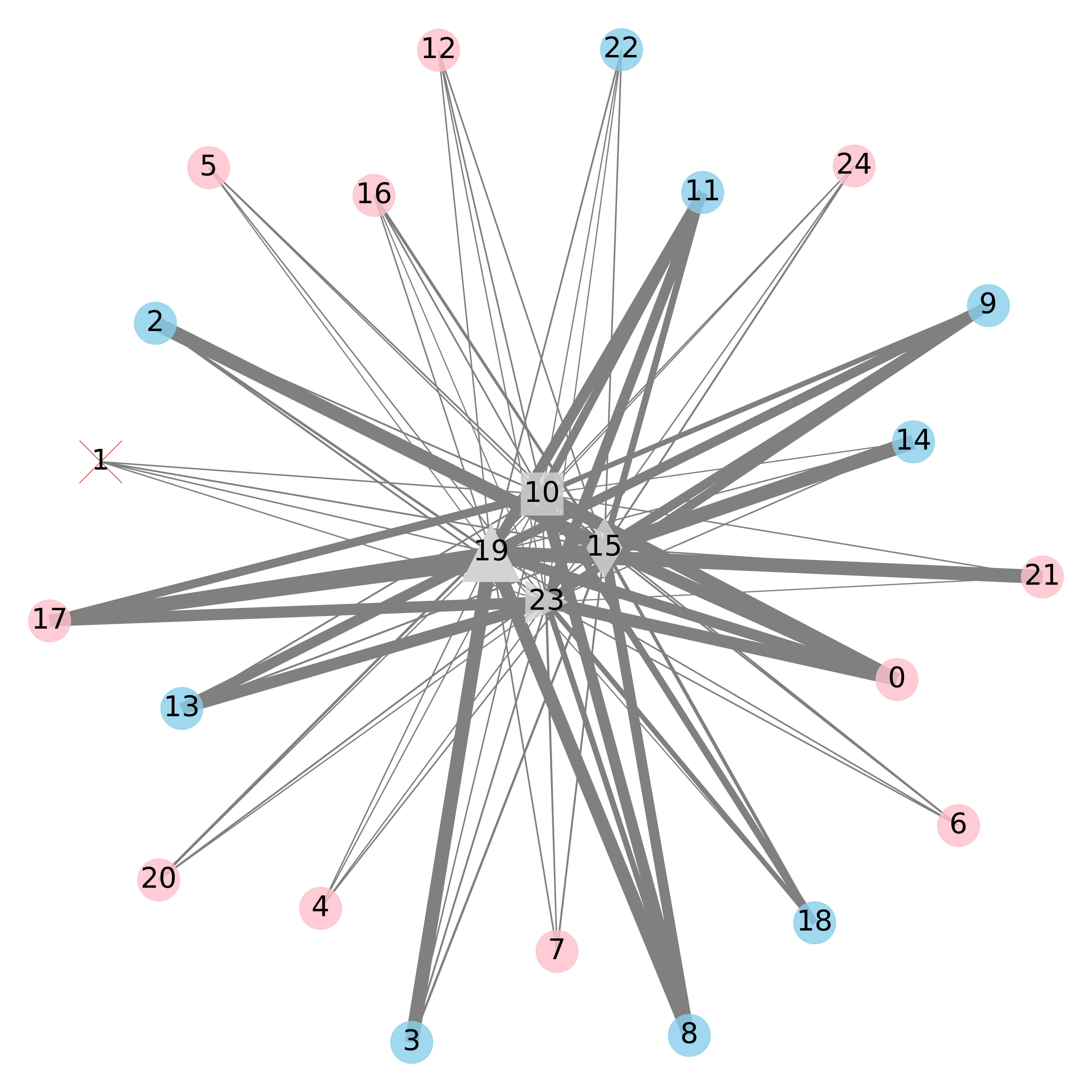}\\
        (c) FP: benign $\rightarrow$ fraudulent (case 3) & (f) FN: fraudulent $\rightarrow$ benign (case 3)\\[6pt]
\end{tabular}
    \caption{{Case studies using hybrid learner weights: false positives (a,b,c on the left) and false negatives (d,e,f on the right).}}
    \label{fig:fp-fn-cases}
\end{figure*}

\paragraph{\textbf{{False positive (FP): benign $\rightarrow$ fraud.}}}
{In the community illustrated in Figure~\ref{fig:fp-fn-cases} (a), we have one buyer/email address/payment token and two shipping addresses. One shipping address (node 6) is much more frequently used than the other one. Note that this heavily used shipping address (node 6) is linked to many fraudulent transactions. This makes the prediction of seed node 39 prone to fraudulent (probability = 0.972 in xFraud detector+), because this transaction is directly linked to seed 6, which is considered highly suspicious by learning from its neighbors in the community. The most important edges in generating the prediction are marked in bold (0-6, 0-8). However, BU reports that this account is not fraudulent after investigation. Similarly, we have more fraudulent cases in the communities than the benign ones in the cases shown in Figure~\ref{fig:fp-fn-cases} (b) and \ref{fig:fp-fn-cases} (c).
}

\paragraph{\textbf{{False negative (FN): fraud $\rightarrow$ benign.}}}
{In a community like Figures~\ref{fig:fp-fn-cases} (d), where multiple buyers/payment tokens/email/shipping addresses are involved, we do not observe that one shipping address is used more frequently in the transaction logs. However, there are much more benign cases in the neighborhood, therefore xFraud detector+ has problems identifying a fraudulent transaction which could be due to a sudden attack, or a delay of user chargebacks. In Figure~\ref{fig:fp-fn-cases} (f) we see a simple (single-buyer) community with more fraudulent transactions than benign ones in the neighborhood. However, since the important (thick) edges are connected to more benign transactions than frauds (benign: 2, 14, 8, 9, 13, 11; fraud: 21, 17, 0), the node 1 is classified wrongly as benign.}

{To empirically investigate the case studies, we show in Table~\ref{tab:conf-matrix} that the different conditions are indeed correlated with communities types. We do not have false positives (benign wrongly recognized as fraudulent) in the complex communities, and xFraud does a better job in communities with more than one buyer.}

{To summarize the \textbf{FP: benign $\rightarrow$ fraud} cases, in all the FP cases we have (6 out of 6), there is only one buyer in the community, one linking entity of each type (i.e., one email, one payment token, one shipping address). Even if there are two linking entities, one is always more frequently used than the other one, and the seed is linked to the entity that is more frequently used. Whenever there is more fraudulent transactions connected to that linking entity, the benign node-to-predict is classified as fraudulent; vice versa for false negatives (2 out of 8). This is preemptive in risk prevention; the user might expect some short-term interruption of the service while being alert and notified by the platform that a fraudulent transaction might be ongoing. }

{In summary, the \textbf{FN: fraud $\rightarrow$ benign} cases show more variability, and our goal is to avoid as much as possible the false negatives. 
In many of the cases (6 out of 8), we see more buyers and linking entities (such as payment tokens and emails) in the community -- we have higher false negatives in the complex communities. It is hard for xFraud detector+ to distinguish signals sent via various buyers and transactions. It could be deliberate ring attacks with accounts “cultivated” over a long time, or could be a one-time attack of one account using an unknown device.} 

\begin{table}[t!]
\centering
\caption{{Confusion matrix of TP, TN, FP, and FN in simple and complex communities. The sample is the 41 communities we present in Sec.~\ref{sec:explainer-eval-quantitative}. We show the frequency and percentage of one condition in one type of communities in the bracket, where a simple community has only one buyer, and a complex one has more than one buyer. "TP": fraudulent transactions, "TN": benign transactions, "FN": missed frauds in detection, "FP": benign flagged as fraud. }}
\label{tab:conf-matrix}
\resizebox{\linewidth}{!}{
\begin{tabular}{rr|rr}
\toprule
\multicolumn{2}{c|}{\textbf{Simple communities (16, 100\%)}} & \multicolumn{2}{c}{\textbf{Complex communities (25, 100\%)}} \\
\midrule
\textbf{FP} (6, 37.5\%)  & \textbf{TP}   (4, 25\%)  & \textbf{FP}    (0, 0\%)     & \textbf{TP}  (6, 24\%)   \\
\textbf{FN}  (2, 12.5\%)  & \textbf{TN}   (4, 25\%)  & \textbf{FN}   (6, 24\%)  & \textbf{TN}   (13, 52\%) \\
\bottomrule
\end{tabular}}
\end{table}

\subsection{System limitation and possible causes.}

{As we see in most of the cases, the misclassification of frauds is related to the ratio of fraudulent/benign cases in the \textit{k}-hop neighborhood of the node-to-predict. It is therefore important to enforce a graph partition constraint of benign/fraudulent-ratio, so that the prediction is not strongly influenced by the frequency of cases. This goes into the direction we discussed before in the scalability experiments of \textit{eBay-xlarge}: how to scale out training in a distributed setting on a heterogeneous graph.}

{Although our system can flag some guest checkouts that are linked to suspicious entities, it still remains a difficult use case to capture with both ML and GNN methods. Guest checkout allows users to make purchases without logging and to stay anonymous. Our xFraud detector is designed for graph, therefore it requires purchases linked to some existing and nontrivial entities. Image a case where the user chooses a guest checkout to make a purchase. The email is newly registered, so it is not linked to any existing entities. The ip address is from a public places such as cafe or gas station. The payment token is a credit card which have not been used in our checkout system. The shipping address is a public transshipment warehouse. In such a case, none of the trivial entities can be linked by this purchase, so that our xFraud detector can hardly retrieve any useful information to make accurate predictions on the purchase.}

\section{More evaluation metrics of experiments on \textbf{eBay-large}} \label{app:more-metrics-eval}

{\subsection{True positive rate (TPR)/Recall, true negative rate (TNR), false positive rate (FPR), false negative rate (FNR)}}\label{app:tpr-tnr-fpr-fnr}

{We document the TPR, TNR, FPR, and FNR on the test set of \textit{eBay-xlarge} in Tables~\ref{tab:tpr-fnr-tnr-fpr-xlarge-part1}, ~\ref{tab:tpr-fnr-tnr-fpr-xlarge-part2}, and ~\ref{tab:tpr-fnr-tnr-fpr-xlarge-part3}. Note that FPR = 1 - TNR and FNR = 1 - TPR. TPR is also known as recall. } 

{\subsection{Precision and recall at various thresholds}}\label{app:precision-recall-xlarge}

{In Tables~\ref{tab:pr-rc-xlarge-part1}, ~\ref{tab:pr-rc-xlarge-part2}, and ~\ref{tab:pr-rc-xlarge-part3}, we list the precision and recall scores calculated at various thresholds on the test set of \textit{eBay-xlarge}. }

{\subsection{\textit{ROC curve on \textit{eBay-large}.}}}
{Besides the ROC curve in Figure~\ref{fig:roc-curve-xlarge-0.1} showing FPR <0.1 in flagging frauds on \textit{eBay-large}, we also present the ROC curves in the full range of FPR. 
The xFraud detector+ outperforms all baselines in ROC-AUC in the two experiments on 8 and 16 machines, respectively. }

{\subsection{Discussion: Performance analysis of xFraud in production}} \label{app:eval-discussion}
{Reading from Tables~\ref{tab:pr-rc-xlarge-part2} and \ref{tab:pr-rc-xlarge-part3}, on the whole test set of \textit{eBay-xlarge} with 4.33\% fraud rate, our method 
\begin{itemize} [wide=0pt]
    \item with threshold 0.983 has precision = 0.9822 and recall = 0.1091,
    \item with threshold 0.977 has precision = 0.9539 and recall = 0.2063, 
    \item with threshold 0.960 has precision = 0.9217 and recall = 0.2930.
\end{itemize}

 }
{As we discussed in Appendix~\ref{app:dataset}, we filtered and sampled the transaction labels in the \textit{eBay-xlarge} dataset before applying GNN models. These operations lead to the change of fraud rate in the transaction labels in each step:}
{
\begin{center}
\text{(1) the original data stream (0.016\% fraud)} \\ 

[\text{filtered by rules}] \bigg\Downarrow \\

\text {(2) the filtered data stream (0.043\% fraud)} \\

[\text{sampled all frauds \& 1\% benign}] \bigg\Downarrow\\

\text{(3) the sampled data stream (4.33\% fraud)}   .
\end{center}
}

{In this paper, we report all numbers on step (3). \textit{How would the algorithm perform on the dataset on step (2)?} We further report precision on step (2). In this case, a 0.98 precision on (3) corresponds to 0.32 precision on (2), with ~0.1 recall; and a 0.95 precision on (3) corresponds to 0.16 precision on (2), with ~0.2 recall.

A 0.32 precision means that for ~3 fraud candidates investigated by the business unit, 1 will be a real fraud. This, with 0.1 recall is acceptable in our scenario. Even a 0.16 precision, with 0.2 recall, is reasonable -- for every ~6 fraud candidates investigated by the business unit, 1 will be a real fraud.


When our model is deployed into production, there 
will be more considerations beyond this single GNN model --- one has to consider the entire pipeline, including data preprocessing, feature engineering, and downsampling.
For example, in many eBay applications, less risky transactions are filtered out by rule-based or ML-based strategies before using more complicated models such as GNN; and GEM~\cite{liu2018heterogeneous} has also pre-filtered isolated transactions. }

{\subsection{Potential production scenario using xFraud}}
\label{app:production}

{We have used an industrial-scale dataset \textit{eBay-xlarge} using seven months of transaction data produced at eBay. xFraud now has a model trained on historical data. What is also practical is an incremental setting of online model training and fine-tuning. For example, we can perform model updates on a daily basis to ensure the model timeliness. 
We can also build models in an incremental setting. For instance, we use the data from the $T-1$ week (or month) to flag the transactions produced in the $T$ week (or month).
However, there is a caveat here: many fraudulent behaviors are planned for the long-term attacks. The defaulters would "cultivate" a set of accounts for many months to gain the trust of the platform and then launch attacks. Hence, it is crucial to use both historical and up-to-date data to train our system and combine their predictions in production.
After our delicate optimization for the distributed xFraud system,
xFraud now can efficiently train a GNN model on a  billion-scale graph dataset
(38s per epoch on \textit{eBay-xlarge} using 16 machines).
With xFraud's powerful processing capabilities,
we can train a new model within several hours in eBay, and hence
support real production scenarios.
Besides, xFraud can also accelerate the inference task significantly in a distributed setting (less than 0.1 seconds to process a batch of 640 nodes using a single machine).
To help downstream tasks in a production scenario,
the inference scores for the transactions are attached to the corresponding entities, represented as risk scores for the upcoming transaction events. 
These scores are treated as features for downstream risk models and as variables for the rule-based defense systems.}

\begin{table*}[!]
\centering
\caption{{TPR, FNR, FPR, and TNR on \textit{eBay-xlarge} by varying thresholds from 0.1 to 0.9 on the prediction scores. \\ FNR = 1 - TPR. The higher TPR is, the better; the lower FNR, the better. 
\\ FPR = 1 - TNR. The higher TNR is, the better; the lower FPR, the better. }}
\label{tab:tpr-fnr-tnr-fpr-xlarge-part1}
\resizebox{0.85\linewidth}{!}{
\begin{tabular}{cccccccccccc}
\toprule
\multirow{2}{*}{\textbf{Model}}   & \multirow{2}{*}{\textbf{\# of machines}} & \multirow{2}{*}{\textbf{Seed}} & \multicolumn{9}{c}{\textbf{TPR@threshold}}                                                                                           \\ \cline{4-12}
                                  &                                          &                                & \textbf{0.1} & \textbf{0.2} & \textbf{0.3} & \textbf{0.4} & \textbf{0.5} & \textbf{0.6} & \textbf{0.7} & \textbf{0.8} & \textbf{0.9} \\ \hline
\multirow{4}{*}{GAT}              & \multirow{2}{*}{8}                       & A                              & 0.7379       & 0.6718       & 0.6264       & 0.5891       & 0.5535       & 0.5158       & 0.4730        & 0.4187       & 0.3170       \\
                                  &                                          & B                              & 0.7443 &	0.6664 &	0.6184 &	0.5787 &	0.5391 &	0.5030 &	0.4577 &	0.3972 &	0.2780
     \\ \cline{2-12}
                                  & \multirow{2}{*}{16}                      & A                              & 0.7439       & 0.6789       & 0.6311       & 0.5923       & 0.5483       & 0.5112       & 0.4614       & 0.3790       & 0.2152       \\
                                  &                                          & B                              & 0.7137       & 0.6287       & 0.5839       & 0.5396       & 0.4930       & 0.4345       & 0.3488       & 0.2262       & 0.0432       \\ \hline
\multirow{4}{*}{GEM}              & \multirow{2}{*}{8}                       & A                              & 0.7125       & 0.6028       & 0.5182       & 0.4608       & 0.4058       & 0.3563       & 0.3086       & 0.2501       & 0.1599       \\
                                  &                                          & B                              & 0.6737       & 0.5611       & 0.4913       & 0.4363       & 0.3788       & 0.3349       & 0.2902       & 0.2344       & 0.1360       \\
\bottomrule 
\end{tabular}
}
\end{table*}
\begin{table*}[!]
\centering
\caption{{TPR, FNR, FPR, and TNR on \textit{eBay-xlarge} by varying thresholds from 0.95 to 0.977 on the prediction scores. \\ FNR = 1 - TPR. The higher TPR is, the better; the lower FNR, the better. \\ FPR = 1 - TNR. The higher TNR is, the better; the lower FPR, the better. \\ ``-" denotes that the scores do not exist for scores $\geqslant$ threshold.}}
\label{tab:tpr-fnr-tnr-fpr-xlarge-part2}
\resizebox{0.9\linewidth}{!}{
\begin{tabular}{ccccccccccccc}
\toprule
\multirow{2}{*}{\textbf{Model}}   & \multirow{2}{*}{\textbf{\# of machines}} & \multirow{2}{*}{\textbf{Seed}} & \multicolumn{10}{c}{\textbf{TPR@threshold}}                                                                                                                                       \\ \cline{4-13}
                                  &                                          &                                & \textbf{0.95}   & \textbf{0.96}   & \textbf{0.97}   & \textbf{0.971}  & \textbf{0.972}  & \textbf{0.973}  & \textbf{0.974}  & \textbf{0.975}  & \textbf{0.976}  & \textbf{0.977}  \\ \hline
\multirow{4}{*}{GAT}              & \multirow{2}{*}{8}                       & A                              & 0.1820          & 0.1303          & 0.0717          & 0.0640          & 0.0595          & 0.0541          & 0.0471          & 0.0412          & 0.0362          & 0.0319          \\
                                  &                                          & B                              & 0.1408 &	0.0903 &	0.0411 &	0.0363 &	0.0326 &	0.0285 &	0.0234 &	0.0188 &	0.0153 &	0.0123
    \\ \cline{2-13}
                                  & \multirow{2}{*}{16}                      & A                              & 0.0380          & 0.0065          & -          & -          & -          & -          & -          & -          & -          & -                \\
                                  &                                          & B                              & 0.0004          & -          & -          & -          & -          & -          & -          & -          & -          & -             \\ \hline
\multirow{4}{*}{GEM}              & \multirow{2}{*}{8}                       & A                              & 0.0699          & 0.0477          & 0.0192          & 0.0168          & 0.0142          & 0.0131          & 0.0113          & 0.0093          & 0.0080          & 0.0057          \\
\bottomrule   
\end{tabular}
}
\end{table*}
\begin{table*}[!]
\centering
\caption{{TPR, FNR, FPR, and TNR on \textit{eBay-xlarge}  by varying thresholds from 0.978 to 0.987 on the prediction scores. \\ FNR = 1 - TPR. The higher TPR is, the better; the lower FNR, the better.  \\ FPR = 1 - TNR. The higher TNR is, the better; the lower FPR, the better. \\ ``-" denotes that the scores do not exist for scores $\geqslant$ threshold.}}
\label{tab:tpr-fnr-tnr-fpr-xlarge-part3}
\resizebox{0.87\linewidth}{!}{
\begin{tabular}{ccccccccccccc}
\toprule
\multirow{2}{*}{\textbf{Model}}   & \multirow{2}{*}{\textbf{\# of machines}} & \multirow{2}{*}{\textbf{Seed}} & \multicolumn{10}{c}{\textbf{TPR@threshold}}                                                                                                                            \\ \cline{4-13}
                                  &                                          &                                & \textbf{0.978} & \textbf{0.979} & \textbf{0.98} & \textbf{0.981} & \textbf{0.982} & \textbf{0.983} & \textbf{0.984} & \textbf{0.985} & \textbf{0.986} & \textbf{0.987} \\ \hline
\multirow{4}{*}{GAT}              & \multirow{2}{*}{8}                       & A                              & 0.0265         & 0.0181         & 0.0130        & 0.0088         & 0.0038         & 0.0008         & 0.0002         & -         & -         & -         \\
                                  &                                          & B                              & 0.0095 &	0.0056	& 0.0036 &	0.0028 &	0.0018 &	0.0006
     & -         & -         & -         \\ \cline{2-13}
                                  & \multirow{2}{*}{16}                      & A                              & -              & -              & -             & -              & -              & -              & -              & -              & -              & -              \\
                                  &                                          & B                              & -              & -              & -             & -              & -              & -              & -              & -              & -              & -          \\ \hline
\multirow{4}{*}{GEM}              & \multirow{2}{*}{8}                       & A                              & 0.0045         & 0.0027         & 0.0018        & 0.0005         & 0.0005         & 0.0004         & -              & -              & -              & -          \\
\bottomrule     
\end{tabular}
}
\end{table*}
\begin{table*}[!]
\centering
\caption{{Precision and recall on \textit{eBay-xlarge} by varying thresholds from 0.1 to 0.9 on the prediction scores. }}

\label{tab:pr-rc-xlarge-part1}
\resizebox{\linewidth}{!}{
\begin{tabular}{cccccccccccc}
\toprule
\multirow{2}{*}{\textbf{Model}}   & \multirow{2}{*}{\textbf{\# of machines}} & \multirow{2}{*}{\textbf{Seed}} & \multicolumn{9}{c}{\textbf{Precision@threshold}}                                                                                     \\ \cline{4-12}
                                  &                                          &                                & \textbf{0.1} & \textbf{0.2} & \textbf{0.3} & \textbf{0.4} & \textbf{0.5} & \textbf{0.6} & \textbf{0.7} & \textbf{0.8} & \textbf{0.9} \\  \hline
\multirow{4}{*}{GAT}              & \multirow{2}{*}{8}                       & A                              & 0.2010       & 0.2478       & 0.2792       & 0.3074       & 0.3362       & 0.3666       & 0.4071       & 0.4592       & 0.6021       \\
                                  &                                          & B                              & 0.1943 &	0.2391 &	0.2706 &	0.2968 &	0.3219 &	0.3544 &	0.3917 &	0.4488 &	0.5755
       \\ \cline{2-12}
                                  & \multirow{2}{*}{16}                      & A                              & 0.1996       & 0.2456       & 0.2797       & 0.3098       & 0.3420       & 0.3852       & 0.4397       & 0.5124       & 0.7124       \\
                                  &                                          & B                              & 0.2109       & 0.2542       & 0.2916       & 0.3285       & 0.3750       & 0.4217       & 0.4796       & 0.6028       & 0.8057       \\ \hline
\multirow{4}{*}{GEM}              & \multirow{2}{*}{8}                       & A                              & 0.2472       & 0.3276       & 0.3822       & 0.4351       & 0.4812       & 0.5329       & 0.5908       & 0.6741       & 0.7904       \\
\bottomrule   
\end{tabular}
}
\end{table*}

\begin{table*}[!]
\centering
\caption{{Precision and recall on \textit{eBay-xlarge} by varying thresholds from 0.95 to 0.977 on the prediction scores. ``-" denotes that the scores do not exist for scores $\geqslant$ threshold.}}

\label{tab:pr-rc-xlarge-part2}
\resizebox{\linewidth}{!}{
\begin{tabular}{ccccccccccccc}
\toprule
\multirow{2}{*}{\textbf{Model}}   & \multirow{2}{*}{\textbf{\# of machines}} & \multirow{2}{*}{\textbf{Seed}} & \multicolumn{10}{c}{\textbf{Precision@threshold}}                                                                                                                    \\ \cline{4-13}
                                  &                                          &                                & \textbf{0.95} & \textbf{0.96} & \textbf{0.97} & \textbf{0.971} & \textbf{0.972} & \textbf{0.973} & \textbf{0.974} & \textbf{0.975} & \textbf{0.976} & \textbf{0.977} \\ \hline
\multirow{4}{*}{GAT}              & \multirow{2}{*}{8}                       & A                              & 0.7832        & 0.8682        & 0.9392        & 0.9568         & 0.9606         & 0.9634         & 0.9642         & 0.9661         & 0.9814         & 0.9882         \\
                                  &                                          & B                              & 0.7647 &	0.8358 &	0.9000	& 0.9034 &	0.9224 &	0.9144 &	0.9331	& 0.9251 &	0.9481 &	0.9759
   \\ \cline{2-13}
                                  & \multirow{2}{*}{16}                      & A                              & 0.9784        & 1.0000        & -             & -              & -              & -              & -              & -              & -              & -              \\
                                  &                                          & B                              & 1.0000        & -             & -             & -              & -              & -              & -              & -              & -              & -              \\ \hline
\multirow{4}{*}{GEM}              & \multirow{2}{*}{8}                       & A                              & 0.8930        & 0.9372        & 0.9655        & 0.9607         & 0.9541         & 0.9503         & 0.9430         & 0.9313         & 0.9813         & 1.0000         \\
\bottomrule
\end{tabular}
}
\end{table*}

\begin{table*}[!]
\centering
\caption{{Precision and recall on \textit{eBay-xlarge} by varying thresholds from 0.978 to 0.987 on the prediction scores. ``-" denotes that the scores do not exist for scores $\geqslant$ threshold.}}

\label{tab:pr-rc-xlarge-part3}
\resizebox{\linewidth}{!}{
\begin{tabular}{ccccccccccccc}
\toprule
\multirow{2}{*}{\textbf{Model}}   & \multirow{2}{*}{\textbf{\# of machines}} & \multirow{2}{*}{\textbf{Seed}} & \multicolumn{10}{c}{\textbf{Precision@threshold}}                                                                                                                      \\ \cline{4-13}
                                  &                                          &                                & \textbf{0.978} & \textbf{0.979} & \textbf{0.98} & \textbf{0.981} & \textbf{0.982} & \textbf{0.983} & \textbf{0.984} & \textbf{0.985} & \textbf{0.986} & \textbf{0.987} \\ \hline
\multirow{4}{*}{GAT}              & \multirow{2}{*}{8}                       & A                              & 0.9971         & 0.9958         & 0.9942        & 1.0000         & 1.0000         & 1.0000         & 1.0000         & -              & -              & -              \\
                                  &                                          & B                              & 0.9690 &	0.9487 &	0.9216 &	0.9024 &	0.8519 &	0.6667
  & -         & -              & -              & -              \\ \cline{2-13}
                                  & \multirow{2}{*}{16}                      & A                              & -              & -              & -             & -              & -              & -              & -              & -              & -              & -              \\
                                  &                                          & B                              & -              & -              & -             & -              & -              & -              & -              & -              & -              & -              \\ \hline
\multirow{4}{*}{GEM}              & \multirow{2}{*}{8}                       & A                              & 1.0000         & 1.0000         & 1.0000        & 1.0000         & 1.0000         & 1.0000         & -              & -              & -              & -              \\
\bottomrule
\end{tabular}
}
\end{table*}

\end{document}